%% file: paper.tex
\pdfoutput=1
\documentclass[conference]{IEEEtran}

\usepackage{etex}
\reserveinserts{28}

\usepackage[bookmarks=true,colorlinks,breaklinks=true,draft=false,%
pdfauthor={Hatem Alismail and Brett Browning and Simon Lucey},%
pdfsubject={Visual Odometry Using Bit-Planes},%
pdfkeywords={Visual Odometry; Visual SLAM; photometric invariance; binary descriptors; brightness constancy; direct methods}]{hyperref}

\usepackage{times}
\usepackage[numbers]{natbib}
\usepackage{multicol}
\usepackage{bm}
\usepackage{mathtools}
\usepackage{nicefrac}
\usepackage[font=small,labelfont=bf]{subcaption}
\usepackage[font=small,labelfont=bf]{caption}
\usepackage{wrapfig}
\usepackage{verbatim}
\usepackage{pgf}
\usepackage{tikz}
\usepackage{pgfplots}
\usepackage{booktabs}
\usepackage{flushend}
\usetikzlibrary{plotmarks}
\usetikzlibrary{shapes}
\usepgfplotslibrary{units}
\pgfplotsset{compat=newest}
\pgfplotsset{plot coordinates/math parser=false}
\pgfplotsset{try min ticks=3}
\pgfplotsset{max space between ticks=20pt}
\pgfplotsset{invoke before crossref tikzpicture={\tikzexternaldisable},invoke
  after crossref tikzpicture={\tikzexternalenable}}
\pgfplotsset{every x tick label/.append style={font=\footnotesize, yshift=0.5ex}}
\pgfplotsset{every y tick label/.append style={font=\footnotesize, xshift=0.0ex}}
\tikzset{font={\fontsize{8pt}{12}\selectfont}}

\usepackage{listings}
\newlength\fwidth%
\newlength\fheight%

\usepackage{siunitx}
\usepackage{microtype}

\usepackage{xspace}
\DeclareRobustCommand\onedot{\futurelet\let\token\onedot}
\def\onedot{\ifx\let\token.\else.\null\fi\xspace}

\def\eg{\emph{e.g}\onedot} 
\def\ie{\emph{i.e}\onedot}

\usepackage{amsmath,amssymb}
\usepackage{graphicx}
\usepackage{paralist}

\DeclareMathOperator*{\argmin}{argmin}

\newcommand{\Argmin}[1]{\underset{#1}{\argmin}}

\newcommand{\T}{^{\mathstrut\scriptscriptstyle{\top}}} 
\newcommand{\mb}[1]{\ensuremath{\mathbf{#1}}}    
\newcommand{\mbb}[1]{\ensuremath{\mathbb{#1}}}

\renewcommand{\subsection}[1]{\vspace{5pt} \noindent \textbf{#1:}}

\usepackage[capitalise]{cleveref}

\newif\ifarxiv
\arxivtrue

\begin{document}
\title{Direct Visual Odometry using Bit-Planes}

\author{%
\IEEEauthorblockN{Hatem Alismail, Brett Browning, and Simon Lucey}\\
\IEEEauthorblockA{The Robotics Institute\\Carnegie Mellon University\\
Pittsburgh, PA 15213\\
\texttt{\{halismai,brettb,slucey\}@cs.cmu.edu}}}

\maketitle
\begin{abstract}
Feature descriptors, such as SIFT and ORB, are well-known for their robustness to illumination changes, which has made them popular for feature-based VSLAM\@. However, in degraded imaging conditions such as low light, low texture, blur and specular reflections, feature extraction is often unreliable. In contrast, direct VSLAM methods which estimate the camera pose by minimizing the photometric error using raw pixel intensities are often more robust to low textured environments and blur. Nonetheless, at the core of direct VSLAM is the reliance on a consistent photometric appearance across images, otherwise known as the brightness constancy assumption. Unfortunately, brightness constancy seldom holds in real world applications.

In this work, we overcome brightness constancy by incorporating feature descriptors into a direct visual odometry framework. This combination results in an efficient algorithm that combines the strength of both feature-based algorithms and direct methods. Namely, we achieve robustness to arbitrary photometric variations while operating in low-textured and poorly lit environments. Our approach utilizes an efficient binary descriptor, which we call Bit-Planes, and show how it can be used in the gradient-based optimization required by direct methods. Moreover, we show that the squared Euclidean distance between Bit-Planes is equivalent to the Hamming distance. Hence, the descriptor may be used in least squares optimization without sacrificing its photometric invariance. Finally, we present empirical results that demonstrate the robustness of the approach in poorly lit underground environments.
\end{abstract}

\IEEEpeerreviewmaketitle%

\section{Introduction}
\begin{figure}
\centering
\includegraphics[width=0.8\linewidth]{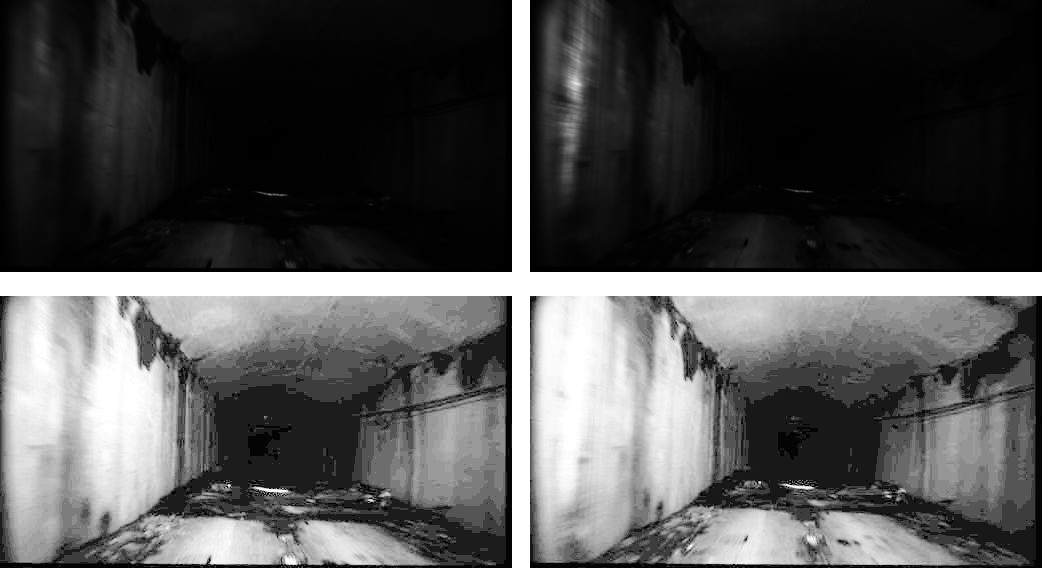}
\caption{Shown at the top row are two consecutive images collected from an underground mine. The bottom row shows a histogram equalization of the images for better visualization. The equalized images may appear noiseless due to its smooth appearance in the document (due to resizing). The data has a low signal-to-noise ratio due to the poor illumination of the scene.}
\label{fig:tunnel}
\end{figure}

\begin{figure}
\centering
\includegraphics[width=0.8\linewidth]{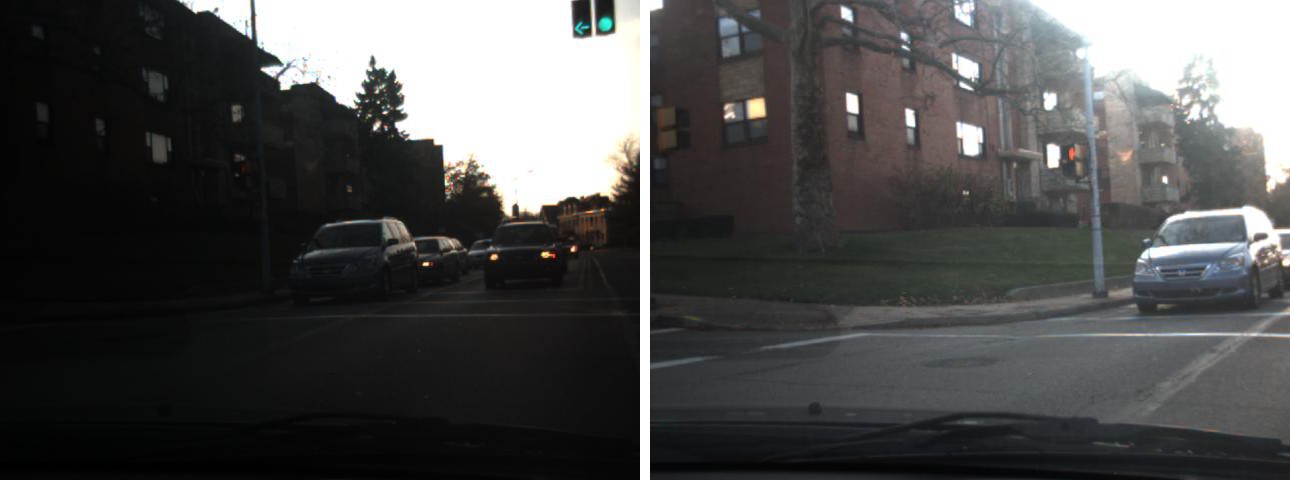}
\caption{An example of the nonlinear intensity deformation caused by the automatic camera settings. A common problem with outdoor applications of robot vision.}
\label{fig:autogain}
\end{figure}

Visual Odometry (VO) is the problem of estimating the relative pose between cameras sharing a common field of view. Due to its importance, VO has received a large amount of attention in the literature as evident by the number of high quality systems freely available to the community~\cite{lsdslam,orbslam2015,forster2014svo,Geiger2012CVPR}. Current systems, however, are not equipped to tackle environments with challenging illumination conditions such as the ones shown in~\cref{fig:tunnel,fig:autogain}. In this paper, our goal is to enable vision-only pose estimation to operate robustly in such challenging environments.

Current state-of-the-art algorithms rely on a \emph{feature-based} pipeline~\cite{torr00}, where keypoint correspondences are used to obtain an estimate of the camera motion (\eg~\cite{nister04,orbslam2015,badino2013visual,mer2,Howard08,Kaess09,fovis,Geiger2011IV}). However, the performance of feature extraction and matching struggles under challenging imaging conditions, such as motion blur, low light, and repetitive texture~\cite{Milford2014,Song2013}. An example environment where keypoint extraction is not easily possible is shown in \cref{fig:tunnel}. If the feature-based pipeline fails, a vision-only system has little hope of recovery.

An alternative to the feature-based pipeline is the use of pixel intensities directly, or what is commonly referred to as \emph{direct methods}~\cite{lk,irani1999direct}. The use of direct methods has been recently popularized with the introduction of the Kinect as a real-time source for high-frame rate depth and RGB data~\cite{klose2013efficient,kerl2013dense,steinbrucker2011real,henry2012rgb,whelan13} as well as monocular SLAM algorithms~\cite{lsdslam,forster2014svo}. When the apparent image motion is small, direct methods deliver more robustness and accuracy as the majority of pixels in the image could be used to estimate a small number of degrees-of-freedom as demonstrated by several authors~\cite{DTAM,comport2010real,forster2014svo,Silveira07}.

Nonetheless, as pointed out by other researchers~\cite{orbslam2015}, the main limitation of direct methods is their reliance on a consistent appearance between the matched pixels, otherwise known as the \emph{brightness constancy} assumption~\cite{horn1981determining,gennert1987relaxing}, which is seldom satisfied in real world applications.

Due to the complexity of real world illumination conditions, an efficient solution to the problem of appearance change for direct VO is challenging. The most common scheme to mitigating the effects of illumination change is to assume a parametric illumination model that is estimated alongside the camera pose, such as the gain and bias model~\cite{klose2013efficient,engel2015_stereo_lsdslam}. This approach is limited by definition and does not address the plethora of the possible non-global and nonlinear intensity deformations. More sophisticated techniques have been proposed~\cite{Silveira07,Yan2015,Meilland12a}, but either assume a certain structure of the scene (such as local planarity) thereby limiting applicability, heavily rely on dense depth estimates (which is not always possible), or significantly increase the dimensionality of the state vector and thereby the computational cost to account for a complex illumination model.

\subsection{Contributions}
In this work, we relax the photometric consistency assumption required by most direct VO algorithms thus allowing them to operate in environments where the  appearance between images vary considerably. We achieve this by combining the illumination invariance afforded by feature descriptors within a direct alignment framework. This in fact is a challenging problem, as conventional illumination invariant feature descriptors are not well-suited to the iterative gradient-based optimization at the heart of direct methods. Our contributions in this work are as follows:

\begin{compactitem}
  \item We combine descriptors from feature-based methods within a direct alignment framework to allow vision-only pose estimation in challenging environments where the current state-of-the-art in feature-based, and direct VSLAM provide unsatisfactory results.
  \item We introduce a binary descriptor~\cite{bitplanes}, show how it can be integrated into a direct method and demonstrate its suitability for linearization. Moreover, we show that under a least squares cost functions the method is equivalent to the Hamming distance over the binary descriptor space. We call this descriptor \emph{Bit-Planes}.
  \item We have previously developed Bit-Planes and demonstrated its performance on template tracking under challenging illumination conditions~\cite{bitplanes}. In this work, we demonstrate the effectiveness of the descriptor on the more challenging task of VO from relatively sparse and noisy depth measurements.
  \item We show that the approach is efficient and robust using a combination of synthetic and real data to evaluate performance as compared to existing method.
\end{compactitem}

\begin{figure}[thbp]
\arxivfalse
  \centering
  \ifarxiv
  \begin{subfigure}[b]{0.25\linewidth}
  \includegraphics{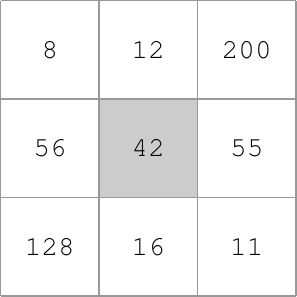}
  \end{subfigure}
  \caption{}\label{fig:ct1}
  \begin{subfigure}[b]{0.25\linewidth}
  \includegraphics{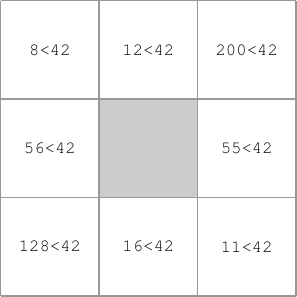}
  \end{subfigure}
  \caption{}\label{fig:ct2}
  \begin{subfigure}[b]{0.25\linewidth}
  \includegraphics{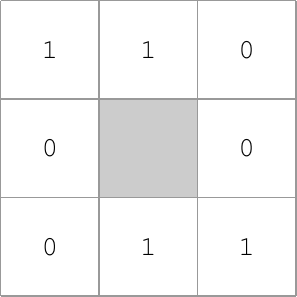}
  \end{subfigure}
  \caption{}\label{fig:ct2}
  \else
  \input{figs/census.tex}
  \fi
  \caption{{\small Local intensity comparisons in a $3\times 3$ neighborhood. In \cref{fig:ct1} the center pixel is highlighted, and compared to its neighbors as shown in \cref{fig:ct2}. The descriptor is obtained by combining the results of each comparison in \cref{fig:ct3} into a single scalar descriptor \cite{lbp,zabih1994non}.}}
  \label{fig:census}
\end{figure}

\begin{figure}
  \centering\small
  \includegraphics[width=\linewidth]{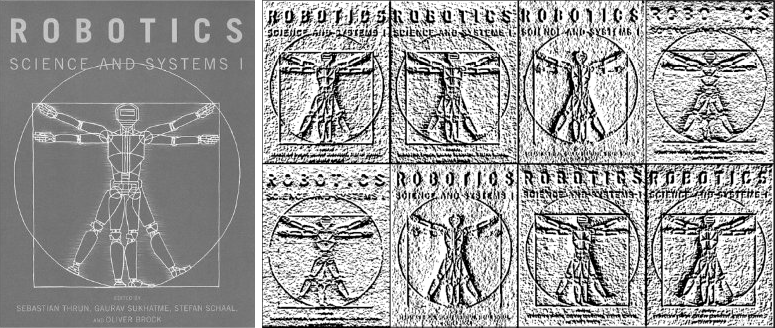}
  \caption{An illustration of our Bit-Planes descriptor where each channel is composed of bits. Since the residual vector is binary, least squares minimization becomes equivalent to minimizing the Hamming distance.}\label{fig:channels}
\end{figure}

\section{Background}
\subsection{Direct Visual Odometry}
Let the intensity, and depth of a pixel coordinate $\mb{p} = {\left(x,~y\right)}\T$ at the \emph{reference} image be respectively given by $\mb{I}(\mb{p}) \in \mbb{R}$ and $\mb{D}(\mb{p}) \in \mbb{R}^+$. Upon a rigid-body motion of the camera a new image is obtained $\mb{I}'(\mb{p}')$. The goal of conventional direct VO is to estimate an increment of the camera motion parameters $\Delta\bm{\theta}\in\mbb{R}^6$ such that the photometric error is minimized
\begin{align}
\label{eq:photo_error}
\Delta\bm{\theta}^\ast = \Argmin{\Delta\bm{\theta}}{%
  \sum_{\mb{p}\in\Omega}{\left\Vert{%
    \mb{I}'\left(\mb{w}(\mb{p}; \bm{\theta} + \Delta\bm{\theta})\right) -
    \mb{I}\left(\mb{p}\right)
  }\right\Vert}^2
},
\end{align}
where $\Omega$ is a subset of pixel coordinates of interest in the reference frame, $\mb{w}\left(\cdot\right)$ is a \emph{warping} function that depends on the parameter vector we seek to estimate, and $\bm{\theta}$ is an initial estimate. After every iteration, the current estimate of parameters is updated additively (\ie~$\bm{\theta} \leftarrow \bm{\theta} + \Delta\bm{\theta}$). The process repeats until convergence, or some termination criteria have been met. This is the well known Lucas and Kanade (forward additive) algorithm~\cite{lk}.

By (conceptually) interchanging the roles of the template and input images, Baker \& Matthews' devise a more efficient alignment techniques known as the Inverse Compositional (IC) algorithm~\cite{baker2004lucas}. Under the IC formulation we seek an update $\Delta\bm{\theta}$ that satisfies
\begin{align}\label{eq:ic}
  {\Delta\bm{\theta}}^\ast=\Argmin{\Delta\bm{\theta}}{\sum_{\mb{p}\in\omega}{\lVert
      \mb{I}\left(\mb{w}(\mb{p}; \Delta\bm{\theta})\right) -
      \mb{I}'\left(\mb{w}(\mb{p}; \bm{\theta})\right) \rVert^2}}.
\end{align}
The optimization problem in \cref{eq:ic} is nonlinear irrespective of the form
of the warping function or the parameters, as in general there is no linear relationship between pixel coordinates and their intensities. By equating the partial derivatives of the first-order Taylor expansion of \cref{eq:ic} to zero, we reach at solution given by the following closed-form (normal equations)
\begin{align}\label{eq:closedform}
  \Delta \bm{\theta} = {\left({\mb{J}}\T{\mb{J}}\right)}^{-1} {\mb{J}}\T \mb{e},
\end{align}
where $\mb{J} = \left({\mb{g}(\mb{p}_1)}\T,~\ldots,~{\mb{g}(\mb{p}_m)}\T\right) \in
\mbb{R}^{m \times p}$ is the matrix of first-order partial derivatives of the objective function, or the Jacobian, $m$ is the number of pixels, and $p = \left\vert \bm{\theta}\right\vert$ is the number of parameters. Each $\mb{g}$ is $\in \mbb{R}^{1\times p}$ and is given by the chain rule as
\begin{align}
\mb{g}(\mb{p})\T = \nabla \mb{I}(\mb{p}) \frac{\partial\mb{w}}{\partial\bm{\theta}},
  \end{align}
where $\nabla \mb{I} = \left(I_x,~I_y\right) \in \mbb{R}^{1 \times 2}$ is the image gradient along the $x$- and $y$- directions respectively. Finally,
\begin{align}
  \mb{e}(\mb{p}) = \mb{I}'(\mb{w}(\mb{p}; \bm{\theta})) -
  \mb{I}(\mb{p})
  \end{align} is the vector of residuals, or the \emph{error image}.
Parameters of the motion model are updated via the IC rule given by %
\begin{align} \label{eq:ic_update}
  \mb{w}\left(\mb{p},
    \bm{\theta}\right) \leftarrow \mb{w}\left(\mb{p}, \bm{\theta}\right) \circ
    {\mb{w}\left(\mb{p}, \Delta\bm{\theta}\right)}^{-1}.
\end{align}
We refer the reader to the excellent series by Baker and Matthews~\cite{baker2004lucas} for a detailed treatment.

\subsection{Image Warping}
Given a rigid body motion $\mb{T}(\bm{\theta})\in SE(3)$ and a depth value $\mb{D}(\mb{p})$ in the coordinate frame of the template image, warping to the coordinate frame of the input image is performed according to
\begin{align}
\mb{p}' = \pi\left(\mb{T}(\bm{\theta})\pi^{-1}(\mb{D}(\mb{p})\right),
\end{align}
where $\pi\left(\cdot\right) : \mbb{R}^3 \rightarrow \mbb{R}^2$ denotes the projection onto a camera with a known intrinsic calibration, and $\pi^{-1}\left(\cdot,\cdot\right) : \mbb{R}^2 \times \mbb{R} \rightarrow \mbb{R}^3$ denotes the inverse of this projection given the camera intrinsics and the pixel's depth. Finally, the intensity values corresponding to the warped input image $\mb{I}(\mb{p}')$ are obtained via a suitable interpolation scheme (bi-linear in this work).

\section{Binary Descriptor Constancy Direct VO}\label{sec:desc_vo}
A limitation of direct method is the reliance on the brightness constancy assumption (\cref{eq:photo_error}). To circumvent this shortcoming, we propose the use of a \emph{descriptor constancy} assumption. Namely, we seek an update to the pose parameters such that
\begin{align}
\label{eq:desc_error}
\Delta\bm{\theta}^\ast = \Argmin{\Delta\bm{\theta}}{%
  {\left\Vert{%
        \bm{\phi}(\mb{I}'\left(\mb{w}(\mb{p}; \bm{\theta} + \Delta\bm{\theta})\right)) -
        \bm{\phi}(\mb{I}\left(\mb{p}\right))
  }\right\Vert}^2
},
\end{align}
where $\bm{\phi(\cdot)}$ is a feature descriptor such as SIFT~\cite{sift}, HOG~\cite{hog}, or ORB~\cite{orb}. The idea of using descriptors in lieu of intensity has been recently explored in optical flow estimation~\cite{SevillaLara2014}, template tracking~\cite{bitplanes}, image-based tracking of a known 3D model~\cite{crivellaro2014robust}, Active Appearance Models~\cite{Antonakos15}, and inter-object category alignment~\cite{bristow14}, in which results are very encouraging and consistently outperform classical minimization of the photometric error. To date, however, the idea has not been explored in the context of real-time direct VO with relatively sparse and noisy depth estimates.

The descriptor constancy objective in \cref{eq:desc_error} is more complicated than its brightness counterpart in \cref{eq:photo_error} as interesting feature descriptors are high dimensional and the suitability of their linearization remains unclear, and is only beginning to be investigated in the literature~\cite{Antonakos15,bristow14}.

In VO, however, another challenge is the need for an efficient feature descriptor suitable for real-time implementation. Densely computing classical descriptors such as HOG~\cite{hog} and SIFT~\cite{sift} becomes infeasible for real-time performance with a limited computational budget. Simpler descriptors, such as photometrically normalized image patches, or the gradient-constraint~\cite{Brox2004}, are efficient to compute, but do not possess sufficient invariance to radiometric changes in the wild. Additionally, since reliable depth estimates from stereo are sparse, warping feature descriptors is challenging as it harder to reason about visibility and occlusions from sparse 3D points.

In this work, we propose a novel descriptor that satisfies the requirements for efficient direct alignment under challenging illumination, we also show its suitability for VO from relatively sparse depth data. Namely, our descriptor has the following properties:
\begin{compactitem}
\item Invariance to monotonic changes in intensity, which is important as many robotic applications rely on automatic camera gain and exposure control. Gain adjustment (gamma correction) such as the one shown in \cref{fig:autogain} results in nonlinear changes in image appearance that are not captured by additive, or multiplicative terms.
\item Computational efficiency, even on embedded devices, which is required for real-time VO, and
\item Suitability for least squares minimization (\eg\cref{eq:desc_error}). This is important for two reasons. First, solutions to least squares problems are the among the most computationally efficient optimization problems with a plethora of ready to use software packages. Second, due to the small residuals nature of least squares, only first-order derivatives are required to compute a good approximation of the Hessian. Hence, increasing efficiency and avoiding numerical errors associated with estimating second-order derivatives that arise, for example, when using alignment algorithms based on intrinsically robust objectives~\cite{Dowson2008,Irani98,Dame2010}.
\end{compactitem}

\subsection{The Bit-Planes Descriptor} The rationale behind binary descriptors is that using relative changes of pixel intensities is more robust than working with the raw values for correspondence estimation tasks. As with all binary descriptors, we perform local comparisons between the pixel and its immediate neighbors as shown in \cref{fig:census}. We found that a $3\times 3$ neighborhood is sufficient when working with high-frame rate data and is the most efficient to compute. This is step is identical to the Census Transform~\cite{zabih1994non}, also known as LBP~\cite{lbp}. Choice of the comparison operator is arbitrary and will be denoted with $\bowtie \in\{>,\ge,<,\le\}$. Since the binary representation of the descriptor requires eight comparisons in a $3\times 3$ neighborhood, it is commonly compactly stored as a single byte according to
\begin{align}
\bm{\phi}_\textsc{byte}(\mb{x}) = \sum_{i=1}^8 2^{i-1}
  \big[ \mb{I}(\mb{x}) \bowtie \mb{I}(\mb{x} + \Delta \mb{x}_{i})\big],
\end{align}
where~${\{\Delta \mb{x}_{i}\}}_{i=1}^{8}$ is the set of the eight relative coordinate displacements that are possible within a $3 \times 3$ neighborhood around the center pixel location~$\mb{x}$.

In order for the descriptor to maintain its morphological invariance to intensity changes it must be matched under a binary norm, such as the Hamming distance. The reason for this is easy to illustrate with an example. Consider two bit patterns differing at a single bit --- which so happens to be at the most significant position ---
\begin{align*}
\mb{a} &= \left\{\mathtt{\textcolor{red}{1},0,1,0,1,1,1,0}\right\},\text{ and}\\ \mb{b} &= \left\{\mathtt{\textcolor{red}{0},0,1,0,1,1,1,0}\right\}.
\end{align*}
The two bit-strings are clearly similar and their distance under the Hamming norm is $1$. However, if the decimal representation is used and matched under the squared Euclidean distance, their distance becomes $128^2 = 16384$, which does not capture their closeness in the descriptor space. Hence, using the descriptor in its decimal form is no longer invariant to monotonic intensity change. Nonetheless, it is not possible to use the Hamming distance in least squares optimization, and so it is usually approximated~\cite{Vogel2013}, but at the cost of reduced invariance.

In our proposed descriptor, we avoid the approximation of the Hamming distance and instead store each bit/coordinate of the descriptor as its own image, namely
\begin{align}
\bm{\phi}_{\textsc{bp}}(\mb{x}) = \begin{bmatrix} \mb{I}(\mb{x}) \bowtie \mb{I}(\mb{x} + \Delta \mb{x}_{1}) \\
\vdots \\ \mb{I}(\mb{x}) \bowtie \mb{I}(\mb{x} + \Delta \mb{x}_{8})
\end{bmatrix} \in \mbb{R}^8 \;.
\label{eq:bitplanes}
\end{align}
Since each coordinate of the 8-vector descriptor is binary (as shown in \cref{fig:channels}), we call our descriptor ``Bit-Planes.'' Using this descriptor it is now possible to minimize an equivalent form of the Hamming distance using ordinary least squares.

\subsection{Bit-Planes implementation details} In order to reduce the sensitivity of the descriptor to noise, the image is smoothed with a Gaussian filter in a $3\times 3$ neighborhood ($\sigma=0.5$). The effect of this smoothing will be investigated in \cref{sec:experiments}. Since the operations involved in extracting the descriptor are simple and data parallel, they can be done efficiently with SIMD (Single Instruction Multiple Data) instructions~\cite{lbpsimd1,lbpsimd2,lbpsimd3}. A  pseudo code of our implementation is shown in \cref{code:simd}.
\begin{lstlisting}[language=C,frame=single,%
  caption={\small Data parallel impelmentation, where \texttt{src} and \texttt{dst} indicate pointers to the source and destination images. The symbol \texttt{load} denotes loading the line of data into a SIMD register. With SSE2, for example, 16 pixels are processed at once. Similarly, for the \texttt{set} operator that fills the register with a single value. The comparision operator '\texttt{>}' along with the logical AND '\texttt{\&}' and logical OR '\texttt{|}' are overloaded for SIMD types using the appropriate machine instructions.},label=code:simd]
  c = set(src[0]);
  dst = ((load(src[-stride-1] > c)) & 0x01) |
        ((load(src[-stride  ] > c)) & 0x02) |
        ((load(src[-stride+1] > c)) & 0x04) |
        ((load(src[       +1] > c)) & 0x08) |
        ((load(src[       -1] > c)) & 0x10) |
        ((load(src[+stride-1] > c)) & 0x20) |
        ((load(src[+stride  ] > c)) & 0x40) |
        ((load(src[+stride+1] > c)) & 0x80) ;
\end{lstlisting}

\subsection{Pre-computing descriptors for efficiency} Descriptor constancy as stated in~\cref{eq:desc_error} requires re-computing the descriptors after every iteration of image warping. In addition to the extra computational cost of repeated applications of the descriptor, it is difficult to warp individual pixel locations with sparse depth. An approximation to the descriptor constancy objective in~\cref{eq:desc_error} is to pre-compute the descriptors and minimize the following expression instead:
\begin{align}\label{eq:desc_error_approx}
\min_{\Delta\bm{\theta}} \sum_{\mb{p}\in\Omega}{%
\sum_{i=1}^8{%
\lVert\bm{\Phi}'_i(\mb{w}(\mb{p}; \bm{\theta} + \Delta\bm{\theta})) - \bm{\Phi}_i(\mb{p})\rVert^2}
},
\end{align}
where $\bm{\Phi}_i$ indicates the $i$-th coordinate of the pre-computed descriptor. We found that the loss of accuracy caused by using~\cref{eq:desc_error_approx} instead of~\cref{eq:desc_error} to be insignificant in comparison to the computational savings.

\section{VO using Bit-Planes}
We will use \cref{eq:desc_error_approx} as our objective function, which we minimize using the Baker and Matthews' IC formulation~\cite{baker2004lucas}, allowing us to pre-compute the Jacobian of the cost function. The Jacobian is given by
\begin{align}
\sum_{\mb{p} \in \Omega}
\sum_{i=1}^8 \mb{g}_i(\mb{p};\bm{\theta})\T \mb{g}_i(\mb{p};\bm{\theta}),
\end{align}
where
\begin{align}
\label{eq:grad}
\mb{g}_i(\mb{q};\bm{\theta}) = \frac{\partial\bm{\Phi}_i}{\partial\mb{p}}\Bigr\rvert_{\mb{p}=\mb{q}}
\frac{\partial \mb{w}}{\partial\bm{\theta}}
\Bigr\rvert_{\mb{p}=\mb{q}, \bm{\theta}=\mb{0}} \;\;.
\end{align}

Similar to other direct VO algorithms~\cite{kerl13icra,klose2013efficient,alismail14dds} pose parameters are represented with the exponential map, \ie $\bm{\theta}=\left[\bm{\omega},~\bm{\nu}\right]\T\in\mbb{R}^6$. A rigid-body transformation $\mb{T}(\bm{\theta})\in SE(3)$ is obtained in closed-form for $\theta  = \lVert\bm{\omega}\rVert \ne 0$, $s = \sin\theta$, and $\bar{c}=1-\cos\theta$ as
\begin{align}
\exp\left(\bm{\theta}\right) &=
\begin{bmatrix}
\mb{I} + s\left[\bm{\omega}\right]_\times +
\bar{c}\left[\bm{\omega}\right]_\times^2 & \mb{A}(\bm{\omega}) \bm{\nu} \\
\mb{0}\T & 1
\end{bmatrix} \; \in SE(3), \\
\mb{A}(\bm{\omega}) &= \mb{I} + \frac{\bar{c}}{\theta} \left[\bm{\omega}\right]_\times + \frac{\theta - s}{\theta}
\left[\bm{\omega}\right]_\times^2\; \in \mbb{R}^{3\times 3}\;.
\end{align}

\begin{figure}
\centering
\includegraphics[width=0.5\linewidth]{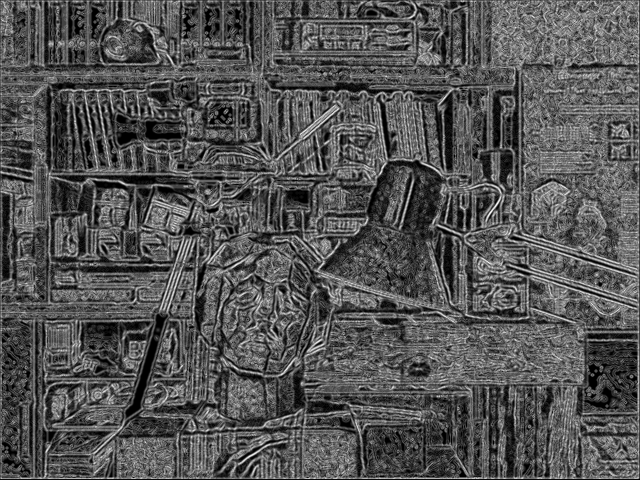}
\caption{The absolute gradient magnitude of Bit-Planes over all channels. Darker values are smaller.}
\label{fig:grad_map}
\end{figure}

To improve the computational running time of the algorithm, we subsample pixel locations for use in direct VO\@. A pixel is selected for the optimization if its absolute gradient magnitude is non-zero and is a strict local maxima in a $3\times 3$ neighborhood. The intuition for this pixel selection procedure is that pixels with a small gradient magnitude contribute little, if any, to the objective function as the term in \cref{eq:grad} vanishes. We compute the pixel saliency map for all eight Bit-Planes coordinate as
\begin{align}
\mb{G} = \sum_{i=1}^8 \sum_{\mb{p}} \left|\nabla_x\bm{\Phi}_i(\mb{p})\right| +
\left|\nabla_y\bm{\Phi}_i(\mb{p})\right|.
\end{align}
An example of this saliency map is shown in \cref{fig:grad_map}. This pixel selection procedure is performed if the image resolution is at least $320\times 240$. For lower resolution images (coarser pyramid levels) we use all pixels with non-zero saliency without the non-maxima suppression step.

Minimization of the objective function is performed using an iteratively re-weighted Gauss-Newton algorithms. The weights are computed using the Tukey bi-weight function~\cite{tukey} given by
\begin{align}
  \rho(r_i; \tau) = \begin{cases}
    {\left(1 - {\left(r_i / \tau\right)}^2\right)}^2 &\text{if } |r_i| \le \tau;\\
    0 &\text{otherwise}.
    \end{cases}
\end{align}
where $r_i$ is the $i$-th residual. The cutoff threshold $\tau$ is set to $4.6851$ to obtain a $95\%$ asymptotic efficiency of the normal distribution. The threshold assumes normalized residuals with unit deviations. For this purpose, we use a robust estimator of standard deviation. For $m$ observations and $p$ parameters, the robust standard deviation is given by:
\begin{align}\label{eq:stddev}
  \hat{\sigma} = 1.4826\left[1 + 5/(m - p)\right] \underset{i}{\mathrm{median}} |r_i|.
\end{align}
The constant $1.4826$ is used to obtain the same efficiency of least squares under Gaussian noise, while $\left[1 + 5/(m-p)\right]$ is used to compensate for small data~\cite{zhang1997parameter}. In practice, $m\gg p$ and the small data constant vanishes.

The approach is implemented in coarse-to-fine manner. The number of pyramid octaves is selected such that the minimum image dimension at the coarsest level is at least $40$ pixels. Termination conditions for Gauss-Newton are fixed to either a maximum number of iterations ($50$), or if the relative change in the estimated parameters, or the relative reduction of the objective function value, falls below $1\times 10^{-6}$.

Finally, we implement a simple keyframing strategy to reduce drift accumulation over time. A keyframe is created if the magnitude of motion exceeds a threshold (data dependent), or if the percentage of ``good points'' falls below $60\%$. A point is deemed good if its weight from the M-Estimator is at the top $80$-percentile. Points that project outside the image (\ie no longer visible) are assigned zero weight.

In the experiments section, we use a calibrated stereo rig to fixate the scale ambiguity and obtain a metric reconstruction directly. The photometric (and descriptor) error is evaluated using the left image only. The approach is possible to extend to monocular systems using a suitable depth initialization scheme~\cite{lsdslam,forster2014svo}, and is readily available for RGB-D sensors.

\section{Experiments \& Results}\label{sec:experiments}
\subsection{Effect of pre-computing descriptors} 
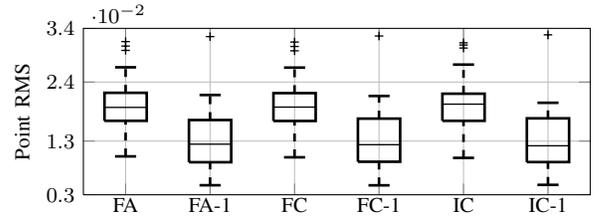
\begin{figure}
  \centering
  \small
  \setlength\fwidth{0.8\linewidth}
  \setlength\fheight{0.25\linewidth}
  \input{figs/alg_cmp.tex}
  \caption{Recomputing descriptors \emph{after} image warping shows consistently better performance than warping feature images when tested with several LK variants. FA: Forward Addition, FC: Forward Compositional, IC: Inverse Compositional. The suffix `-1` indicates \emph{recomputed} features on warped images.}\label{fig:features_recompute}
\end{figure}

Using the approximated descriptor constancy term in~\cref{eq:desc_error_approx} instead of~\cref{eq:desc_error} is slightly less accurate as shown in~\cref{fig:features_recompute}. But, favoring the computational savings, we opt to use the approximated form of the descriptor constancy.

\subsection{Effect of smoothing}
\begin{figure}
\centering
\arxivtrue
\ifarxiv
\includegraphics{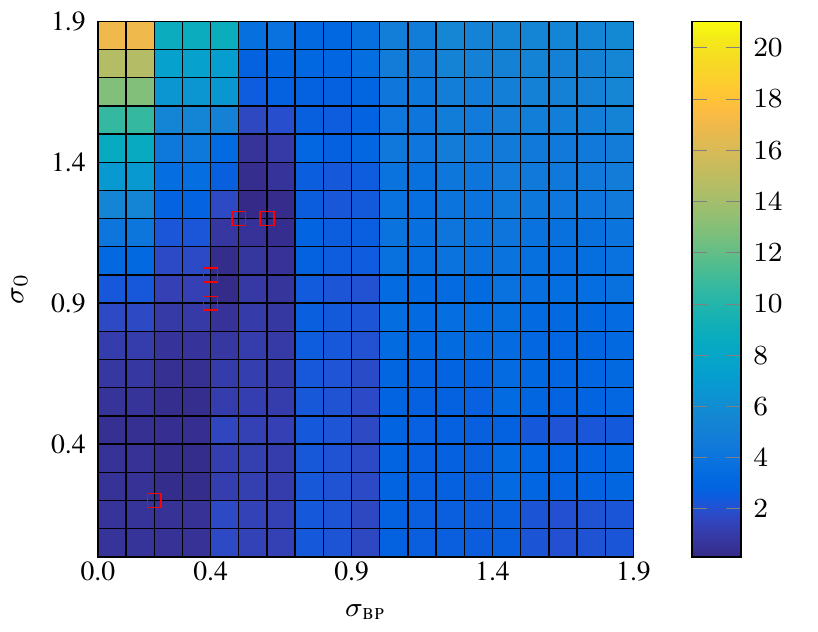}
\else
\setlength\fwidth{3.0in}
\setlength\fheight{2.5in}
\input{figs/sigma.tex}
\fi
\caption{Error as function of pre-smoothing the image with a Gaussian kernel of standard deviation of $\sigma_0$ as as well smoothing the Bit-Planes with $\sigma_1$. The lowest error is associated with smaller kernels.}
\label{fig:sigma}
\end{figure}

\cref{fig:sigma} shows the effect of smoothing the image prior to computing Bit-Planes. The experiment is performed on synthetic data with a small translational shift. Higher smoothing kernels tend to washout the image details required to estimate small motions. Hence, we use a $3\times 3$ kernel with $\sigma=0.5$.

\subsection{Runtime} There are two steps to the algorithm. First, is pre-computing the Jacobian of the cost function as well as the Bit-Planes descriptor for the reference image. This is required only when a new keyframe is created. We call this step \textbf{Jacobians}. The second step, which is repeated at every iteration consists of: (i) image warping using the current estimate of pose, (ii) computing the Bit-Planes error, (iii) computing the residuals and estimating their standard deviations, and (iv) building the weighted linear system and solving it. We call this image \textbf{Linearization}. The running time for each step is summarized in \cref{table:runtime} as a function of image resolution and the selected image points. A typical number of iterations for a run using stereo computed with block matching is shown in \cref{fig:tsukuba_intensity_stats,fig:tsukuba_bitplanes_stats}. The bottleneck in the \textbf{Linearization} step is computing the median absolute deviation of the residuals, which could be mitigated by approximating the median with histograms~\cite{klose2013efficient}. Timing for each of the major steps is shown in \cref{table:runtime,table:timing_steps_kitti}.

\begin{table}
\centering
\caption{{\small Execution time of each major step in the algorithm reported in milliseconds (ms) and using four pyramid levels. The construction of the image pyramid is a common step for both raw intensity and Bit-Planes. Descriptor computation step for raw intensity amounts to converting the image to floating point. Jacobian pre-computation is required only when creating a new keyframe. The most commonly performed operation is descriptor warping, which is not significantly more expensive than warping a single channel of raw intensity. Note, both descriptors (raw intensity and Bit-Planes) use the same generic code base. It is possible to obtain additional speedups by optimizing the code for Bit-Planes specifically. For larger images see \cref{table:timing_steps_kitti}.}}
\label{table:runtime}
\begin{tabular}{r cc}
\toprule
   & Raw Intensity & Bit-Planes \\ \midrule
Pyramid construction   & \multicolumn{2}{c}{$0.31$}  \\
Descriptor computation & $0.18$  & $4.33$            \\
Jacobian pre-computation & $3.34$ & $10.47$          \\
Descriptor warping       & $0.35$ & $1.65$ \\ \bottomrule
\end{tabular}
\end{table}
\begin{table}
\centering
\caption{Execution time of each major step of the algorithm using the KITTI dataset with image size $1024\times 376$. Please refer to \cref{table:runtime} for details.}
\label{table:timing_steps_kitti}
\begin{tabular}{r cc}
\toprule
   & Raw Intensity & Bit-Planes \\ \midrule
Pyramid construction   & \multicolumn{2}{c}{$0.44$}  \\
Descriptor computation & $0.28$  & $5.55$            \\
Jacobian pre-computation & $5.00$ & $13.92$          \\
Descriptor warping       & $0.30$ & $1.74$ \\ \bottomrule
\end{tabular}
\end{table}

\begin{figure*}
\centering
\setlength\fwidth{0.4\linewidth}
\setlength\fheight{0.3\linewidth}
\input{figs/tsukuba_intensity_9_iters.tex}
\input{figs/tsukuba_intensity_9_time.tex}
\caption{Number of iterations and runtime on the first $500$ frames of the New Tsukuba dataset using raw intensity only. On average, the algorithm runs at more than $100$ Hz.}
\label{fig:tsukuba_intensity_stats}
\end{figure*}
\begin{figure*}
\centering
\setlength\fwidth{0.4\linewidth}
\setlength\fheight{0.3\linewidth}
\input{figs/tsukuba_bitplanes_9_iters.tex}
\input{figs/tsukuba_bitplanes_9_time.tex}
\caption{Number of iterations and run time using on the first $500$ frames of the New Tsukuba dataset using the Bit-Planes descriptors. On average, the algorithm runs at $15$ Hz.}
\label{fig:tsukuba_bitplanes_stats}
\end{figure*}

\subsection{Experiments with synthetic data}
\begin{figure}
\centering
\includegraphics[width=0.75\linewidth]{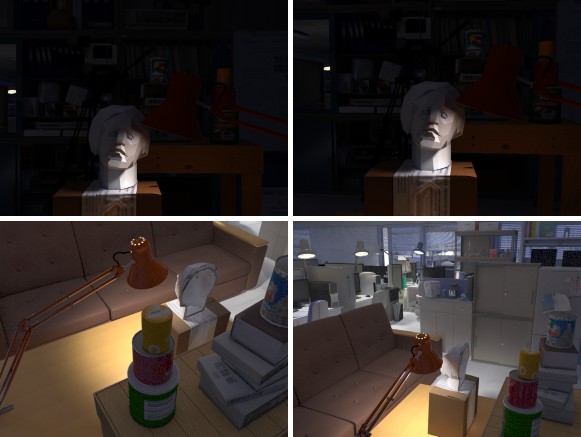}
\caption{Example images from the ``lamps'' sequence.}
\label{fig:tsukuba_example}
\end{figure}

We use the ``New Tsukuba'' dataset~\cite{tsukuba_1,tsukuba_2} to compare the performance of our algorithm against two representative algorithms from the state-of-the-art. The first, is FOVIS~\cite{fovis}, which we use as a representative of feature-based methods. The second, is DVO~\cite{kerl13icra} as representative of direct methods using the brightness constancy assumption. The most challenging illumination condition provided by the Tsukuba dataset is when the scene is lit by ``lamps'' as shown in \cref{fig:tsukuba_example}, which we use in our evaluation.

\begin{figure}
\centering
\arxivtrue
\ifarxiv
\includegraphics{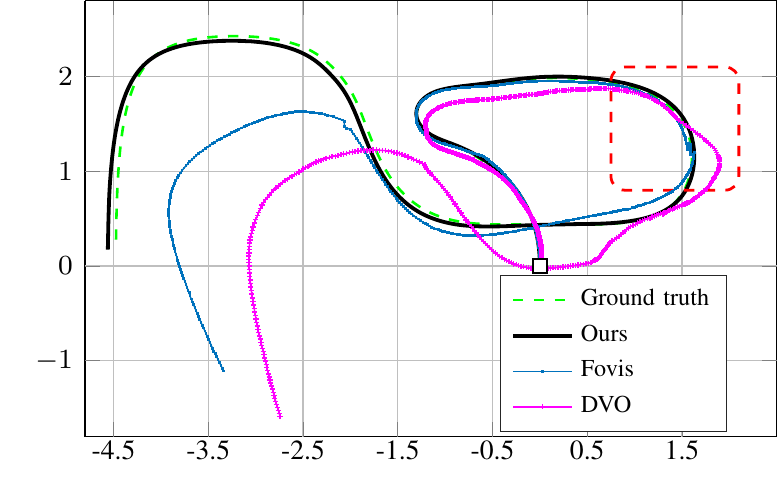}
\else
\setlength\fwidth{\linewidth}
\setlength\fheight{0.5\linewidth}
\input{figs/eval_tsukuba_gt.tex}
\fi
\caption{Evaluation on the synthetic Tsukuba sequence \cite{tsukuba_1} using the illumination provided by ``lamps'' in comparison to other VO algorithms. The figure shows a bird's eye view of the estimated trajectory of the camera form each algorithm in comparison to the ground truth. The highlighted area is shown with more details in \cref{fig:tsukuba_zoom}. Example images are in \cref{fig:tsukuba_example}.}
\label{fig:tsukuba}
\end{figure}

\begin{figure}
\centering
\arxivtrue
\ifarxiv
\includegraphics{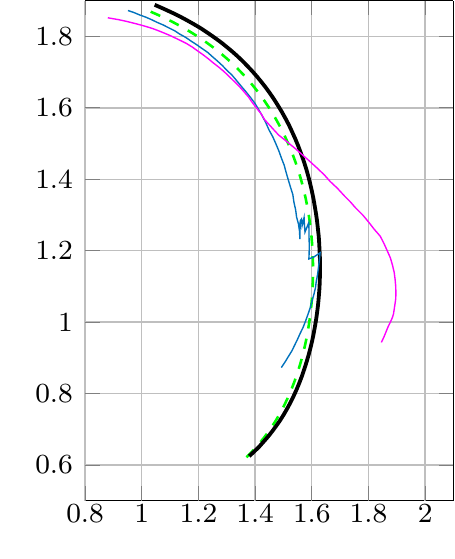}
\else
\setlength\fwidth{\linewidth}
\setlength\fheight{2.0in}
\input{figs/eval_tsukuba_gt_zoom.tex}
\fi
\caption{Estimated camera path details for each of the algorithms shown in~\cref{fig:tsukuba}.}
\label{fig:tsukuba_zoom}.
\end{figure}

Our goal is this experiment is to assess the utility of our proposed descriptor in handling the arbitrary change in illumination visible in all frames of the dataset. Hence, we initialize all algorithms with the ground truth disparity/depth map. In this manner, any pose estimation errors are caused by failures to extract and match features, or failure in minimizing the photometric error. As shown in~\cref{fig:tsukuba} and~\cref{fig:tsukuba_zoom} the robustness of our approach far exceeds the conventional state-of-the-art. Also, as expected, feature-based methods in this case (FOVIS) slightly outperforms direct methods (DVO) due to the challenging illumination of the scene.

\subsection{Evaluation on the KITTI benchmark}
The KITTI benchmark~\cite{Geiger2012CVPR} presents a challenging dataset for our algorithm, and all direct methods in general, as the motion between consecutive frames is large. The effect of large motions can be observed in \cref{fig:kitti_eval}, where the performance of our algorithm noticeably degrades at higher vehicle speeds. This limitation could be mitigated by using a higher camera frame rate, or providing a suitable initialization.

\begin{figure*}
\centering
\setlength\fwidth{0.8\linewidth}
\setlength\fheight{2.5in}
\input{figs/kitti_eval_latest.tex}
\caption{Performance on the training data of the KITTI benchmark in comparison to VISO2~\cite{Geiger2011IV}. The large baseline between consecutive frames presents a challenge to direct methods as can been seen by observing the error as a function of speed. Nonetheless, rotation accuracy of our method remains high.}
\label{fig:kitti_eval}
\end{figure*}

\subsection{Real data from underground mines}
\begin{figure*}
\centering
\begin{subfigure}[t]{\textwidth}
\includegraphics[width=\linewidth]{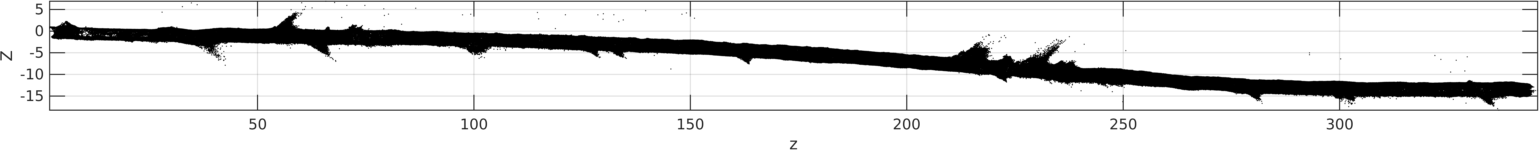}
\caption{Long section of $\approx 400$ meters of robust VO in a poorly lit underground environments.}
\label{fig:tunnel_1}
\end{subfigure}
\begin{subfigure}[t]{0.19\textwidth}
\includegraphics[width=\linewidth]{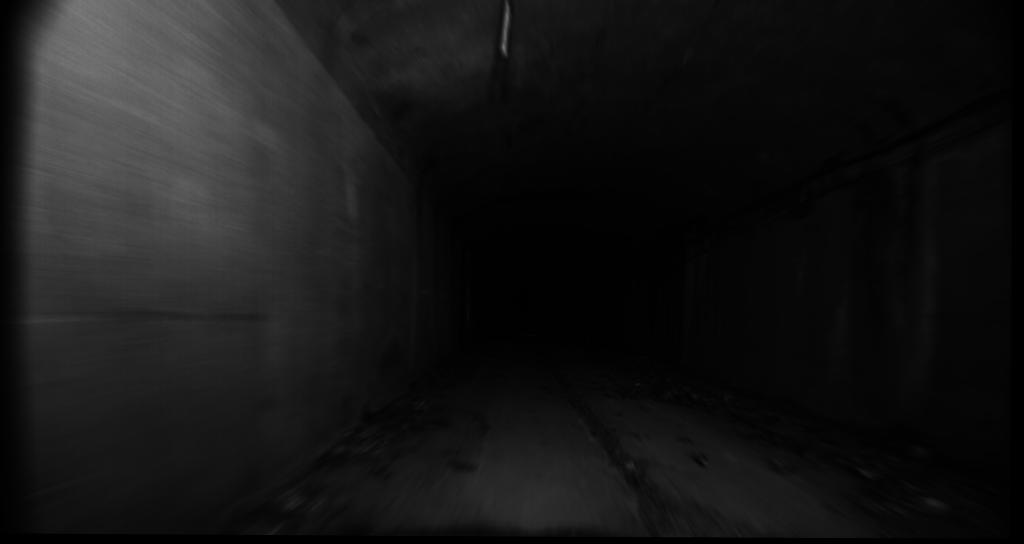}
\end{subfigure}
\begin{subfigure}[t]{0.19\textwidth}
\includegraphics[width=\linewidth]{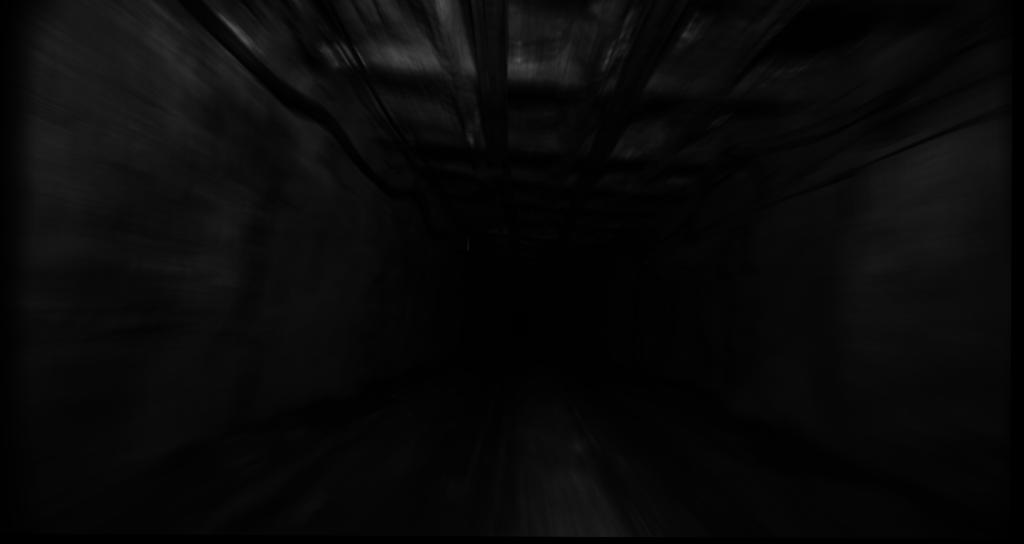}
\end{subfigure}
\begin{subfigure}[t]{0.19\textwidth}
\includegraphics[width=\linewidth]{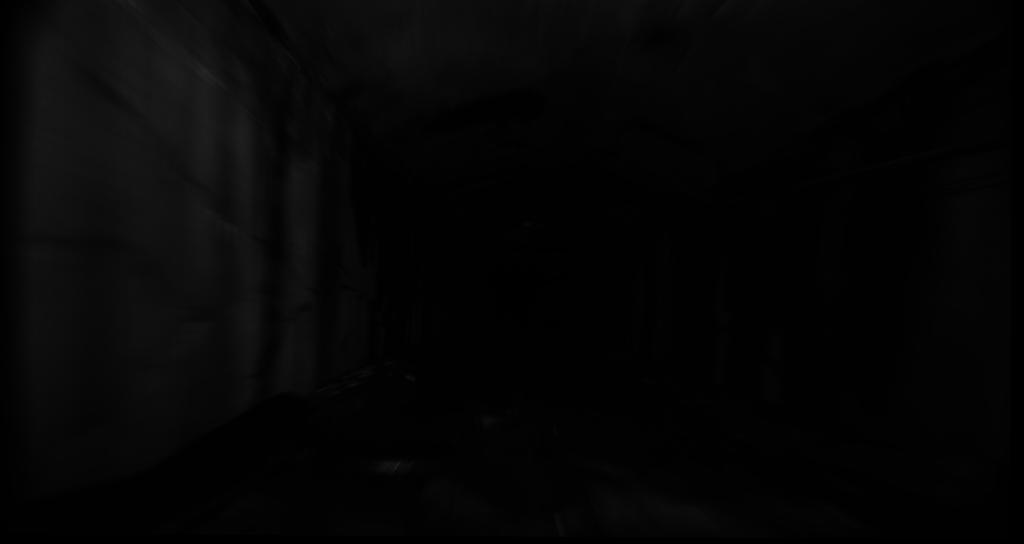}
\end{subfigure}
\begin{subfigure}[t]{0.19\textwidth}
\includegraphics[width=\linewidth]{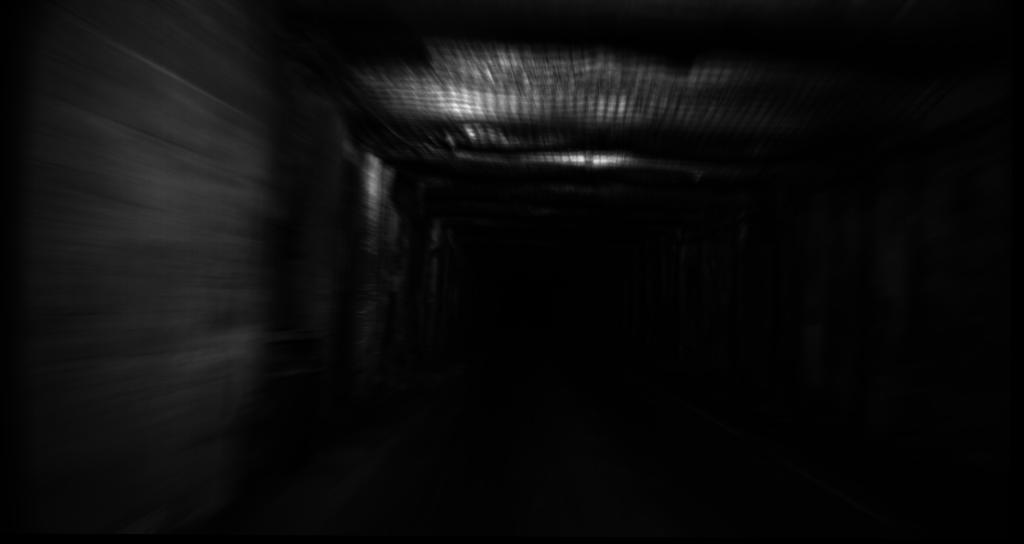}
\end{subfigure}
\begin{subfigure}[t]{0.19\textwidth}
\includegraphics[width=\linewidth]{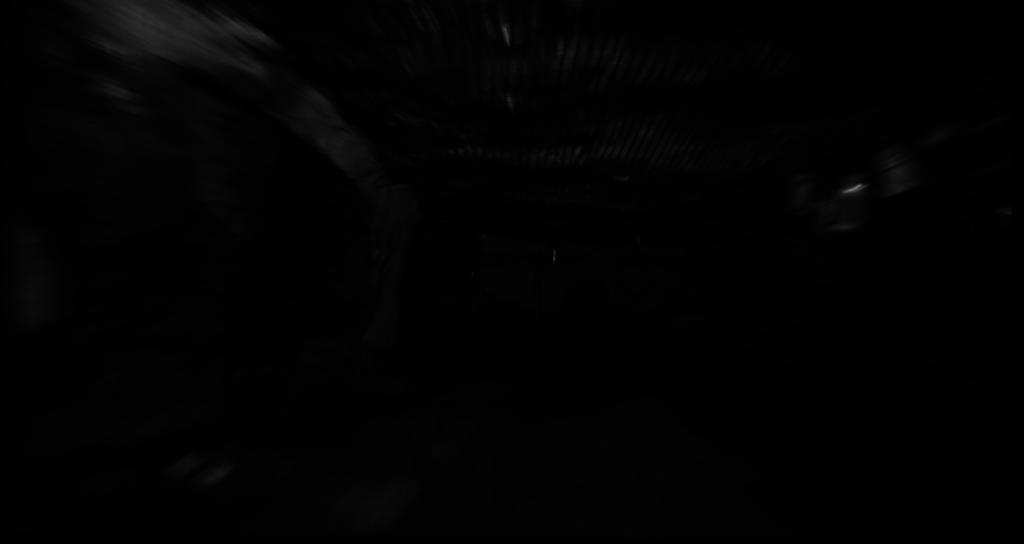}
\end{subfigure}

\begin{subfigure}[t]{0.19\textwidth}
\includegraphics[width=\linewidth]{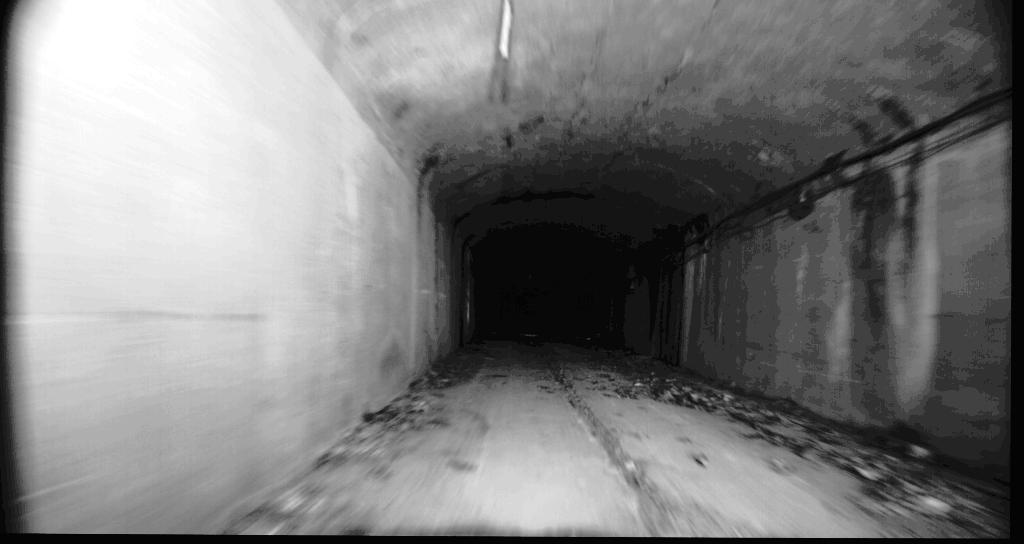}
\end{subfigure}
\begin{subfigure}[t]{0.19\textwidth}
\includegraphics[width=\linewidth]{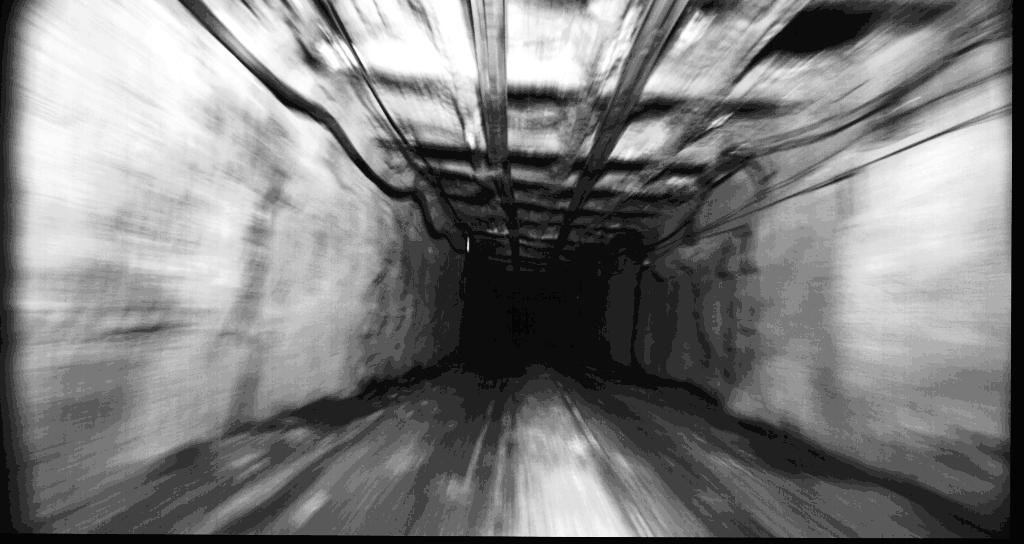}
\end{subfigure}
\begin{subfigure}[t]{0.19\textwidth}
\includegraphics[width=\linewidth]{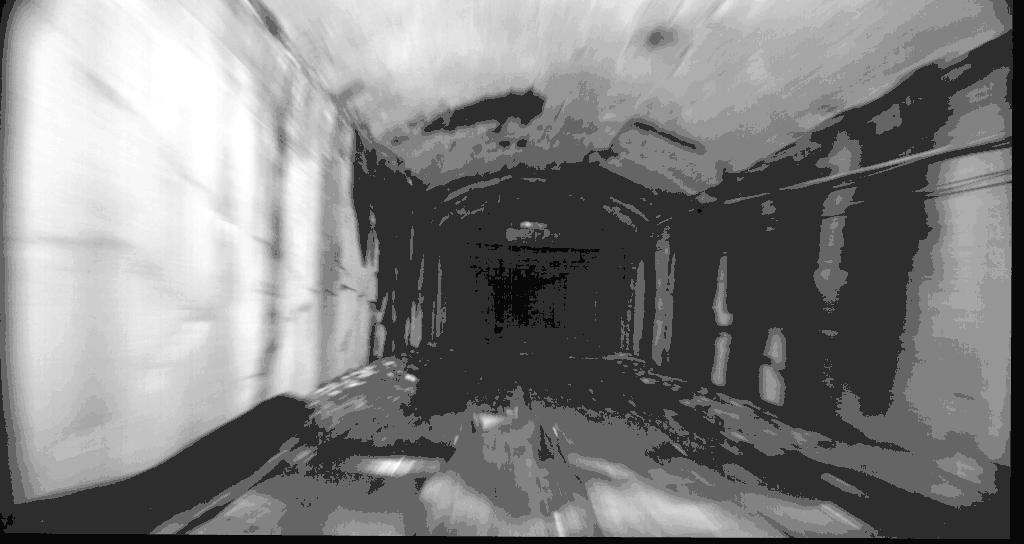}
\end{subfigure}
\begin{subfigure}[t]{0.19\textwidth}
\includegraphics[width=\linewidth]{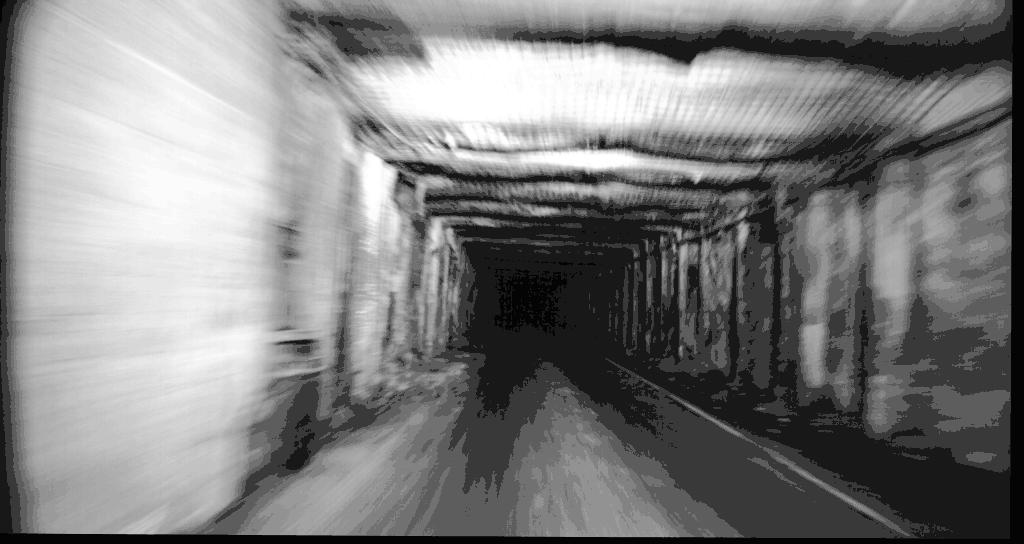}
\end{subfigure}
\begin{subfigure}[t]{0.19\textwidth}
\includegraphics[width=\linewidth]{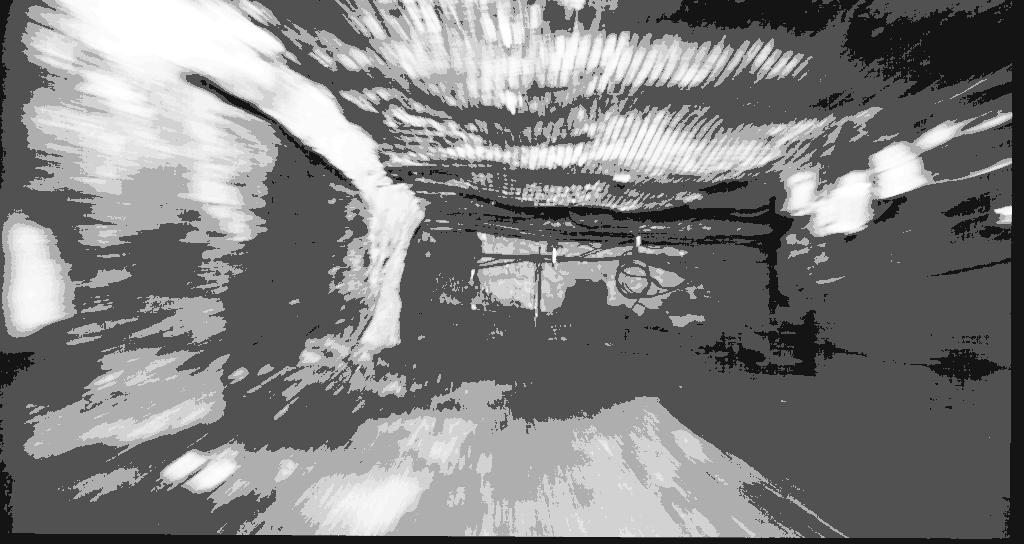}
\end{subfigure}
\caption{Example result and representative images from the first mine sequence (top row) and a histogram equalized version for visualization (bottom row).}
\label{fig:tunnel_results}
\end{figure*}

We demonstrate the robustness of our algorithm using data collected in underground mines. Our robot is equipped with a stereo camera with $7$cm baseline that outputs grayscale images of size $1024\times 544$ and computes an estimate of disparity using a hardware implementation of SGM~\cite{sgm}. An example VO result along with a sample of the data is shown in~\cref{fig:tunnel_results}.

Due to lack of lighting in underground mines, the robot carries its own source of LED light. However, the LEDs are insufficient to uniformly illuminate the scene due to power constraints in the system. We have attempted to use other open source VSLAM/VO packages~\cite{fovis,orbslam2015,Geiger2011IV}, but they all fail too often due to the severely degraded illumination conditions.

\begin{figure}
\centering
\includegraphics[width=0.75\linewidth]{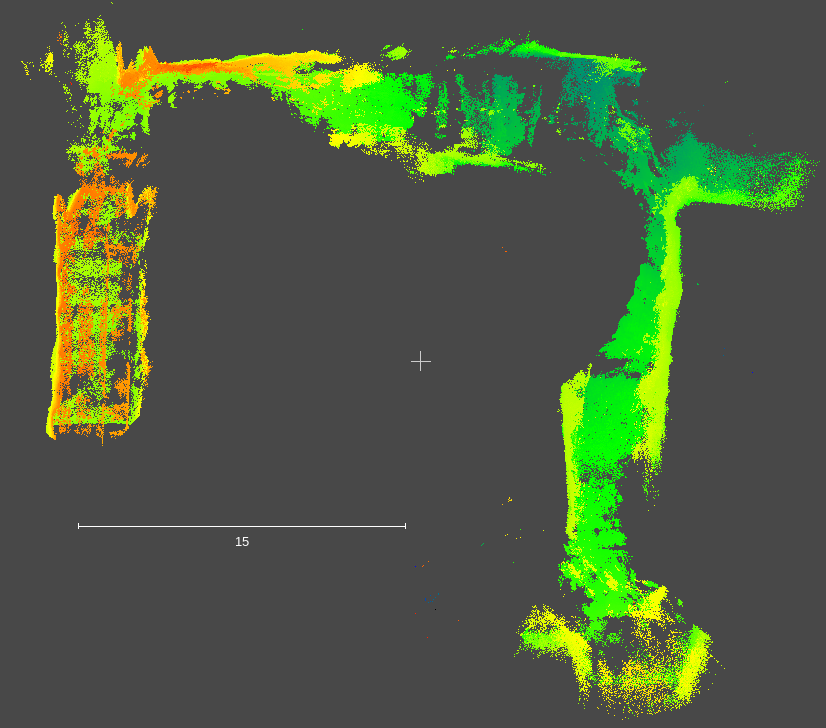}
\caption{VO map colorized by height showing the robot transitioning between different levels in the second mine dataset.}
\label{fig:tunnel_height}
\end{figure}

In \cref{fig:tunnel_height}, we show another result from a different underground environment where the stereo 3D points are colorized by height. The large empty areas in the generated map is due to lack of disparity estimates in large portions of the input images. Due to lack of ground-truth we are unable to assess the accuracy of the system. But, visual inspection of the created 3D maps indicate minimal drift, which is expected when operating in an open loop fashion.

\subsection{Reconstruction density}
\begin{figure}
\centering
\includegraphics[width=0.95\linewidth,keepaspectratio]{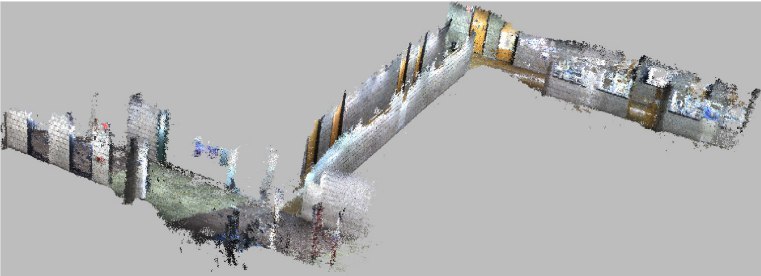}
\includegraphics[width=0.95\linewidth,keepaspectratio]{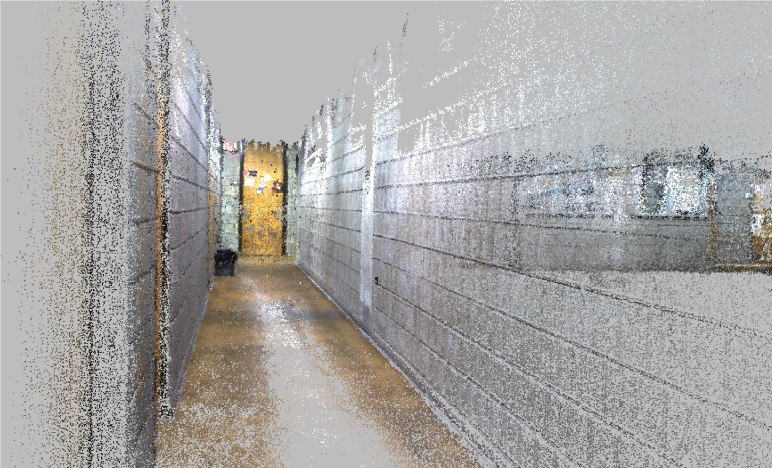}
\caption{Reconstruction density using an indoor dataset collected by us~\cite{alismail10}.}
\label{fig:wean_dense}
\end{figure}

\begin{figure}
\centering
\includegraphics[width=0.95\linewidth]{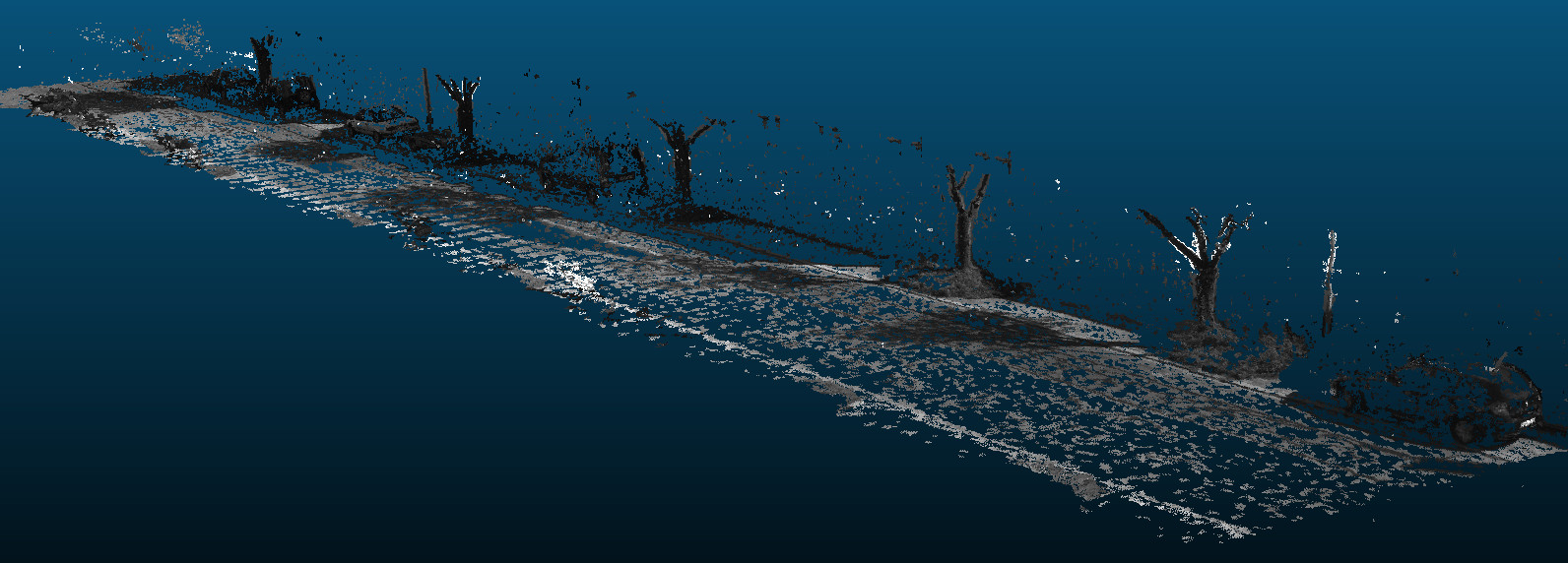}
\caption{Reconstruction density on a section of the KITTI dataset.}
\label{fig:kitti_dense}
\end{figure}

Density of the reconstructed point cloud is demonstrated in \cref{fig:wean_dense} and \cref{fig:kitti_dense}. Denser output is possible by eliminating the pixel selection step at the expense of increased computational time.

\subsection{Failure cases}
\begin{figure}
\centering
\includegraphics[width=0.75\linewidth]{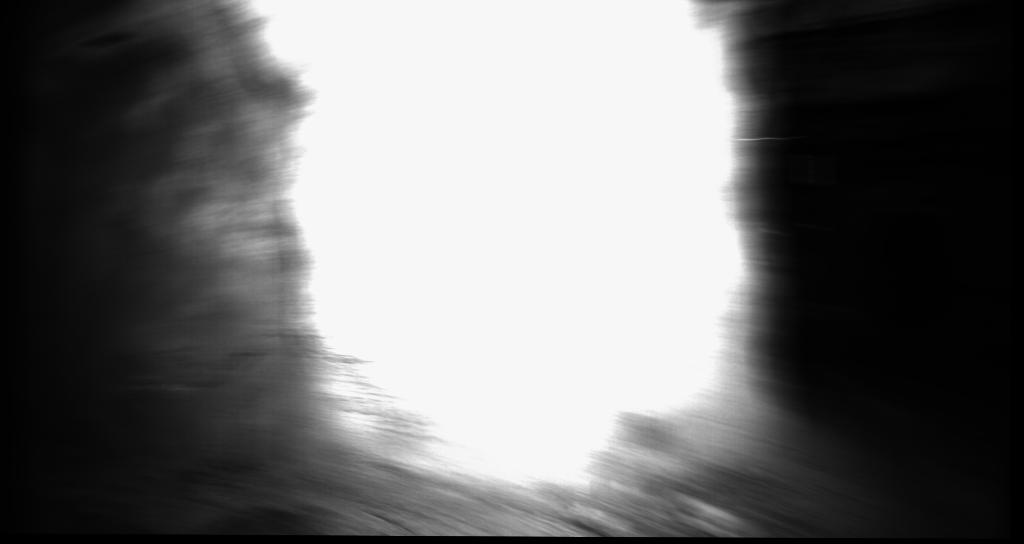}
\caption{An example of a failure case where most of the image details are washed out causing complete loss of stereo.}
\label{fig:failure}
\end{figure}

Most failure cases are due to a complete image washout. An example is shown in \cref{fig:failure}. Theses cases occur when the robot is navigating a tight turn such that all of the LED output is constrained very closely to the camera. Addressing such cases, form vision-only data, is a good avenue of future work.

\section{Conclusion}
\label{sec:conclusion}
In this work, we presented a VO system capable of operating in challenging environments where the illumination of the scene is poor and non-uniform. The approach is based on direct alignment of feature descriptors. In particular, we designed an efficient to compute binary descriptor that is invariant to monotonic changes in intensity. By using this descriptor constancy, we allow vision-only pose estimation to operate robustly in environments that lack keypoints and lack the photometric consistency required by direct methods.

Our descriptor, Bit-Planes, is designed for efficiency. However, other descriptors could be used instead (such ORB and/or SIFT) if computational demands are not an issues. A comparison of performance between difference descriptors in a direct framework is an interesting direction of future work as their amenability to linearization may differ.

The approach is simple to implement, and can be readily integrated into existing direct VSLAM algorithms with a small additional computational overhead.

{
\small\bibliographystyle{plainnat}
\bibliography{bib.bib}
}

\end{document}

%% file: figs/census.tex
\begin{subfigure}[b]{0.25\linewidth}
  \centering
  \resizebox{\textwidth}{!}{\begin{tikzpicture}
      \fill[black!20!white] (0,0) rectangle (1,1);
      \draw[step=1cm,color=black!40!white] (-1,-1) grid (2,2);
      \node at (-0.5,+1.5) {\texttt{8}};
      \node at (+0.5,+1.5) {\texttt{12}};
      \node at (+1.5,+1.5) {\texttt{200}};

      \node at (-0.5,+0.5) {\texttt{56}};
      \node at (+0.5,+0.5) {\texttt{42}};
      \node at (+1.5,+0.5) {\texttt{55}};

      \node at (-0.5,-0.5) {\texttt{128}};
      \node at (+0.5,-0.5) {\texttt{16}};
      \node at (+1.5,-0.5) {\texttt{11}};
  \end{tikzpicture}}
  \caption{}\label{fig:ct1}
\end{subfigure}
\begin{subfigure}[b]{0.25\linewidth}
  \centering
  \resizebox{\textwidth}{!}{\begin{tikzpicture}
      \fill[black!20!white] (0,0) rectangle (1,1);
      \draw[step=1cm,color=black!40!white] (-1,-1) grid (2,2);
      \node at (-0.5,+1.5) {\tiny\texttt{8<42}};
      \node at (+0.5,+1.5) {\tiny\texttt{12<42}};
      \node at (+1.5,+1.5) {\tiny\texttt{200<42}};

      \node at (-0.5,+0.5) {\tiny\texttt{56<42}};
      \node at (+0.5,+0.5) {};
      \node at (+1.5,+0.5) {\tiny\texttt{55<42}};

      \node at (-0.5,-0.5) {\tiny\texttt{128<42}};
      \node at (+0.5,-0.5) {\tiny\texttt{16<42}};
      \node at (+1.5,-0.5) {\tiny\texttt{11<42}};
  \end{tikzpicture}}
  \caption{}\label{fig:ct2}
\end{subfigure}
\begin{subfigure}[b]{0.25\linewidth}
  \centering
  \resizebox{\textwidth}{!}{\begin{tikzpicture}
      \fill[black!20!white] (0,0) rectangle (1,1);
      \draw[step=1cm,color=black!40!white] (-1,-1) grid (2,2);
      \node at (-0.5,+1.5) {\texttt{1}};
      \node at (+0.5,+1.5) {\texttt{1}};
      \node at (+1.5,+1.5) {\texttt{0}};

      \node at (-0.5,+0.5) {\texttt{0}};
      \node at (+0.5,+0.5) {};
      \node at (+1.5,+0.5) {\texttt{0}};

      \node at (-0.5,-0.5) {\texttt{0}};
      \node at (+0.5,-0.5) {\texttt{1}};
      \node at (+1.5,-0.5) {\texttt{1}};
  \end{tikzpicture}}
  \caption{}\label{fig:ct3}
\end{subfigure}

%% file: figs/alg_cmp.tex
%
%
\begin{tikzpicture}

\begin{axis}[%
width=0.951\fwidth,
height=\fheight,
at={(0\fwidth,0\fheight)},
scale only axis,
xmin=0.5,
xmax=6.5,
xtick={1,2,3,4,5,6},
xticklabels={{FA},{FA-1},{FC},{FC-1},{IC},{IC-1}},
xmajorgrids,
ymin=0.003,
ymax=0.034,
ytick={0.003, 0.013, 0.024, 0.034},
ylabel={\footnotesize Point RMS},
ymajorgrids,
axis background/.style={fill=white}
]
\addplot [color=black,dashed,forget plot,line width=1pt]
  table[row sep=crcr]{%
1	0.0219965386492519\\
1	0.0267463805233228\\
};
\addplot [color=black,dashed,forget plot,line width=1pt]
  table[row sep=crcr]{%
2	0.016927533850461\\
2	0.0215901214882594\\
};
\addplot [color=black,dashed,forget plot,line width=1pt]
  table[row sep=crcr]{%
3	0.0219392720132254\\
3	0.0266905952340818\\
};
\addplot [color=black,dashed,forget plot,line width=1pt]
  table[row sep=crcr]{%
4	0.0171894031341911\\
4	0.0214073522841\\
};
\addplot [color=black,dashed,forget plot,line width=1pt]
  table[row sep=crcr]{%
5	0.0218301777680187\\
5	0.0272483709238291\\
};
\addplot [color=black,dashed,forget plot,line width=1pt]
  table[row sep=crcr]{%
6	0.0172426324557945\\
6	0.0201376979001312\\
};
\addplot [color=black,dashed,forget plot,line width=1pt]
  table[row sep=crcr]{%
1	0.0101549521263182\\
1	0.0167785662686705\\
};
\addplot [color=black,dashed,forget plot,line width=1pt]
  table[row sep=crcr]{%
2	0.00477795020616553\\
2	0.00906372711435472\\
};
\addplot [color=black,dashed,forget plot,line width=1pt]
  table[row sep=crcr]{%
3	0.00998399442991195\\
3	0.0167505134902565\\
};
\addplot [color=black,dashed,forget plot,line width=1pt]
  table[row sep=crcr]{%
4	0.004764155222485\\
4	0.00915619016271643\\
};
\addplot [color=black,dashed,forget plot,line width=1pt]
  table[row sep=crcr]{%
5	0.00987462756016741\\
5	0.0167748278249468\\
};
\addplot [color=black,dashed,forget plot,line width=1pt]
  table[row sep=crcr]{%
6	0.00484840061398891\\
6	0.00909445761151399\\
};
\addplot [color=black,solid,forget plot,line width=1pt]
  table[row sep=crcr]{%
0.875	0.0267463805233228\\
1.125	0.0267463805233228\\
};
\addplot [color=black,solid,forget plot,line width=1pt]
  table[row sep=crcr]{%
1.875	0.0215901214882594\\
2.125	0.0215901214882594\\
};
\addplot [color=black,solid,forget plot,line width=1pt]
  table[row sep=crcr]{%
2.875	0.0266905952340818\\
3.125	0.0266905952340818\\
};
\addplot [color=black,solid,forget plot,line width=1pt]
  table[row sep=crcr]{%
3.875	0.0214073522841\\
4.125	0.0214073522841\\
};
\addplot [color=black,solid,forget plot,line width=1pt]
  table[row sep=crcr]{%
4.875	0.0272483709238291\\
5.125	0.0272483709238291\\
};
\addplot [color=black,solid,forget plot,line width=1pt]
  table[row sep=crcr]{%
5.875	0.0201376979001312\\
6.125	0.0201376979001312\\
};
\addplot [color=black,solid,forget plot,line width=1pt]
  table[row sep=crcr]{%
0.875	0.0101549521263182\\
1.125	0.0101549521263182\\
};
\addplot [color=black,solid,forget plot,line width=1pt]
  table[row sep=crcr]{%
1.875	0.00477795020616553\\
2.125	0.00477795020616553\\
};
\addplot [color=black,solid,forget plot,line width=1pt]
  table[row sep=crcr]{%
2.875	0.00998399442991195\\
3.125	0.00998399442991195\\
};
\addplot [color=black,solid,forget plot,line width=1pt]
  table[row sep=crcr]{%
3.875	0.004764155222485\\
4.125	0.004764155222485\\
};
\addplot [color=black,solid,forget plot,line width=1pt]
  table[row sep=crcr]{%
4.875	0.00987462756016741\\
5.125	0.00987462756016741\\
};
\addplot [color=black,solid,forget plot,line width=1pt]
  table[row sep=crcr]{%
5.875	0.00484840061398891\\
6.125	0.00484840061398891\\
};
\addplot [color=black,solid,forget plot,line width=1pt]
  table[row sep=crcr]{%
0.75	0.0167785662686705\\
0.75	0.0219965386492519\\
1.25	0.0219965386492519\\
1.25	0.0167785662686705\\
0.75	0.0167785662686705\\
};
\addplot [color=black,solid,forget plot,line width=1pt]
  table[row sep=crcr]{%
1.75	0.00906372711435472\\
1.75	0.016927533850461\\
2.25	0.016927533850461\\
2.25	0.00906372711435472\\
1.75	0.00906372711435472\\
};
\addplot [color=black,solid,forget plot,line width=1pt]
  table[row sep=crcr]{%
2.75	0.0167505134902565\\
2.75	0.0219392720132254\\
3.25	0.0219392720132254\\
3.25	0.0167505134902565\\
2.75	0.0167505134902565\\
};
\addplot [color=black,solid,forget plot,line width=1pt]
  table[row sep=crcr]{%
3.75	0.00915619016271643\\
3.75	0.0171894031341911\\
4.25	0.0171894031341911\\
4.25	0.00915619016271643\\
3.75	0.00915619016271643\\
};
\addplot [color=black,solid,forget plot,line width=1pt]
  table[row sep=crcr]{%
4.75	0.0167748278249468\\
4.75	0.0218301777680187\\
5.25	0.0218301777680187\\
5.25	0.0167748278249468\\
4.75	0.0167748278249468\\
};
\addplot [color=black,solid,forget plot,line width=1pt]
  table[row sep=crcr]{%
5.75	0.00909445761151399\\
5.75	0.0172426324557945\\
6.25	0.0172426324557945\\
6.25	0.00909445761151399\\
5.75	0.00909445761151399\\
};
\addplot [color=black,solid,forget plot,line width=0.5pt]
  table[row sep=crcr]{%
0.75	0.0192789290561288\\
1.25	0.0192789290561288\\
};
\addplot [color=black,solid,forget plot,line width=0.5pt]
  table[row sep=crcr]{%
1.75	0.0124313460751006\\
2.25	0.0124313460751006\\
};
\addplot [color=black,solid,forget plot,line width=0.5pt]
  table[row sep=crcr]{%
2.75	0.0193351590014576\\
3.25	0.0193351590014576\\
};
\addplot [color=black,solid,forget plot,line width=0.5pt]
  table[row sep=crcr]{%
3.75	0.0123304803730193\\
4.25	0.0123304803730193\\
};
\addplot [color=black,solid,forget plot,line width=0.5pt]
  table[row sep=crcr]{%
4.75	0.0198846631329035\\
5.25	0.0198846631329035\\
};
\addplot [color=black,solid,forget plot,line width=0.5pt]
  table[row sep=crcr]{%
5.75	0.0121303766839266\\
6.25	0.0121303766839266\\
};
\addplot [color=black,only marks,mark=+,mark options={mark size=1.618pt,solid,draw=black,line width=0.5pt},forget plot,line width=1pt]
  table[row sep=crcr]{%
1	0.0298973063996007\\
1	0.030699767734481\\
1	0.031537226638695\\
};
\addplot [color=black,only marks,mark=+,mark options={mark size=1.618pt,solid,draw=black,line width=0.5pt},forget plot,line width=1pt]
  table[row sep=crcr]{%
2	0.0324658482181881\\
};
\addplot [color=black,only marks,mark=+,mark options={mark size=1.618pt,solid,draw=black,line width=0.5pt},forget plot,line width=1pt]
  table[row sep=crcr]{%
3	0.0298197133123936\\
3	0.0306239843196846\\
3	0.0314495936320871\\
};
\addplot [color=black,only marks,mark=+,mark options={mark size=1.618pt,solid,draw=black,line width=0.5pt},forget plot,line width=1pt]
  table[row sep=crcr]{%
4	0.0325730055171678\\
};
\addplot [color=black,only marks,mark=+,mark options={mark size=1.618pt,solid,draw=black,line width=0.5pt},forget plot,line width=1pt]
  table[row sep=crcr]{%
5	0.0303325900399528\\
5	0.03089606613773\\
5	0.0313123617947183\\
};
\addplot [color=black,only marks,mark=+,mark options={mark size=1.618pt,solid,draw=black,line width=0.5pt},forget plot,line width=1pt]
  table[row sep=crcr]{%
6	0.0327950716427288\\
};
\end{axis}
\end{tikzpicture}%

%% file: figs/sigma.tex
%
%
\begin{tikzpicture}

\begin{axis}[%
width=\fwidth,
height=0.857\fheight,
at={(0\fwidth,0\fheight)},
unit vector ratio = 1 1,
scale only axis,
point meta min=0.121840964167431,
point meta max=21.0328982798014,
xmin=1,
xmax=20,
xtick={1,5,10,15,20},
xticklabels={{0.0},{0.4},{0.9},{1.4},{1.9},{0.50},{0.60},{0.70},{0.80},{0.90},{1.00},{1.10},{1.20},{1.30},{1.40},{1.50},{1.60},{1.70},{1.80},{1.90},{}},
xlabel={$\sigma_\textsc{bp}$},
ymin=1,
ymax=20,
ytick={5,10,15,20},
yticklabels={{0.4},{0.9},{1.4},{1.9},{1.00},{1.10},{1.20},{1.30},{1.40},{1.50},{1.60},{1.70},{1.80},{1.90}},
ylabel={$\sigma_0$},
axis background/.style={fill=white},
colormap={mymap}{[1pt] rgb(0pt)=(0.2081,0.1663,0.5292); rgb(1pt)=(0.211624,0.189781,0.577676); rgb(2pt)=(0.212252,0.213771,0.626971); rgb(3pt)=(0.2081,0.2386,0.677086); rgb(4pt)=(0.195905,0.264457,0.7279); rgb(5pt)=(0.170729,0.291938,0.779248); rgb(6pt)=(0.125271,0.324243,0.830271); rgb(7pt)=(0.0591333,0.359833,0.868333); rgb(8pt)=(0.0116952,0.38751,0.881957); rgb(9pt)=(0.00595714,0.408614,0.882843); rgb(10pt)=(0.0165143,0.4266,0.878633); rgb(11pt)=(0.0328524,0.443043,0.871957); rgb(12pt)=(0.0498143,0.458571,0.864057); rgb(13pt)=(0.0629333,0.47369,0.855438); rgb(14pt)=(0.0722667,0.488667,0.8467); rgb(15pt)=(0.0779429,0.503986,0.838371); rgb(16pt)=(0.0793476,0.520024,0.831181); rgb(17pt)=(0.0749429,0.537543,0.826271); rgb(18pt)=(0.0640571,0.556986,0.823957); rgb(19pt)=(0.0487714,0.577224,0.822829); rgb(20pt)=(0.0343429,0.596581,0.819852); rgb(21pt)=(0.0265,0.6137,0.8135); rgb(22pt)=(0.0238905,0.628662,0.803762); rgb(23pt)=(0.0230905,0.641786,0.791267); rgb(24pt)=(0.0227714,0.653486,0.776757); rgb(25pt)=(0.0266619,0.664195,0.760719); rgb(26pt)=(0.0383714,0.674271,0.743552); rgb(27pt)=(0.0589714,0.683757,0.725386); rgb(28pt)=(0.0843,0.692833,0.706167); rgb(29pt)=(0.113295,0.7015,0.685857); rgb(30pt)=(0.145271,0.709757,0.664629); rgb(31pt)=(0.180133,0.717657,0.642433); rgb(32pt)=(0.217829,0.725043,0.619262); rgb(33pt)=(0.258643,0.731714,0.595429); rgb(34pt)=(0.302171,0.737605,0.571186); rgb(35pt)=(0.348167,0.742433,0.547267); rgb(36pt)=(0.395257,0.7459,0.524443); rgb(37pt)=(0.44201,0.748081,0.503314); rgb(38pt)=(0.487124,0.749062,0.483976); rgb(39pt)=(0.530029,0.749114,0.466114); rgb(40pt)=(0.570857,0.748519,0.44939); rgb(41pt)=(0.609852,0.747314,0.433686); rgb(42pt)=(0.6473,0.7456,0.4188); rgb(43pt)=(0.683419,0.743476,0.404433); rgb(44pt)=(0.71841,0.741133,0.390476); rgb(45pt)=(0.752486,0.7384,0.376814); rgb(46pt)=(0.785843,0.735567,0.363271); rgb(47pt)=(0.818505,0.732733,0.34979); rgb(48pt)=(0.850657,0.7299,0.336029); rgb(49pt)=(0.882433,0.727433,0.3217); rgb(50pt)=(0.913933,0.725786,0.306276); rgb(51pt)=(0.944957,0.726114,0.288643); rgb(52pt)=(0.973895,0.731395,0.266648); rgb(53pt)=(0.993771,0.745457,0.240348); rgb(54pt)=(0.999043,0.765314,0.216414); rgb(55pt)=(0.995533,0.786057,0.196652); rgb(56pt)=(0.988,0.8066,0.179367); rgb(57pt)=(0.978857,0.827143,0.163314); rgb(58pt)=(0.9697,0.848138,0.147452); rgb(59pt)=(0.962586,0.870514,0.1309); rgb(60pt)=(0.958871,0.8949,0.113243); rgb(61pt)=(0.959824,0.921833,0.0948381); rgb(62pt)=(0.9661,0.951443,0.0755333); rgb(63pt)=(0.9763,0.9831,0.0538)},
colorbar
]

\addplot[%
surf,
shader=flat corner,draw=black,colormap={mymap}{[1pt] rgb(0pt)=(0.2081,0.1663,0.5292); rgb(1pt)=(0.211624,0.189781,0.577676); rgb(2pt)=(0.212252,0.213771,0.626971); rgb(3pt)=(0.2081,0.2386,0.677086); rgb(4pt)=(0.195905,0.264457,0.7279); rgb(5pt)=(0.170729,0.291938,0.779248); rgb(6pt)=(0.125271,0.324243,0.830271); rgb(7pt)=(0.0591333,0.359833,0.868333); rgb(8pt)=(0.0116952,0.38751,0.881957); rgb(9pt)=(0.00595714,0.408614,0.882843); rgb(10pt)=(0.0165143,0.4266,0.878633); rgb(11pt)=(0.0328524,0.443043,0.871957); rgb(12pt)=(0.0498143,0.458571,0.864057); rgb(13pt)=(0.0629333,0.47369,0.855438); rgb(14pt)=(0.0722667,0.488667,0.8467); rgb(15pt)=(0.0779429,0.503986,0.838371); rgb(16pt)=(0.0793476,0.520024,0.831181); rgb(17pt)=(0.0749429,0.537543,0.826271); rgb(18pt)=(0.0640571,0.556986,0.823957); rgb(19pt)=(0.0487714,0.577224,0.822829); rgb(20pt)=(0.0343429,0.596581,0.819852); rgb(21pt)=(0.0265,0.6137,0.8135); rgb(22pt)=(0.0238905,0.628662,0.803762); rgb(23pt)=(0.0230905,0.641786,0.791267); rgb(24pt)=(0.0227714,0.653486,0.776757); rgb(25pt)=(0.0266619,0.664195,0.760719); rgb(26pt)=(0.0383714,0.674271,0.743552); rgb(27pt)=(0.0589714,0.683757,0.725386); rgb(28pt)=(0.0843,0.692833,0.706167); rgb(29pt)=(0.113295,0.7015,0.685857); rgb(30pt)=(0.145271,0.709757,0.664629); rgb(31pt)=(0.180133,0.717657,0.642433); rgb(32pt)=(0.217829,0.725043,0.619262); rgb(33pt)=(0.258643,0.731714,0.595429); rgb(34pt)=(0.302171,0.737605,0.571186); rgb(35pt)=(0.348167,0.742433,0.547267); rgb(36pt)=(0.395257,0.7459,0.524443); rgb(37pt)=(0.44201,0.748081,0.503314); rgb(38pt)=(0.487124,0.749062,0.483976); rgb(39pt)=(0.530029,0.749114,0.466114); rgb(40pt)=(0.570857,0.748519,0.44939); rgb(41pt)=(0.609852,0.747314,0.433686); rgb(42pt)=(0.6473,0.7456,0.4188); rgb(43pt)=(0.683419,0.743476,0.404433); rgb(44pt)=(0.71841,0.741133,0.390476); rgb(45pt)=(0.752486,0.7384,0.376814); rgb(46pt)=(0.785843,0.735567,0.363271); rgb(47pt)=(0.818505,0.732733,0.34979); rgb(48pt)=(0.850657,0.7299,0.336029); rgb(49pt)=(0.882433,0.727433,0.3217); rgb(50pt)=(0.913933,0.725786,0.306276); rgb(51pt)=(0.944957,0.726114,0.288643); rgb(52pt)=(0.973895,0.731395,0.266648); rgb(53pt)=(0.993771,0.745457,0.240348); rgb(54pt)=(0.999043,0.765314,0.216414); rgb(55pt)=(0.995533,0.786057,0.196652); rgb(56pt)=(0.988,0.8066,0.179367); rgb(57pt)=(0.978857,0.827143,0.163314); rgb(58pt)=(0.9697,0.848138,0.147452); rgb(59pt)=(0.962586,0.870514,0.1309); rgb(60pt)=(0.958871,0.8949,0.113243); rgb(61pt)=(0.959824,0.921833,0.0948381); rgb(62pt)=(0.9661,0.951443,0.0755333); rgb(63pt)=(0.9763,0.9831,0.0538)},mesh/rows=20]
table[row sep=crcr, point meta=\thisrow{c}] {%
x	y	c\\
1	1	0.601773585526416\\
1	2	0.601773585526416\\
1	3	0.533079782598099\\
1	4	0.533079782598099\\
1	5	0.407967827747521\\
1	6	0.606077607215727\\
1	7	0.801258654548371\\
1	8	1.13463944355154\\
1	9	1.68959051116671\\
1	10	2.30710616965786\\
1	11	3.18695951530515\\
1	12	4.18290571798534\\
1	13	5.37431357985142\\
1	14	6.8412316481041\\
1	15	8.46883368076472\\
1	16	10.6123952480867\\
1	17	12.868114179531\\
1	18	14.6185557313362\\
1	19	16.960320214738\\
1	20	21.0328982798014\\
2	1	0.601773585526416\\
2	2	0.601773585526416\\
2	3	0.533079782598099\\
2	4	0.533079782598099\\
2	5	0.407967827747521\\
2	6	0.606077607215727\\
2	7	0.801258654548371\\
2	8	1.13463944355154\\
2	9	1.68959051116671\\
2	10	2.30710616965786\\
2	11	3.18695951530515\\
2	12	4.18290571798534\\
2	13	5.37431357985142\\
2	14	6.8412316481041\\
2	15	8.46883368076472\\
2	16	10.6123952480867\\
2	17	12.868114179531\\
2	18	14.6185557313362\\
2	19	16.960320214738\\
2	20	21.0328982798014\\
3	1	0.533860178890239\\
3	2	0.533860178890239\\
3	3	0.270368595663499\\
3	4	0.270368595663499\\
3	5	0.379393832761171\\
3	6	0.313163091024908\\
3	7	0.431739939935912\\
3	8	0.629292069310066\\
3	9	0.947668659549938\\
3	10	1.2779522311052\\
3	11	1.73686401054901\\
3	12	2.23238595580373\\
3	13	2.79994480570582\\
3	14	3.51829548857543\\
3	15	4.3354061499731\\
3	16	5.42159855124547\\
3	17	6.60043741354323\\
3	18	7.52744326280558\\
3	19	8.80963013318852\\
3	20	11.1246105135035\\
4	1	0.533860178890239\\
4	2	0.533860178890239\\
4	3	0.270368595663499\\
4	4	0.270368595663499\\
4	5	0.379393832761171\\
4	6	0.313163091024908\\
4	7	0.431739939935912\\
4	8	0.629292069310066\\
4	9	0.947668659549938\\
4	10	1.2779522311052\\
4	11	1.73686401054901\\
4	12	2.23238595580373\\
4	13	2.79994480570582\\
4	14	3.51829548857543\\
4	15	4.3354061499731\\
4	16	5.42159855124547\\
4	17	6.60043741354323\\
4	18	7.52744326280558\\
4	19	8.80963013318852\\
4	20	11.1246105135035\\
5	1	1.69054805910887\\
5	2	1.69054805910887\\
5	3	1.18364914450856\\
5	4	1.18364914450856\\
5	5	1.55883207176508\\
5	6	1.13650743320553\\
5	7	1.01980416502451\\
5	8	0.822361925944476\\
5	9	0.541402734579495\\
5	10	0.168172450998388\\
5	11	0.265742011174323\\
5	12	0.799055776866712\\
5	13	1.71077576664384\\
5	14	2.63637690667379\\
5	15	3.29442762229347\\
5	16	5.13117393332546\\
5	17	6.66005206335964\\
5	18	7.24317748207386\\
5	19	8.83633072680545\\
5	20	12.4888839539142\\
6	1	1.36750943929596\\
6	2	1.36750943929596\\
6	3	1.20390950282885\\
6	4	1.20390950282885\\
6	5	1.31106558045724\\
6	6	1.19078859839176\\
6	7	1.1585688775675\\
6	8	1.08294196789308\\
6	9	1.01089126349599\\
6	10	0.893859795921433\\
6	11	0.764845122287839\\
6	12	0.564806606304019\\
6	13	0.121840964167431\\
6	14	0.329677188401901\\
6	15	0.601436554156844\\
6	16	1.64714481140419\\
6	17	2.49775928524664\\
6	18	2.75506313956994\\
6	19	3.6726171209789\\
6	20	5.93901038202504\\
7	1	1.3416722683286\\
7	2	1.3416722683286\\
7	3	1.11825221212288\\
7	4	1.11825221212288\\
7	5	1.27368985459572\\
7	6	1.09832414191469\\
7	7	1.04986703754269\\
7	8	0.955897774847504\\
7	9	0.849196204010554\\
7	10	0.694222321257353\\
7	11	0.539381814942474\\
7	12	0.329801123609627\\
7	13	0.223601200311911\\
7	14	0.651106387317377\\
7	15	0.956341049849701\\
7	16	1.93627839252864\\
7	17	2.74633058725032\\
7	18	3.00238475900909\\
7	19	3.85236272275462\\
7	20	5.93702614627948\\
8	1	2.31997395291494\\
8	2	2.31997395291494\\
8	3	2.29092636446042\\
8	4	2.29092636446042\\
8	5	2.32265213082581\\
8	6	2.29645492058584\\
8	7	2.31088282290949\\
8	8	2.4923448221352\\
8	9	2.57820931938359\\
8	10	2.4112533332764\\
8	11	2.75670508994784\\
8	12	2.8394282014601\\
8	13	2.51725591969636\\
8	14	2.54795832660415\\
8	15	3.01039482837892\\
8	16	2.64263905780709\\
8	17	2.74998861460287\\
8	18	3.12845995409227\\
8	19	3.19989530177364\\
8	20	3.60104559537409\\
9	1	2.17307348430414\\
9	2	2.17307348430414\\
9	3	2.13658173898383\\
9	4	2.13658173898383\\
9	5	2.17980307096211\\
9	6	2.13843077133505\\
9	7	2.14366180355511\\
9	8	2.30491440118859\\
9	9	2.36381761405017\\
9	10	2.18910885143077\\
9	11	2.48442821385301\\
9	12	2.54170461762422\\
9	13	2.2668296380186\\
9	14	2.31253903362726\\
9	15	2.69165847699164\\
9	16	2.50193603448319\\
9	17	2.68232315411635\\
9	18	2.93865194019698\\
9	19	3.0850821545393\\
9	20	3.75030227917835\\
10	1	1.65728446978543\\
10	2	1.65728446978543\\
10	3	1.76303553183078\\
10	4	1.76303553183078\\
10	5	1.70262971198072\\
10	6	1.77912239169264\\
10	7	1.81840645780727\\
10	8	2.05165310875066\\
10	9	2.18252851069821\\
10	10	2.05619865375418\\
10	11	2.45749186314725\\
10	12	2.5984356172298\\
10	13	2.37121146569023\\
10	14	2.49178705402405\\
10	15	3.02749402855006\\
10	16	2.78095694098746\\
10	17	2.96364621218473\\
10	18	3.51378325683238\\
10	19	3.70276362175427\\
10	20	3.77137664099446\\
11	1	2.75976944409825\\
11	2	2.75976944409825\\
11	3	2.84559000863166\\
11	4	2.84559000863166\\
11	5	2.8467069643833\\
11	6	2.87790102264746\\
11	7	2.95395165144263\\
11	8	3.30097079118994\\
11	9	3.45001548087999\\
11	10	3.36898437945999\\
11	11	3.77548101507806\\
11	12	3.94001726074963\\
11	13	3.83662779527997\\
11	14	3.97804478743337\\
11	15	4.39857210799578\\
11	16	4.22372622598582\\
11	17	4.32276020449745\\
11	18	4.75256483916209\\
11	19	4.84611786848128\\
11	20	4.55819697110772\\
12	1	2.58903461442124\\
12	2	2.58903461442124\\
12	3	2.66115617528683\\
12	4	2.66115617528683\\
12	5	2.67469478338653\\
12	6	2.69240763203373\\
12	7	2.76579932785472\\
12	8	3.11287699888985\\
12	9	3.25372971033523\\
12	10	3.16503130564968\\
12	11	3.56321659870304\\
12	12	3.72286180015378\\
12	13	3.63751725862959\\
12	14	3.79316644489562\\
12	15	4.19817375730404\\
12	16	4.10082683402548\\
12	17	4.2471203294519\\
12	18	4.64247246910203\\
12	19	4.78399394232682\\
12	20	4.70742823197633\\
13	1	2.58165607578721\\
13	2	2.58165607578721\\
13	3	2.70866487091115\\
13	4	2.70866487091115\\
13	5	2.69808535742373\\
13	6	2.75298795518748\\
13	7	2.85696558279117\\
13	8	3.31264007603332\\
13	9	3.50333033627963\\
13	10	3.41185924794157\\
13	11	3.903841590938\\
13	12	4.10480466692284\\
13	13	4.04337662191087\\
13	14	4.25151731743098\\
13	15	4.70979200108372\\
13	16	4.67759036619891\\
13	17	4.88906402700378\\
13	18	5.32843423138878\\
13	19	5.54138083612411\\
13	20	5.57631807243666\\
14	1	2.5178572117787\\
14	2	2.5178572117787\\
14	3	2.62783532630095\\
14	4	2.62783532630095\\
14	5	2.63773027429268\\
14	6	2.67217918776684\\
14	7	2.77557245303015\\
14	8	3.24093747815155\\
14	9	3.42121967900364\\
14	10	3.31521600331682\\
14	11	3.79351973699936\\
14	12	3.97779136200874\\
14	13	3.91686532314659\\
14	14	4.11465814087629\\
14	15	4.52720428890796\\
14	16	4.52039232894111\\
14	17	4.72209365060298\\
14	18	5.08504383872518\\
14	19	5.27618801105823\\
14	20	5.37608971417941\\
15	1	2.62562563651661\\
15	2	2.62562563651661\\
15	3	3.2421182293775\\
15	4	3.2421182293775\\
15	5	2.74491449194064\\
15	6	3.28422406184786\\
15	7	3.38068078786127\\
15	8	3.39748741027203\\
15	9	3.58637952095262\\
15	10	3.85705466051973\\
15	11	3.95338933620429\\
15	12	4.12791731179042\\
15	13	4.35624781814799\\
15	14	4.51528091750094\\
15	15	4.64175406123134\\
15	16	4.84388653104458\\
15	17	5.00533073347947\\
15	18	5.17345197616106\\
15	19	5.36071925857199\\
15	20	5.5389100091966\\
16	1	2.20261403279602\\
16	2	2.20261403279602\\
16	3	2.90779585313601\\
16	4	2.90779585313601\\
16	5	2.33377843537093\\
16	6	2.95651612168225\\
16	7	3.06842033275543\\
16	8	3.04743004159365\\
16	9	3.25327068565321\\
16	10	3.62877889805167\\
16	11	3.6605334062471\\
16	12	3.8597668248202\\
16	13	4.24711499448762\\
16	14	4.45602773592831\\
16	15	4.48506979295153\\
16	16	4.91115597583485\\
16	17	5.14457632479317\\
16	18	5.18873910493074\\
16	19	5.44298952528067\\
16	20	5.92998786041943\\
17	1	2.01669054365314\\
17	2	2.01669054365314\\
17	3	2.76583979206137\\
17	4	2.76583979206137\\
17	5	2.16022353574649\\
17	6	2.81778167244898\\
17	7	2.93694004372027\\
17	8	2.91948514938002\\
17	9	3.133893815896\\
17	10	3.53138269258424\\
17	11	3.55712532602484\\
17	12	3.76399446607698\\
17	13	4.18720646098186\\
17	14	4.40906044714938\\
17	15	4.41556694621668\\
17	16	4.89379542885315\\
17	17	5.14196328885066\\
17	18	5.1446689901232\\
17	19	5.40522618020373\\
17	20	5.97347342839754\\
18	1	2.12352924683733\\
18	2	2.12352924683733\\
18	3	2.77404346212151\\
18	4	2.77404346212151\\
18	5	2.2546385196593\\
18	6	2.82359730038322\\
18	7	2.93825842804973\\
18	8	2.96957787945815\\
18	9	3.17545866789658\\
18	10	3.52909021695137\\
18	11	3.586307187151\\
18	12	3.79042637855058\\
18	13	4.1961096822341\\
18	14	4.42057440052575\\
18	15	4.43500444563092\\
18	16	4.89920932531345\\
18	17	5.14174952042666\\
18	18	5.151495078737\\
18	19	5.40757117481277\\
18	20	5.93851372911009\\
19	1	2.20867952624761\\
19	2	2.20867952624761\\
19	3	2.96220324110918\\
19	4	2.96220324110918\\
19	5	2.35840177618675\\
19	6	3.01593976485171\\
19	7	3.14004152552415\\
19	8	3.1522302207137\\
19	9	3.37574300898759\\
19	10	3.77680976261856\\
19	11	3.82034438301006\\
19	12	4.04093563411991\\
19	13	4.49710307478758\\
19	14	4.74019018079368\\
19	15	4.73961692531442\\
19	16	5.26191092557217\\
19	17	5.5258122349052\\
19	18	5.5108446896824\\
19	19	5.78278720848181\\
19	20	6.3890875575806\\
20	1	1.67627148413229\\
20	2	1.67627148413229\\
20	3	2.60303152722318\\
20	4	2.60303152722318\\
20	5	1.84332900092578\\
20	6	2.66379767549772\\
20	7	2.80320037528387\\
20	8	2.72754037100851\\
20	9	2.97799868180199\\
20	10	3.51160282139681\\
20	11	3.4830046277025\\
20	12	3.73613287026731\\
20	13	4.324152166954\\
20	14	4.60323262298045\\
20	15	4.55680549547018\\
20	16	5.21428314776456\\
20	17	5.52636997781283\\
20	18	5.48404488366894\\
20	19	5.81257416777109\\
20	20	6.5589797828443\\
};
\addplot [color=red,only marks,mark=square,mark options={solid},forget plot]
  table[row sep=crcr]{%
6	13\\
};
\addplot [color=red,only marks,mark=square,mark options={solid},forget plot]
  table[row sep=crcr]{%
5	10\\
};
\addplot [color=red,only marks,mark=square,mark options={solid},forget plot]
  table[row sep=crcr]{%
7	13\\
};
\addplot [color=red,only marks,mark=square,mark options={solid},forget plot]
  table[row sep=crcr]{%
5	11\\
};
\addplot [color=red,only marks,mark=square,mark options={solid},forget plot]
  table[row sep=crcr]{%
3	3\\
};
\end{axis}
\end{tikzpicture}%

%% file: figs/tsukuba_intensity_9_iters.tex
%
%
\begin{tikzpicture}

\begin{axis}[%
width=0.951\fwidth,
height=\fheight,
at={(0\fwidth,0\fheight)},
scale only axis,
xmin=1,
xmax=501,
xlabel={Frame number},
xmajorgrids,
ymin=0,
ymax=34,
ylabel={\# iterations},
ymajorgrids,
axis background/.style={fill=white},
title style={font=\bfseries},
title={Raw Intensity},
legend style={at={(0.03,0.97)},anchor=north west,legend cell align=left,align=left,draw=white!15!black}
]
\addplot [color=white!30!black,solid,mark=+,mark options={solid}]
  table[row sep=crcr]{%
1	0\\
2	4\\
3	5\\
4	6\\
5	5\\
6	5\\
7	4\\
8	5\\
9	6\\
10	4\\
11	5\\
12	6\\
13	6\\
14	6\\
15	6\\
16	6\\
17	6\\
18	6\\
19	6\\
20	6\\
21	6\\
22	6\\
23	5\\
24	5\\
25	7\\
26	6\\
27	7\\
28	5\\
29	5\\
30	7\\
31	6\\
32	5\\
33	6\\
34	5\\
35	5\\
36	6\\
37	7\\
38	6\\
39	5\\
40	7\\
41	7\\
42	6\\
43	5\\
44	6\\
45	6\\
46	5\\
47	6\\
48	5\\
49	6\\
50	5\\
51	5\\
52	6\\
53	6\\
54	5\\
55	5\\
56	7\\
57	4\\
58	4\\
59	5\\
60	6\\
61	5\\
62	7\\
63	6\\
64	5\\
65	6\\
66	6\\
67	7\\
68	8\\
69	6\\
70	6\\
71	5\\
72	6\\
73	4\\
74	4\\
75	5\\
76	4\\
77	5\\
78	4\\
79	4\\
80	4\\
81	5\\
82	5\\
83	5\\
84	5\\
85	5\\
86	6\\
87	5\\
88	6\\
89	5\\
90	5\\
91	5\\
92	5\\
93	5\\
94	5\\
95	5\\
96	7\\
97	5\\
98	5\\
99	5\\
100	5\\
101	4\\
102	5\\
103	7\\
104	6\\
105	6\\
106	5\\
107	5\\
108	5\\
109	5\\
110	6\\
111	5\\
112	5\\
113	6\\
114	7\\
115	9\\
116	8\\
117	12\\
118	12\\
119	7\\
120	10\\
121	7\\
122	8\\
123	13\\
124	15\\
125	6\\
126	19\\
127	6\\
128	7\\
129	8\\
130	15\\
131	29\\
132	5\\
133	10\\
134	27\\
135	5\\
136	13\\
137	17\\
138	17\\
139	17\\
140	26\\
141	17\\
142	18\\
143	19\\
144	19\\
145	18\\
146	22\\
147	13\\
148	11\\
149	16\\
150	17\\
151	16\\
152	13\\
153	10\\
154	12\\
155	8\\
156	12\\
157	13\\
158	10\\
159	7\\
160	7\\
161	9\\
162	9\\
163	8\\
164	7\\
165	7\\
166	6\\
167	6\\
168	4\\
169	5\\
170	6\\
171	8\\
172	9\\
173	7\\
174	8\\
175	6\\
176	10\\
177	28\\
178	17\\
179	17\\
180	25\\
181	34\\
182	25\\
183	20\\
184	17\\
185	15\\
186	16\\
187	10\\
188	18\\
189	18\\
190	14\\
191	17\\
192	15\\
193	17\\
194	12\\
195	14\\
196	11\\
197	13\\
198	18\\
199	6\\
200	9\\
201	8\\
202	9\\
203	9\\
204	15\\
205	7\\
206	7\\
207	6\\
208	9\\
209	8\\
210	10\\
211	8\\
212	8\\
213	8\\
214	9\\
215	15\\
216	10\\
217	8\\
218	9\\
219	10\\
220	9\\
221	8\\
222	11\\
223	8\\
224	6\\
225	7\\
226	9\\
227	8\\
228	12\\
229	7\\
230	7\\
231	7\\
232	8\\
233	9\\
234	8\\
235	6\\
236	7\\
237	8\\
238	9\\
239	8\\
240	12\\
241	14\\
242	10\\
243	10\\
244	12\\
245	11\\
246	16\\
247	17\\
248	18\\
249	22\\
250	16\\
251	19\\
252	27\\
253	13\\
254	13\\
255	6\\
256	11\\
257	9\\
258	11\\
259	6\\
260	8\\
261	8\\
262	7\\
263	7\\
264	10\\
265	9\\
266	8\\
267	8\\
268	9\\
269	9\\
270	9\\
271	10\\
272	6\\
273	5\\
274	7\\
275	8\\
276	7\\
277	7\\
278	7\\
279	7\\
280	8\\
281	8\\
282	10\\
283	12\\
284	6\\
285	5\\
286	7\\
287	7\\
288	6\\
289	6\\
290	8\\
291	8\\
292	9\\
293	11\\
294	5\\
295	9\\
296	8\\
297	8\\
298	5\\
299	11\\
300	10\\
301	12\\
302	5\\
303	8\\
304	7\\
305	6\\
306	7\\
307	6\\
308	5\\
309	6\\
310	5\\
311	5\\
312	6\\
313	6\\
314	7\\
315	6\\
316	6\\
317	6\\
318	6\\
319	6\\
320	7\\
321	6\\
322	6\\
323	5\\
324	6\\
325	6\\
326	6\\
327	6\\
328	5\\
329	4\\
330	5\\
331	5\\
332	6\\
333	6\\
334	5\\
335	7\\
336	4\\
337	4\\
338	4\\
339	5\\
340	5\\
341	5\\
342	6\\
343	5\\
344	6\\
345	6\\
346	6\\
347	7\\
348	5\\
349	4\\
350	5\\
351	5\\
352	5\\
353	4\\
354	5\\
355	4\\
356	6\\
357	7\\
358	4\\
359	6\\
360	5\\
361	6\\
362	7\\
363	5\\
364	7\\
365	5\\
366	7\\
367	5\\
368	7\\
369	7\\
370	5\\
371	5\\
372	7\\
373	7\\
374	8\\
375	9\\
376	6\\
377	5\\
378	7\\
379	5\\
380	6\\
381	6\\
382	5\\
383	5\\
384	5\\
385	7\\
386	5\\
387	9\\
388	8\\
389	9\\
390	13\\
391	8\\
392	9\\
393	14\\
394	4\\
395	8\\
396	7\\
397	9\\
398	6\\
399	7\\
400	9\\
401	11\\
402	5\\
403	6\\
404	7\\
405	8\\
406	6\\
407	5\\
408	7\\
409	8\\
410	5\\
411	6\\
412	8\\
413	8\\
414	5\\
415	7\\
416	7\\
417	10\\
418	7\\
419	6\\
420	7\\
421	7\\
422	7\\
423	7\\
424	5\\
425	7\\
426	4\\
427	6\\
428	7\\
429	7\\
430	5\\
431	5\\
432	5\\
433	6\\
434	4\\
435	6\\
436	7\\
437	8\\
438	6\\
439	6\\
440	5\\
441	6\\
442	5\\
443	6\\
444	7\\
445	8\\
446	5\\
447	7\\
448	7\\
449	5\\
450	5\\
451	7\\
452	7\\
453	8\\
454	5\\
455	6\\
456	6\\
457	7\\
458	5\\
459	5\\
460	6\\
461	6\\
462	5\\
463	5\\
464	6\\
465	6\\
466	5\\
467	5\\
468	5\\
469	5\\
470	5\\
471	7\\
472	7\\
473	8\\
474	6\\
475	7\\
476	7\\
477	8\\
478	8\\
479	7\\
480	6\\
481	7\\
482	6\\
483	5\\
484	5\\
485	6\\
486	5\\
487	5\\
488	7\\
489	7\\
490	5\\
491	6\\
492	6\\
493	7\\
494	6\\
495	6\\
496	7\\
497	8\\
498	6\\
499	5\\
500	5\\
501	5\\
};
\addlegendentry{data};

\addplot [color=black,dashed,line width=2.0pt]
  table[row sep=crcr]{%
1	7.83632734530938\\
2	7.83632734530938\\
3	7.83632734530938\\
4	7.83632734530938\\
5	7.83632734530938\\
6	7.83632734530938\\
7	7.83632734530938\\
8	7.83632734530938\\
9	7.83632734530938\\
10	7.83632734530938\\
11	7.83632734530938\\
12	7.83632734530938\\
13	7.83632734530938\\
14	7.83632734530938\\
15	7.83632734530938\\
16	7.83632734530938\\
17	7.83632734530938\\
18	7.83632734530938\\
19	7.83632734530938\\
20	7.83632734530938\\
21	7.83632734530938\\
22	7.83632734530938\\
23	7.83632734530938\\
24	7.83632734530938\\
25	7.83632734530938\\
26	7.83632734530938\\
27	7.83632734530938\\
28	7.83632734530938\\
29	7.83632734530938\\
30	7.83632734530938\\
31	7.83632734530938\\
32	7.83632734530938\\
33	7.83632734530938\\
34	7.83632734530938\\
35	7.83632734530938\\
36	7.83632734530938\\
37	7.83632734530938\\
38	7.83632734530938\\
39	7.83632734530938\\
40	7.83632734530938\\
41	7.83632734530938\\
42	7.83632734530938\\
43	7.83632734530938\\
44	7.83632734530938\\
45	7.83632734530938\\
46	7.83632734530938\\
47	7.83632734530938\\
48	7.83632734530938\\
49	7.83632734530938\\
50	7.83632734530938\\
51	7.83632734530938\\
52	7.83632734530938\\
53	7.83632734530938\\
54	7.83632734530938\\
55	7.83632734530938\\
56	7.83632734530938\\
57	7.83632734530938\\
58	7.83632734530938\\
59	7.83632734530938\\
60	7.83632734530938\\
61	7.83632734530938\\
62	7.83632734530938\\
63	7.83632734530938\\
64	7.83632734530938\\
65	7.83632734530938\\
66	7.83632734530938\\
67	7.83632734530938\\
68	7.83632734530938\\
69	7.83632734530938\\
70	7.83632734530938\\
71	7.83632734530938\\
72	7.83632734530938\\
73	7.83632734530938\\
74	7.83632734530938\\
75	7.83632734530938\\
76	7.83632734530938\\
77	7.83632734530938\\
78	7.83632734530938\\
79	7.83632734530938\\
80	7.83632734530938\\
81	7.83632734530938\\
82	7.83632734530938\\
83	7.83632734530938\\
84	7.83632734530938\\
85	7.83632734530938\\
86	7.83632734530938\\
87	7.83632734530938\\
88	7.83632734530938\\
89	7.83632734530938\\
90	7.83632734530938\\
91	7.83632734530938\\
92	7.83632734530938\\
93	7.83632734530938\\
94	7.83632734530938\\
95	7.83632734530938\\
96	7.83632734530938\\
97	7.83632734530938\\
98	7.83632734530938\\
99	7.83632734530938\\
100	7.83632734530938\\
101	7.83632734530938\\
102	7.83632734530938\\
103	7.83632734530938\\
104	7.83632734530938\\
105	7.83632734530938\\
106	7.83632734530938\\
107	7.83632734530938\\
108	7.83632734530938\\
109	7.83632734530938\\
110	7.83632734530938\\
111	7.83632734530938\\
112	7.83632734530938\\
113	7.83632734530938\\
114	7.83632734530938\\
115	7.83632734530938\\
116	7.83632734530938\\
117	7.83632734530938\\
118	7.83632734530938\\
119	7.83632734530938\\
120	7.83632734530938\\
121	7.83632734530938\\
122	7.83632734530938\\
123	7.83632734530938\\
124	7.83632734530938\\
125	7.83632734530938\\
126	7.83632734530938\\
127	7.83632734530938\\
128	7.83632734530938\\
129	7.83632734530938\\
130	7.83632734530938\\
131	7.83632734530938\\
132	7.83632734530938\\
133	7.83632734530938\\
134	7.83632734530938\\
135	7.83632734530938\\
136	7.83632734530938\\
137	7.83632734530938\\
138	7.83632734530938\\
139	7.83632734530938\\
140	7.83632734530938\\
141	7.83632734530938\\
142	7.83632734530938\\
143	7.83632734530938\\
144	7.83632734530938\\
145	7.83632734530938\\
146	7.83632734530938\\
147	7.83632734530938\\
148	7.83632734530938\\
149	7.83632734530938\\
150	7.83632734530938\\
151	7.83632734530938\\
152	7.83632734530938\\
153	7.83632734530938\\
154	7.83632734530938\\
155	7.83632734530938\\
156	7.83632734530938\\
157	7.83632734530938\\
158	7.83632734530938\\
159	7.83632734530938\\
160	7.83632734530938\\
161	7.83632734530938\\
162	7.83632734530938\\
163	7.83632734530938\\
164	7.83632734530938\\
165	7.83632734530938\\
166	7.83632734530938\\
167	7.83632734530938\\
168	7.83632734530938\\
169	7.83632734530938\\
170	7.83632734530938\\
171	7.83632734530938\\
172	7.83632734530938\\
173	7.83632734530938\\
174	7.83632734530938\\
175	7.83632734530938\\
176	7.83632734530938\\
177	7.83632734530938\\
178	7.83632734530938\\
179	7.83632734530938\\
180	7.83632734530938\\
181	7.83632734530938\\
182	7.83632734530938\\
183	7.83632734530938\\
184	7.83632734530938\\
185	7.83632734530938\\
186	7.83632734530938\\
187	7.83632734530938\\
188	7.83632734530938\\
189	7.83632734530938\\
190	7.83632734530938\\
191	7.83632734530938\\
192	7.83632734530938\\
193	7.83632734530938\\
194	7.83632734530938\\
195	7.83632734530938\\
196	7.83632734530938\\
197	7.83632734530938\\
198	7.83632734530938\\
199	7.83632734530938\\
200	7.83632734530938\\
201	7.83632734530938\\
202	7.83632734530938\\
203	7.83632734530938\\
204	7.83632734530938\\
205	7.83632734530938\\
206	7.83632734530938\\
207	7.83632734530938\\
208	7.83632734530938\\
209	7.83632734530938\\
210	7.83632734530938\\
211	7.83632734530938\\
212	7.83632734530938\\
213	7.83632734530938\\
214	7.83632734530938\\
215	7.83632734530938\\
216	7.83632734530938\\
217	7.83632734530938\\
218	7.83632734530938\\
219	7.83632734530938\\
220	7.83632734530938\\
221	7.83632734530938\\
222	7.83632734530938\\
223	7.83632734530938\\
224	7.83632734530938\\
225	7.83632734530938\\
226	7.83632734530938\\
227	7.83632734530938\\
228	7.83632734530938\\
229	7.83632734530938\\
230	7.83632734530938\\
231	7.83632734530938\\
232	7.83632734530938\\
233	7.83632734530938\\
234	7.83632734530938\\
235	7.83632734530938\\
236	7.83632734530938\\
237	7.83632734530938\\
238	7.83632734530938\\
239	7.83632734530938\\
240	7.83632734530938\\
241	7.83632734530938\\
242	7.83632734530938\\
243	7.83632734530938\\
244	7.83632734530938\\
245	7.83632734530938\\
246	7.83632734530938\\
247	7.83632734530938\\
248	7.83632734530938\\
249	7.83632734530938\\
250	7.83632734530938\\
251	7.83632734530938\\
252	7.83632734530938\\
253	7.83632734530938\\
254	7.83632734530938\\
255	7.83632734530938\\
256	7.83632734530938\\
257	7.83632734530938\\
258	7.83632734530938\\
259	7.83632734530938\\
260	7.83632734530938\\
261	7.83632734530938\\
262	7.83632734530938\\
263	7.83632734530938\\
264	7.83632734530938\\
265	7.83632734530938\\
266	7.83632734530938\\
267	7.83632734530938\\
268	7.83632734530938\\
269	7.83632734530938\\
270	7.83632734530938\\
271	7.83632734530938\\
272	7.83632734530938\\
273	7.83632734530938\\
274	7.83632734530938\\
275	7.83632734530938\\
276	7.83632734530938\\
277	7.83632734530938\\
278	7.83632734530938\\
279	7.83632734530938\\
280	7.83632734530938\\
281	7.83632734530938\\
282	7.83632734530938\\
283	7.83632734530938\\
284	7.83632734530938\\
285	7.83632734530938\\
286	7.83632734530938\\
287	7.83632734530938\\
288	7.83632734530938\\
289	7.83632734530938\\
290	7.83632734530938\\
291	7.83632734530938\\
292	7.83632734530938\\
293	7.83632734530938\\
294	7.83632734530938\\
295	7.83632734530938\\
296	7.83632734530938\\
297	7.83632734530938\\
298	7.83632734530938\\
299	7.83632734530938\\
300	7.83632734530938\\
301	7.83632734530938\\
302	7.83632734530938\\
303	7.83632734530938\\
304	7.83632734530938\\
305	7.83632734530938\\
306	7.83632734530938\\
307	7.83632734530938\\
308	7.83632734530938\\
309	7.83632734530938\\
310	7.83632734530938\\
311	7.83632734530938\\
312	7.83632734530938\\
313	7.83632734530938\\
314	7.83632734530938\\
315	7.83632734530938\\
316	7.83632734530938\\
317	7.83632734530938\\
318	7.83632734530938\\
319	7.83632734530938\\
320	7.83632734530938\\
321	7.83632734530938\\
322	7.83632734530938\\
323	7.83632734530938\\
324	7.83632734530938\\
325	7.83632734530938\\
326	7.83632734530938\\
327	7.83632734530938\\
328	7.83632734530938\\
329	7.83632734530938\\
330	7.83632734530938\\
331	7.83632734530938\\
332	7.83632734530938\\
333	7.83632734530938\\
334	7.83632734530938\\
335	7.83632734530938\\
336	7.83632734530938\\
337	7.83632734530938\\
338	7.83632734530938\\
339	7.83632734530938\\
340	7.83632734530938\\
341	7.83632734530938\\
342	7.83632734530938\\
343	7.83632734530938\\
344	7.83632734530938\\
345	7.83632734530938\\
346	7.83632734530938\\
347	7.83632734530938\\
348	7.83632734530938\\
349	7.83632734530938\\
350	7.83632734530938\\
351	7.83632734530938\\
352	7.83632734530938\\
353	7.83632734530938\\
354	7.83632734530938\\
355	7.83632734530938\\
356	7.83632734530938\\
357	7.83632734530938\\
358	7.83632734530938\\
359	7.83632734530938\\
360	7.83632734530938\\
361	7.83632734530938\\
362	7.83632734530938\\
363	7.83632734530938\\
364	7.83632734530938\\
365	7.83632734530938\\
366	7.83632734530938\\
367	7.83632734530938\\
368	7.83632734530938\\
369	7.83632734530938\\
370	7.83632734530938\\
371	7.83632734530938\\
372	7.83632734530938\\
373	7.83632734530938\\
374	7.83632734530938\\
375	7.83632734530938\\
376	7.83632734530938\\
377	7.83632734530938\\
378	7.83632734530938\\
379	7.83632734530938\\
380	7.83632734530938\\
381	7.83632734530938\\
382	7.83632734530938\\
383	7.83632734530938\\
384	7.83632734530938\\
385	7.83632734530938\\
386	7.83632734530938\\
387	7.83632734530938\\
388	7.83632734530938\\
389	7.83632734530938\\
390	7.83632734530938\\
391	7.83632734530938\\
392	7.83632734530938\\
393	7.83632734530938\\
394	7.83632734530938\\
395	7.83632734530938\\
396	7.83632734530938\\
397	7.83632734530938\\
398	7.83632734530938\\
399	7.83632734530938\\
400	7.83632734530938\\
401	7.83632734530938\\
402	7.83632734530938\\
403	7.83632734530938\\
404	7.83632734530938\\
405	7.83632734530938\\
406	7.83632734530938\\
407	7.83632734530938\\
408	7.83632734530938\\
409	7.83632734530938\\
410	7.83632734530938\\
411	7.83632734530938\\
412	7.83632734530938\\
413	7.83632734530938\\
414	7.83632734530938\\
415	7.83632734530938\\
416	7.83632734530938\\
417	7.83632734530938\\
418	7.83632734530938\\
419	7.83632734530938\\
420	7.83632734530938\\
421	7.83632734530938\\
422	7.83632734530938\\
423	7.83632734530938\\
424	7.83632734530938\\
425	7.83632734530938\\
426	7.83632734530938\\
427	7.83632734530938\\
428	7.83632734530938\\
429	7.83632734530938\\
430	7.83632734530938\\
431	7.83632734530938\\
432	7.83632734530938\\
433	7.83632734530938\\
434	7.83632734530938\\
435	7.83632734530938\\
436	7.83632734530938\\
437	7.83632734530938\\
438	7.83632734530938\\
439	7.83632734530938\\
440	7.83632734530938\\
441	7.83632734530938\\
442	7.83632734530938\\
443	7.83632734530938\\
444	7.83632734530938\\
445	7.83632734530938\\
446	7.83632734530938\\
447	7.83632734530938\\
448	7.83632734530938\\
449	7.83632734530938\\
450	7.83632734530938\\
451	7.83632734530938\\
452	7.83632734530938\\
453	7.83632734530938\\
454	7.83632734530938\\
455	7.83632734530938\\
456	7.83632734530938\\
457	7.83632734530938\\
458	7.83632734530938\\
459	7.83632734530938\\
460	7.83632734530938\\
461	7.83632734530938\\
462	7.83632734530938\\
463	7.83632734530938\\
464	7.83632734530938\\
465	7.83632734530938\\
466	7.83632734530938\\
467	7.83632734530938\\
468	7.83632734530938\\
469	7.83632734530938\\
470	7.83632734530938\\
471	7.83632734530938\\
472	7.83632734530938\\
473	7.83632734530938\\
474	7.83632734530938\\
475	7.83632734530938\\
476	7.83632734530938\\
477	7.83632734530938\\
478	7.83632734530938\\
479	7.83632734530938\\
480	7.83632734530938\\
481	7.83632734530938\\
482	7.83632734530938\\
483	7.83632734530938\\
484	7.83632734530938\\
485	7.83632734530938\\
486	7.83632734530938\\
487	7.83632734530938\\
488	7.83632734530938\\
489	7.83632734530938\\
490	7.83632734530938\\
491	7.83632734530938\\
492	7.83632734530938\\
493	7.83632734530938\\
494	7.83632734530938\\
495	7.83632734530938\\
496	7.83632734530938\\
497	7.83632734530938\\
498	7.83632734530938\\
499	7.83632734530938\\
500	7.83632734530938\\
501	7.83632734530938\\
};
\addlegendentry{mean 7.84};

\addplot [color=black,solid,line width=3.0pt]
  table[row sep=crcr]{%
1	6\\
2	6\\
3	6\\
4	6\\
5	6\\
6	6\\
7	6\\
8	6\\
9	6\\
10	6\\
11	6\\
12	6\\
13	6\\
14	6\\
15	6\\
16	6\\
17	6\\
18	6\\
19	6\\
20	6\\
21	6\\
22	6\\
23	6\\
24	6\\
25	6\\
26	6\\
27	6\\
28	6\\
29	6\\
30	6\\
31	6\\
32	6\\
33	6\\
34	6\\
35	6\\
36	6\\
37	6\\
38	6\\
39	6\\
40	6\\
41	6\\
42	6\\
43	6\\
44	6\\
45	6\\
46	6\\
47	6\\
48	6\\
49	6\\
50	6\\
51	6\\
52	6\\
53	6\\
54	6\\
55	6\\
56	6\\
57	6\\
58	6\\
59	6\\
60	6\\
61	6\\
62	6\\
63	6\\
64	6\\
65	6\\
66	6\\
67	6\\
68	6\\
69	6\\
70	6\\
71	6\\
72	6\\
73	6\\
74	6\\
75	6\\
76	6\\
77	6\\
78	6\\
79	6\\
80	6\\
81	6\\
82	6\\
83	6\\
84	6\\
85	6\\
86	6\\
87	6\\
88	6\\
89	6\\
90	6\\
91	6\\
92	6\\
93	6\\
94	6\\
95	6\\
96	6\\
97	6\\
98	6\\
99	6\\
100	6\\
101	6\\
102	6\\
103	6\\
104	6\\
105	6\\
106	6\\
107	6\\
108	6\\
109	6\\
110	6\\
111	6\\
112	6\\
113	6\\
114	6\\
115	6\\
116	6\\
117	6\\
118	6\\
119	6\\
120	6\\
121	6\\
122	6\\
123	6\\
124	6\\
125	6\\
126	6\\
127	6\\
128	6\\
129	6\\
130	6\\
131	6\\
132	6\\
133	6\\
134	6\\
135	6\\
136	6\\
137	6\\
138	6\\
139	6\\
140	6\\
141	6\\
142	6\\
143	6\\
144	6\\
145	6\\
146	6\\
147	6\\
148	6\\
149	6\\
150	6\\
151	6\\
152	6\\
153	6\\
154	6\\
155	6\\
156	6\\
157	6\\
158	6\\
159	6\\
160	6\\
161	6\\
162	6\\
163	6\\
164	6\\
165	6\\
166	6\\
167	6\\
168	6\\
169	6\\
170	6\\
171	6\\
172	6\\
173	6\\
174	6\\
175	6\\
176	6\\
177	6\\
178	6\\
179	6\\
180	6\\
181	6\\
182	6\\
183	6\\
184	6\\
185	6\\
186	6\\
187	6\\
188	6\\
189	6\\
190	6\\
191	6\\
192	6\\
193	6\\
194	6\\
195	6\\
196	6\\
197	6\\
198	6\\
199	6\\
200	6\\
201	6\\
202	6\\
203	6\\
204	6\\
205	6\\
206	6\\
207	6\\
208	6\\
209	6\\
210	6\\
211	6\\
212	6\\
213	6\\
214	6\\
215	6\\
216	6\\
217	6\\
218	6\\
219	6\\
220	6\\
221	6\\
222	6\\
223	6\\
224	6\\
225	6\\
226	6\\
227	6\\
228	6\\
229	6\\
230	6\\
231	6\\
232	6\\
233	6\\
234	6\\
235	6\\
236	6\\
237	6\\
238	6\\
239	6\\
240	6\\
241	6\\
242	6\\
243	6\\
244	6\\
245	6\\
246	6\\
247	6\\
248	6\\
249	6\\
250	6\\
251	6\\
252	6\\
253	6\\
254	6\\
255	6\\
256	6\\
257	6\\
258	6\\
259	6\\
260	6\\
261	6\\
262	6\\
263	6\\
264	6\\
265	6\\
266	6\\
267	6\\
268	6\\
269	6\\
270	6\\
271	6\\
272	6\\
273	6\\
274	6\\
275	6\\
276	6\\
277	6\\
278	6\\
279	6\\
280	6\\
281	6\\
282	6\\
283	6\\
284	6\\
285	6\\
286	6\\
287	6\\
288	6\\
289	6\\
290	6\\
291	6\\
292	6\\
293	6\\
294	6\\
295	6\\
296	6\\
297	6\\
298	6\\
299	6\\
300	6\\
301	6\\
302	6\\
303	6\\
304	6\\
305	6\\
306	6\\
307	6\\
308	6\\
309	6\\
310	6\\
311	6\\
312	6\\
313	6\\
314	6\\
315	6\\
316	6\\
317	6\\
318	6\\
319	6\\
320	6\\
321	6\\
322	6\\
323	6\\
324	6\\
325	6\\
326	6\\
327	6\\
328	6\\
329	6\\
330	6\\
331	6\\
332	6\\
333	6\\
334	6\\
335	6\\
336	6\\
337	6\\
338	6\\
339	6\\
340	6\\
341	6\\
342	6\\
343	6\\
344	6\\
345	6\\
346	6\\
347	6\\
348	6\\
349	6\\
350	6\\
351	6\\
352	6\\
353	6\\
354	6\\
355	6\\
356	6\\
357	6\\
358	6\\
359	6\\
360	6\\
361	6\\
362	6\\
363	6\\
364	6\\
365	6\\
366	6\\
367	6\\
368	6\\
369	6\\
370	6\\
371	6\\
372	6\\
373	6\\
374	6\\
375	6\\
376	6\\
377	6\\
378	6\\
379	6\\
380	6\\
381	6\\
382	6\\
383	6\\
384	6\\
385	6\\
386	6\\
387	6\\
388	6\\
389	6\\
390	6\\
391	6\\
392	6\\
393	6\\
394	6\\
395	6\\
396	6\\
397	6\\
398	6\\
399	6\\
400	6\\
401	6\\
402	6\\
403	6\\
404	6\\
405	6\\
406	6\\
407	6\\
408	6\\
409	6\\
410	6\\
411	6\\
412	6\\
413	6\\
414	6\\
415	6\\
416	6\\
417	6\\
418	6\\
419	6\\
420	6\\
421	6\\
422	6\\
423	6\\
424	6\\
425	6\\
426	6\\
427	6\\
428	6\\
429	6\\
430	6\\
431	6\\
432	6\\
433	6\\
434	6\\
435	6\\
436	6\\
437	6\\
438	6\\
439	6\\
440	6\\
441	6\\
442	6\\
443	6\\
444	6\\
445	6\\
446	6\\
447	6\\
448	6\\
449	6\\
450	6\\
451	6\\
452	6\\
453	6\\
454	6\\
455	6\\
456	6\\
457	6\\
458	6\\
459	6\\
460	6\\
461	6\\
462	6\\
463	6\\
464	6\\
465	6\\
466	6\\
467	6\\
468	6\\
469	6\\
470	6\\
471	6\\
472	6\\
473	6\\
474	6\\
475	6\\
476	6\\
477	6\\
478	6\\
479	6\\
480	6\\
481	6\\
482	6\\
483	6\\
484	6\\
485	6\\
486	6\\
487	6\\
488	6\\
489	6\\
490	6\\
491	6\\
492	6\\
493	6\\
494	6\\
495	6\\
496	6\\
497	6\\
498	6\\
499	6\\
500	6\\
501	6\\
};
\addlegendentry{median 6.00};

\end{axis}
\end{tikzpicture}%

%% file: figs/tsukuba_intensity_9_time.tex
%
%
\begin{tikzpicture}

\begin{axis}[%
width=0.951\fwidth,
height=\fheight,
at={(0\fwidth,0\fheight)},
scale only axis,
xmin=1,
xmax=501,
xlabel={Frame number},
xmajorgrids,
ymin=2,
ymax=26,
ylabel={Time (ms)},
ymajorgrids,
axis background/.style={fill=white},
title style={font=\bfseries},
title={Raw Intensity},
legend style={at={(0.03,0.97)},anchor=north west,legend cell align=left,align=left,draw=white!15!black}
]
\addplot [color=white!30!black,solid,mark=+,mark options={solid}]
  table[row sep=crcr]{%
1	5\\
2	4\\
3	4\\
4	4\\
5	4\\
6	4\\
7	5\\
8	6\\
9	5\\
10	4\\
11	4\\
12	4\\
13	4\\
14	4\\
15	5\\
16	5\\
17	4\\
18	4\\
19	6\\
20	4\\
21	5\\
22	4\\
23	5\\
24	5\\
25	4\\
26	13\\
27	4\\
28	4\\
29	5\\
30	5\\
31	4\\
32	4\\
33	5\\
34	4\\
35	5\\
36	4\\
37	5\\
38	5\\
39	5\\
40	5\\
41	16\\
42	6\\
43	5\\
44	5\\
45	6\\
46	4\\
47	5\\
48	5\\
49	4\\
50	5\\
51	4\\
52	4\\
53	4\\
54	4\\
55	4\\
56	6\\
57	6\\
58	4\\
59	6\\
60	4\\
61	4\\
62	6\\
63	6\\
64	6\\
65	4\\
66	6\\
67	6\\
68	15\\
69	6\\
70	5\\
71	6\\
72	4\\
73	3\\
74	4\\
75	6\\
76	6\\
77	4\\
78	4\\
79	4\\
80	4\\
81	5\\
82	4\\
83	5\\
84	5\\
85	4\\
86	16\\
87	4\\
88	4\\
89	5\\
90	4\\
91	3\\
92	3\\
93	4\\
94	5\\
95	4\\
96	10\\
97	3\\
98	4\\
99	3\\
100	3\\
101	3\\
102	3\\
103	15\\
104	4\\
105	4\\
106	4\\
107	4\\
108	5\\
109	4\\
110	10\\
111	3\\
112	3\\
113	3\\
114	10\\
115	4\\
116	4\\
117	12\\
118	4\\
119	10\\
120	5\\
121	3\\
122	11\\
123	4\\
124	4\\
125	10\\
126	5\\
127	12\\
128	4\\
129	19\\
130	5\\
131	8\\
132	15\\
133	4\\
134	6\\
135	15\\
136	5\\
137	5\\
138	6\\
139	5\\
140	5\\
141	15\\
142	9\\
143	5\\
144	6\\
145	7\\
146	7\\
147	6\\
148	20\\
149	7\\
150	9\\
151	8\\
152	6\\
153	7\\
154	5\\
155	10\\
156	15\\
157	4\\
158	5\\
159	6\\
160	5\\
161	8\\
162	7\\
163	8\\
164	18\\
165	5\\
166	6\\
167	5\\
168	7\\
169	5\\
170	8\\
171	7\\
172	17\\
173	10\\
174	7\\
175	7\\
176	10\\
177	8\\
178	10\\
179	9\\
180	19\\
181	11\\
182	12\\
183	10\\
184	8\\
185	9\\
186	9\\
187	8\\
188	26\\
189	9\\
190	9\\
191	8\\
192	9\\
193	10\\
194	7\\
195	6\\
196	5\\
197	18\\
198	8\\
199	9\\
200	6\\
201	6\\
202	8\\
203	8\\
204	9\\
205	15\\
206	5\\
207	9\\
208	5\\
209	5\\
210	6\\
211	15\\
212	7\\
213	6\\
214	8\\
215	21\\
216	8\\
217	7\\
218	6\\
219	18\\
220	7\\
221	7\\
222	7\\
223	16\\
224	5\\
225	5\\
226	18\\
227	6\\
228	8\\
229	26\\
230	5\\
231	14\\
232	9\\
233	5\\
234	14\\
235	8\\
236	5\\
237	15\\
238	11\\
239	7\\
240	9\\
241	21\\
242	15\\
243	11\\
244	8\\
245	8\\
246	23\\
247	14\\
248	11\\
249	10\\
250	13\\
251	13\\
252	25\\
253	12\\
254	12\\
255	8\\
256	8\\
257	8\\
258	7\\
259	23\\
260	8\\
261	8\\
262	7\\
263	6\\
264	6\\
265	6\\
266	16\\
267	10\\
268	10\\
269	11\\
270	8\\
271	10\\
272	19\\
273	6\\
274	9\\
275	6\\
276	7\\
277	7\\
278	17\\
279	5\\
280	7\\
281	6\\
282	6\\
283	10\\
284	16\\
285	5\\
286	6\\
287	7\\
288	8\\
289	5\\
290	16\\
291	6\\
292	7\\
293	10\\
294	9\\
295	8\\
296	25\\
297	8\\
298	8\\
299	6\\
300	10\\
301	8\\
302	18\\
303	10\\
304	6\\
305	8\\
306	6\\
307	6\\
308	14\\
309	5\\
310	4\\
311	4\\
312	4\\
313	13\\
314	6\\
315	6\\
316	4\\
317	5\\
318	13\\
319	5\\
320	5\\
321	4\\
322	4\\
323	13\\
324	5\\
325	4\\
326	4\\
327	3\\
328	12\\
329	3\\
330	5\\
331	4\\
332	6\\
333	12\\
334	4\\
335	4\\
336	5\\
337	4\\
338	16\\
339	4\\
340	5\\
341	4\\
342	4\\
343	16\\
344	4\\
345	5\\
346	4\\
347	4\\
348	11\\
349	3\\
350	6\\
351	10\\
352	24\\
353	11\\
354	5\\
355	5\\
356	4\\
357	5\\
358	11\\
359	5\\
360	3\\
361	3\\
362	4\\
363	15\\
364	4\\
365	5\\
366	4\\
367	12\\
368	3\\
369	3\\
370	14\\
371	2\\
372	3\\
373	12\\
374	4\\
375	3\\
376	10\\
377	3\\
378	3\\
379	8\\
380	4\\
381	10\\
382	3\\
383	9\\
384	3\\
385	3\\
386	9\\
387	3\\
388	4\\
389	11\\
390	4\\
391	12\\
392	3\\
393	4\\
394	9\\
395	3\\
396	3\\
397	3\\
398	10\\
399	3\\
400	3\\
401	3\\
402	15\\
403	3\\
404	5\\
405	4\\
406	17\\
407	4\\
408	4\\
409	4\\
410	11\\
411	4\\
412	4\\
413	4\\
414	15\\
415	5\\
416	4\\
417	4\\
418	16\\
419	4\\
420	5\\
421	5\\
422	11\\
423	4\\
424	4\\
425	3\\
426	12\\
427	3\\
428	4\\
429	4\\
430	11\\
431	4\\
432	4\\
433	5\\
434	15\\
435	3\\
436	5\\
437	4\\
438	18\\
439	4\\
440	3\\
441	5\\
442	12\\
443	4\\
444	6\\
445	4\\
446	16\\
447	4\\
448	4\\
449	4\\
450	12\\
451	5\\
452	4\\
453	4\\
454	12\\
455	4\\
456	4\\
457	4\\
458	12\\
459	4\\
460	5\\
461	5\\
462	17\\
463	4\\
464	4\\
465	4\\
466	17\\
467	4\\
468	4\\
469	4\\
470	12\\
471	4\\
472	5\\
473	5\\
474	17\\
475	5\\
476	5\\
477	7\\
478	15\\
479	9\\
480	5\\
481	5\\
482	20\\
483	4\\
484	5\\
485	5\\
486	17\\
487	6\\
488	5\\
489	6\\
490	14\\
491	6\\
492	5\\
493	5\\
494	16\\
495	6\\
496	8\\
497	7\\
498	14\\
499	6\\
500	5\\
501	5\\
};
\addlegendentry{data};

\addplot [color=black,dashed,line width=2.0pt]
  table[row sep=crcr]{%
1	7.31736526946108\\
2	7.31736526946108\\
3	7.31736526946108\\
4	7.31736526946108\\
5	7.31736526946108\\
6	7.31736526946108\\
7	7.31736526946108\\
8	7.31736526946108\\
9	7.31736526946108\\
10	7.31736526946108\\
11	7.31736526946108\\
12	7.31736526946108\\
13	7.31736526946108\\
14	7.31736526946108\\
15	7.31736526946108\\
16	7.31736526946108\\
17	7.31736526946108\\
18	7.31736526946108\\
19	7.31736526946108\\
20	7.31736526946108\\
21	7.31736526946108\\
22	7.31736526946108\\
23	7.31736526946108\\
24	7.31736526946108\\
25	7.31736526946108\\
26	7.31736526946108\\
27	7.31736526946108\\
28	7.31736526946108\\
29	7.31736526946108\\
30	7.31736526946108\\
31	7.31736526946108\\
32	7.31736526946108\\
33	7.31736526946108\\
34	7.31736526946108\\
35	7.31736526946108\\
36	7.31736526946108\\
37	7.31736526946108\\
38	7.31736526946108\\
39	7.31736526946108\\
40	7.31736526946108\\
41	7.31736526946108\\
42	7.31736526946108\\
43	7.31736526946108\\
44	7.31736526946108\\
45	7.31736526946108\\
46	7.31736526946108\\
47	7.31736526946108\\
48	7.31736526946108\\
49	7.31736526946108\\
50	7.31736526946108\\
51	7.31736526946108\\
52	7.31736526946108\\
53	7.31736526946108\\
54	7.31736526946108\\
55	7.31736526946108\\
56	7.31736526946108\\
57	7.31736526946108\\
58	7.31736526946108\\
59	7.31736526946108\\
60	7.31736526946108\\
61	7.31736526946108\\
62	7.31736526946108\\
63	7.31736526946108\\
64	7.31736526946108\\
65	7.31736526946108\\
66	7.31736526946108\\
67	7.31736526946108\\
68	7.31736526946108\\
69	7.31736526946108\\
70	7.31736526946108\\
71	7.31736526946108\\
72	7.31736526946108\\
73	7.31736526946108\\
74	7.31736526946108\\
75	7.31736526946108\\
76	7.31736526946108\\
77	7.31736526946108\\
78	7.31736526946108\\
79	7.31736526946108\\
80	7.31736526946108\\
81	7.31736526946108\\
82	7.31736526946108\\
83	7.31736526946108\\
84	7.31736526946108\\
85	7.31736526946108\\
86	7.31736526946108\\
87	7.31736526946108\\
88	7.31736526946108\\
89	7.31736526946108\\
90	7.31736526946108\\
91	7.31736526946108\\
92	7.31736526946108\\
93	7.31736526946108\\
94	7.31736526946108\\
95	7.31736526946108\\
96	7.31736526946108\\
97	7.31736526946108\\
98	7.31736526946108\\
99	7.31736526946108\\
100	7.31736526946108\\
101	7.31736526946108\\
102	7.31736526946108\\
103	7.31736526946108\\
104	7.31736526946108\\
105	7.31736526946108\\
106	7.31736526946108\\
107	7.31736526946108\\
108	7.31736526946108\\
109	7.31736526946108\\
110	7.31736526946108\\
111	7.31736526946108\\
112	7.31736526946108\\
113	7.31736526946108\\
114	7.31736526946108\\
115	7.31736526946108\\
116	7.31736526946108\\
117	7.31736526946108\\
118	7.31736526946108\\
119	7.31736526946108\\
120	7.31736526946108\\
121	7.31736526946108\\
122	7.31736526946108\\
123	7.31736526946108\\
124	7.31736526946108\\
125	7.31736526946108\\
126	7.31736526946108\\
127	7.31736526946108\\
128	7.31736526946108\\
129	7.31736526946108\\
130	7.31736526946108\\
131	7.31736526946108\\
132	7.31736526946108\\
133	7.31736526946108\\
134	7.31736526946108\\
135	7.31736526946108\\
136	7.31736526946108\\
137	7.31736526946108\\
138	7.31736526946108\\
139	7.31736526946108\\
140	7.31736526946108\\
141	7.31736526946108\\
142	7.31736526946108\\
143	7.31736526946108\\
144	7.31736526946108\\
145	7.31736526946108\\
146	7.31736526946108\\
147	7.31736526946108\\
148	7.31736526946108\\
149	7.31736526946108\\
150	7.31736526946108\\
151	7.31736526946108\\
152	7.31736526946108\\
153	7.31736526946108\\
154	7.31736526946108\\
155	7.31736526946108\\
156	7.31736526946108\\
157	7.31736526946108\\
158	7.31736526946108\\
159	7.31736526946108\\
160	7.31736526946108\\
161	7.31736526946108\\
162	7.31736526946108\\
163	7.31736526946108\\
164	7.31736526946108\\
165	7.31736526946108\\
166	7.31736526946108\\
167	7.31736526946108\\
168	7.31736526946108\\
169	7.31736526946108\\
170	7.31736526946108\\
171	7.31736526946108\\
172	7.31736526946108\\
173	7.31736526946108\\
174	7.31736526946108\\
175	7.31736526946108\\
176	7.31736526946108\\
177	7.31736526946108\\
178	7.31736526946108\\
179	7.31736526946108\\
180	7.31736526946108\\
181	7.31736526946108\\
182	7.31736526946108\\
183	7.31736526946108\\
184	7.31736526946108\\
185	7.31736526946108\\
186	7.31736526946108\\
187	7.31736526946108\\
188	7.31736526946108\\
189	7.31736526946108\\
190	7.31736526946108\\
191	7.31736526946108\\
192	7.31736526946108\\
193	7.31736526946108\\
194	7.31736526946108\\
195	7.31736526946108\\
196	7.31736526946108\\
197	7.31736526946108\\
198	7.31736526946108\\
199	7.31736526946108\\
200	7.31736526946108\\
201	7.31736526946108\\
202	7.31736526946108\\
203	7.31736526946108\\
204	7.31736526946108\\
205	7.31736526946108\\
206	7.31736526946108\\
207	7.31736526946108\\
208	7.31736526946108\\
209	7.31736526946108\\
210	7.31736526946108\\
211	7.31736526946108\\
212	7.31736526946108\\
213	7.31736526946108\\
214	7.31736526946108\\
215	7.31736526946108\\
216	7.31736526946108\\
217	7.31736526946108\\
218	7.31736526946108\\
219	7.31736526946108\\
220	7.31736526946108\\
221	7.31736526946108\\
222	7.31736526946108\\
223	7.31736526946108\\
224	7.31736526946108\\
225	7.31736526946108\\
226	7.31736526946108\\
227	7.31736526946108\\
228	7.31736526946108\\
229	7.31736526946108\\
230	7.31736526946108\\
231	7.31736526946108\\
232	7.31736526946108\\
233	7.31736526946108\\
234	7.31736526946108\\
235	7.31736526946108\\
236	7.31736526946108\\
237	7.31736526946108\\
238	7.31736526946108\\
239	7.31736526946108\\
240	7.31736526946108\\
241	7.31736526946108\\
242	7.31736526946108\\
243	7.31736526946108\\
244	7.31736526946108\\
245	7.31736526946108\\
246	7.31736526946108\\
247	7.31736526946108\\
248	7.31736526946108\\
249	7.31736526946108\\
250	7.31736526946108\\
251	7.31736526946108\\
252	7.31736526946108\\
253	7.31736526946108\\
254	7.31736526946108\\
255	7.31736526946108\\
256	7.31736526946108\\
257	7.31736526946108\\
258	7.31736526946108\\
259	7.31736526946108\\
260	7.31736526946108\\
261	7.31736526946108\\
262	7.31736526946108\\
263	7.31736526946108\\
264	7.31736526946108\\
265	7.31736526946108\\
266	7.31736526946108\\
267	7.31736526946108\\
268	7.31736526946108\\
269	7.31736526946108\\
270	7.31736526946108\\
271	7.31736526946108\\
272	7.31736526946108\\
273	7.31736526946108\\
274	7.31736526946108\\
275	7.31736526946108\\
276	7.31736526946108\\
277	7.31736526946108\\
278	7.31736526946108\\
279	7.31736526946108\\
280	7.31736526946108\\
281	7.31736526946108\\
282	7.31736526946108\\
283	7.31736526946108\\
284	7.31736526946108\\
285	7.31736526946108\\
286	7.31736526946108\\
287	7.31736526946108\\
288	7.31736526946108\\
289	7.31736526946108\\
290	7.31736526946108\\
291	7.31736526946108\\
292	7.31736526946108\\
293	7.31736526946108\\
294	7.31736526946108\\
295	7.31736526946108\\
296	7.31736526946108\\
297	7.31736526946108\\
298	7.31736526946108\\
299	7.31736526946108\\
300	7.31736526946108\\
301	7.31736526946108\\
302	7.31736526946108\\
303	7.31736526946108\\
304	7.31736526946108\\
305	7.31736526946108\\
306	7.31736526946108\\
307	7.31736526946108\\
308	7.31736526946108\\
309	7.31736526946108\\
310	7.31736526946108\\
311	7.31736526946108\\
312	7.31736526946108\\
313	7.31736526946108\\
314	7.31736526946108\\
315	7.31736526946108\\
316	7.31736526946108\\
317	7.31736526946108\\
318	7.31736526946108\\
319	7.31736526946108\\
320	7.31736526946108\\
321	7.31736526946108\\
322	7.31736526946108\\
323	7.31736526946108\\
324	7.31736526946108\\
325	7.31736526946108\\
326	7.31736526946108\\
327	7.31736526946108\\
328	7.31736526946108\\
329	7.31736526946108\\
330	7.31736526946108\\
331	7.31736526946108\\
332	7.31736526946108\\
333	7.31736526946108\\
334	7.31736526946108\\
335	7.31736526946108\\
336	7.31736526946108\\
337	7.31736526946108\\
338	7.31736526946108\\
339	7.31736526946108\\
340	7.31736526946108\\
341	7.31736526946108\\
342	7.31736526946108\\
343	7.31736526946108\\
344	7.31736526946108\\
345	7.31736526946108\\
346	7.31736526946108\\
347	7.31736526946108\\
348	7.31736526946108\\
349	7.31736526946108\\
350	7.31736526946108\\
351	7.31736526946108\\
352	7.31736526946108\\
353	7.31736526946108\\
354	7.31736526946108\\
355	7.31736526946108\\
356	7.31736526946108\\
357	7.31736526946108\\
358	7.31736526946108\\
359	7.31736526946108\\
360	7.31736526946108\\
361	7.31736526946108\\
362	7.31736526946108\\
363	7.31736526946108\\
364	7.31736526946108\\
365	7.31736526946108\\
366	7.31736526946108\\
367	7.31736526946108\\
368	7.31736526946108\\
369	7.31736526946108\\
370	7.31736526946108\\
371	7.31736526946108\\
372	7.31736526946108\\
373	7.31736526946108\\
374	7.31736526946108\\
375	7.31736526946108\\
376	7.31736526946108\\
377	7.31736526946108\\
378	7.31736526946108\\
379	7.31736526946108\\
380	7.31736526946108\\
381	7.31736526946108\\
382	7.31736526946108\\
383	7.31736526946108\\
384	7.31736526946108\\
385	7.31736526946108\\
386	7.31736526946108\\
387	7.31736526946108\\
388	7.31736526946108\\
389	7.31736526946108\\
390	7.31736526946108\\
391	7.31736526946108\\
392	7.31736526946108\\
393	7.31736526946108\\
394	7.31736526946108\\
395	7.31736526946108\\
396	7.31736526946108\\
397	7.31736526946108\\
398	7.31736526946108\\
399	7.31736526946108\\
400	7.31736526946108\\
401	7.31736526946108\\
402	7.31736526946108\\
403	7.31736526946108\\
404	7.31736526946108\\
405	7.31736526946108\\
406	7.31736526946108\\
407	7.31736526946108\\
408	7.31736526946108\\
409	7.31736526946108\\
410	7.31736526946108\\
411	7.31736526946108\\
412	7.31736526946108\\
413	7.31736526946108\\
414	7.31736526946108\\
415	7.31736526946108\\
416	7.31736526946108\\
417	7.31736526946108\\
418	7.31736526946108\\
419	7.31736526946108\\
420	7.31736526946108\\
421	7.31736526946108\\
422	7.31736526946108\\
423	7.31736526946108\\
424	7.31736526946108\\
425	7.31736526946108\\
426	7.31736526946108\\
427	7.31736526946108\\
428	7.31736526946108\\
429	7.31736526946108\\
430	7.31736526946108\\
431	7.31736526946108\\
432	7.31736526946108\\
433	7.31736526946108\\
434	7.31736526946108\\
435	7.31736526946108\\
436	7.31736526946108\\
437	7.31736526946108\\
438	7.31736526946108\\
439	7.31736526946108\\
440	7.31736526946108\\
441	7.31736526946108\\
442	7.31736526946108\\
443	7.31736526946108\\
444	7.31736526946108\\
445	7.31736526946108\\
446	7.31736526946108\\
447	7.31736526946108\\
448	7.31736526946108\\
449	7.31736526946108\\
450	7.31736526946108\\
451	7.31736526946108\\
452	7.31736526946108\\
453	7.31736526946108\\
454	7.31736526946108\\
455	7.31736526946108\\
456	7.31736526946108\\
457	7.31736526946108\\
458	7.31736526946108\\
459	7.31736526946108\\
460	7.31736526946108\\
461	7.31736526946108\\
462	7.31736526946108\\
463	7.31736526946108\\
464	7.31736526946108\\
465	7.31736526946108\\
466	7.31736526946108\\
467	7.31736526946108\\
468	7.31736526946108\\
469	7.31736526946108\\
470	7.31736526946108\\
471	7.31736526946108\\
472	7.31736526946108\\
473	7.31736526946108\\
474	7.31736526946108\\
475	7.31736526946108\\
476	7.31736526946108\\
477	7.31736526946108\\
478	7.31736526946108\\
479	7.31736526946108\\
480	7.31736526946108\\
481	7.31736526946108\\
482	7.31736526946108\\
483	7.31736526946108\\
484	7.31736526946108\\
485	7.31736526946108\\
486	7.31736526946108\\
487	7.31736526946108\\
488	7.31736526946108\\
489	7.31736526946108\\
490	7.31736526946108\\
491	7.31736526946108\\
492	7.31736526946108\\
493	7.31736526946108\\
494	7.31736526946108\\
495	7.31736526946108\\
496	7.31736526946108\\
497	7.31736526946108\\
498	7.31736526946108\\
499	7.31736526946108\\
500	7.31736526946108\\
501	7.31736526946108\\
};
\addlegendentry{mean 7.32};

\addplot [color=black,solid,line width=3.0pt]
  table[row sep=crcr]{%
1	5\\
2	5\\
3	5\\
4	5\\
5	5\\
6	5\\
7	5\\
8	5\\
9	5\\
10	5\\
11	5\\
12	5\\
13	5\\
14	5\\
15	5\\
16	5\\
17	5\\
18	5\\
19	5\\
20	5\\
21	5\\
22	5\\
23	5\\
24	5\\
25	5\\
26	5\\
27	5\\
28	5\\
29	5\\
30	5\\
31	5\\
32	5\\
33	5\\
34	5\\
35	5\\
36	5\\
37	5\\
38	5\\
39	5\\
40	5\\
41	5\\
42	5\\
43	5\\
44	5\\
45	5\\
46	5\\
47	5\\
48	5\\
49	5\\
50	5\\
51	5\\
52	5\\
53	5\\
54	5\\
55	5\\
56	5\\
57	5\\
58	5\\
59	5\\
60	5\\
61	5\\
62	5\\
63	5\\
64	5\\
65	5\\
66	5\\
67	5\\
68	5\\
69	5\\
70	5\\
71	5\\
72	5\\
73	5\\
74	5\\
75	5\\
76	5\\
77	5\\
78	5\\
79	5\\
80	5\\
81	5\\
82	5\\
83	5\\
84	5\\
85	5\\
86	5\\
87	5\\
88	5\\
89	5\\
90	5\\
91	5\\
92	5\\
93	5\\
94	5\\
95	5\\
96	5\\
97	5\\
98	5\\
99	5\\
100	5\\
101	5\\
102	5\\
103	5\\
104	5\\
105	5\\
106	5\\
107	5\\
108	5\\
109	5\\
110	5\\
111	5\\
112	5\\
113	5\\
114	5\\
115	5\\
116	5\\
117	5\\
118	5\\
119	5\\
120	5\\
121	5\\
122	5\\
123	5\\
124	5\\
125	5\\
126	5\\
127	5\\
128	5\\
129	5\\
130	5\\
131	5\\
132	5\\
133	5\\
134	5\\
135	5\\
136	5\\
137	5\\
138	5\\
139	5\\
140	5\\
141	5\\
142	5\\
143	5\\
144	5\\
145	5\\
146	5\\
147	5\\
148	5\\
149	5\\
150	5\\
151	5\\
152	5\\
153	5\\
154	5\\
155	5\\
156	5\\
157	5\\
158	5\\
159	5\\
160	5\\
161	5\\
162	5\\
163	5\\
164	5\\
165	5\\
166	5\\
167	5\\
168	5\\
169	5\\
170	5\\
171	5\\
172	5\\
173	5\\
174	5\\
175	5\\
176	5\\
177	5\\
178	5\\
179	5\\
180	5\\
181	5\\
182	5\\
183	5\\
184	5\\
185	5\\
186	5\\
187	5\\
188	5\\
189	5\\
190	5\\
191	5\\
192	5\\
193	5\\
194	5\\
195	5\\
196	5\\
197	5\\
198	5\\
199	5\\
200	5\\
201	5\\
202	5\\
203	5\\
204	5\\
205	5\\
206	5\\
207	5\\
208	5\\
209	5\\
210	5\\
211	5\\
212	5\\
213	5\\
214	5\\
215	5\\
216	5\\
217	5\\
218	5\\
219	5\\
220	5\\
221	5\\
222	5\\
223	5\\
224	5\\
225	5\\
226	5\\
227	5\\
228	5\\
229	5\\
230	5\\
231	5\\
232	5\\
233	5\\
234	5\\
235	5\\
236	5\\
237	5\\
238	5\\
239	5\\
240	5\\
241	5\\
242	5\\
243	5\\
244	5\\
245	5\\
246	5\\
247	5\\
248	5\\
249	5\\
250	5\\
251	5\\
252	5\\
253	5\\
254	5\\
255	5\\
256	5\\
257	5\\
258	5\\
259	5\\
260	5\\
261	5\\
262	5\\
263	5\\
264	5\\
265	5\\
266	5\\
267	5\\
268	5\\
269	5\\
270	5\\
271	5\\
272	5\\
273	5\\
274	5\\
275	5\\
276	5\\
277	5\\
278	5\\
279	5\\
280	5\\
281	5\\
282	5\\
283	5\\
284	5\\
285	5\\
286	5\\
287	5\\
288	5\\
289	5\\
290	5\\
291	5\\
292	5\\
293	5\\
294	5\\
295	5\\
296	5\\
297	5\\
298	5\\
299	5\\
300	5\\
301	5\\
302	5\\
303	5\\
304	5\\
305	5\\
306	5\\
307	5\\
308	5\\
309	5\\
310	5\\
311	5\\
312	5\\
313	5\\
314	5\\
315	5\\
316	5\\
317	5\\
318	5\\
319	5\\
320	5\\
321	5\\
322	5\\
323	5\\
324	5\\
325	5\\
326	5\\
327	5\\
328	5\\
329	5\\
330	5\\
331	5\\
332	5\\
333	5\\
334	5\\
335	5\\
336	5\\
337	5\\
338	5\\
339	5\\
340	5\\
341	5\\
342	5\\
343	5\\
344	5\\
345	5\\
346	5\\
347	5\\
348	5\\
349	5\\
350	5\\
351	5\\
352	5\\
353	5\\
354	5\\
355	5\\
356	5\\
357	5\\
358	5\\
359	5\\
360	5\\
361	5\\
362	5\\
363	5\\
364	5\\
365	5\\
366	5\\
367	5\\
368	5\\
369	5\\
370	5\\
371	5\\
372	5\\
373	5\\
374	5\\
375	5\\
376	5\\
377	5\\
378	5\\
379	5\\
380	5\\
381	5\\
382	5\\
383	5\\
384	5\\
385	5\\
386	5\\
387	5\\
388	5\\
389	5\\
390	5\\
391	5\\
392	5\\
393	5\\
394	5\\
395	5\\
396	5\\
397	5\\
398	5\\
399	5\\
400	5\\
401	5\\
402	5\\
403	5\\
404	5\\
405	5\\
406	5\\
407	5\\
408	5\\
409	5\\
410	5\\
411	5\\
412	5\\
413	5\\
414	5\\
415	5\\
416	5\\
417	5\\
418	5\\
419	5\\
420	5\\
421	5\\
422	5\\
423	5\\
424	5\\
425	5\\
426	5\\
427	5\\
428	5\\
429	5\\
430	5\\
431	5\\
432	5\\
433	5\\
434	5\\
435	5\\
436	5\\
437	5\\
438	5\\
439	5\\
440	5\\
441	5\\
442	5\\
443	5\\
444	5\\
445	5\\
446	5\\
447	5\\
448	5\\
449	5\\
450	5\\
451	5\\
452	5\\
453	5\\
454	5\\
455	5\\
456	5\\
457	5\\
458	5\\
459	5\\
460	5\\
461	5\\
462	5\\
463	5\\
464	5\\
465	5\\
466	5\\
467	5\\
468	5\\
469	5\\
470	5\\
471	5\\
472	5\\
473	5\\
474	5\\
475	5\\
476	5\\
477	5\\
478	5\\
479	5\\
480	5\\
481	5\\
482	5\\
483	5\\
484	5\\
485	5\\
486	5\\
487	5\\
488	5\\
489	5\\
490	5\\
491	5\\
492	5\\
493	5\\
494	5\\
495	5\\
496	5\\
497	5\\
498	5\\
499	5\\
500	5\\
501	5\\
};
\addlegendentry{median 5.00};

\end{axis}
\end{tikzpicture}%

%% file: figs/tsukuba_bitplanes_9_iters.tex
%
%
\begin{tikzpicture}

\begin{axis}[%
width=0.951\fwidth,
height=\fheight,
at={(0\fwidth,0\fheight)},
scale only axis,
xmin=1,
xmax=501,
xlabel={Frame number},
xmajorgrids,
ymin=0,
ymax=100,
ylabel={\# iterations},
ymajorgrids,
axis background/.style={fill=white},
title style={font=\bfseries},
title={Bit-Planes},
legend style={at={(0.03,0.97)},anchor=north west,legend cell align=left,align=left,draw=white!15!black}
]
\addplot [color=white!30!black,solid,mark=+,mark options={solid}]
  table[row sep=crcr]{%
1	0\\
2	4\\
3	4\\
4	4\\
5	4\\
6	5\\
7	5\\
8	5\\
9	5\\
10	4\\
11	5\\
12	4\\
13	4\\
14	4\\
15	4\\
16	5\\
17	5\\
18	5\\
19	6\\
20	6\\
21	6\\
22	5\\
23	5\\
24	6\\
25	6\\
26	7\\
27	7\\
28	7\\
29	7\\
30	7\\
31	7\\
32	5\\
33	5\\
34	4\\
35	4\\
36	5\\
37	5\\
38	5\\
39	5\\
40	6\\
41	6\\
42	6\\
43	6\\
44	6\\
45	6\\
46	6\\
47	6\\
48	6\\
49	6\\
50	6\\
51	6\\
52	6\\
53	6\\
54	5\\
55	5\\
56	4\\
57	5\\
58	5\\
59	5\\
60	6\\
61	6\\
62	5\\
63	5\\
64	6\\
65	5\\
66	6\\
67	5\\
68	5\\
69	4\\
70	4\\
71	5\\
72	6\\
73	5\\
74	6\\
75	6\\
76	5\\
77	5\\
78	6\\
79	4\\
80	5\\
81	5\\
82	5\\
83	6\\
84	4\\
85	5\\
86	4\\
87	4\\
88	5\\
89	4\\
90	5\\
91	4\\
92	5\\
93	5\\
94	5\\
95	5\\
96	4\\
97	4\\
98	5\\
99	5\\
100	5\\
101	4\\
102	5\\
103	5\\
104	6\\
105	5\\
106	6\\
107	7\\
108	7\\
109	8\\
110	8\\
111	7\\
112	8\\
113	8\\
114	8\\
115	6\\
116	6\\
117	6\\
118	6\\
119	7\\
120	6\\
121	8\\
122	9\\
123	8\\
124	10\\
125	10\\
126	6\\
127	7\\
128	10\\
129	4\\
130	16\\
131	16\\
132	16\\
133	9\\
134	7\\
135	9\\
136	10\\
137	9\\
138	15\\
139	24\\
140	24\\
141	28\\
142	28\\
143	28\\
144	21\\
145	6\\
146	7\\
147	16\\
148	24\\
149	26\\
150	27\\
151	20\\
152	16\\
153	16\\
154	13\\
155	6\\
156	9\\
157	14\\
158	16\\
159	17\\
160	15\\
161	9\\
162	12\\
163	5\\
164	11\\
165	14\\
166	11\\
167	19\\
168	17\\
169	16\\
170	12\\
171	11\\
172	10\\
173	15\\
174	8\\
175	18\\
176	27\\
177	30\\
178	53\\
179	26\\
180	43\\
181	23\\
182	38\\
183	54\\
184	17\\
185	20\\
186	52\\
187	42\\
188	24\\
189	26\\
190	26\\
191	26\\
192	23\\
193	20\\
194	100\\
195	24\\
196	22\\
197	19\\
198	23\\
199	13\\
200	10\\
201	17\\
202	35\\
203	24\\
204	16\\
205	19\\
206	16\\
207	12\\
208	15\\
209	18\\
210	28\\
211	22\\
212	19\\
213	15\\
214	10\\
215	15\\
216	15\\
217	23\\
218	19\\
219	14\\
220	11\\
221	9\\
222	11\\
223	13\\
224	13\\
225	11\\
226	15\\
227	12\\
228	13\\
229	14\\
230	11\\
231	18\\
232	14\\
233	11\\
234	12\\
235	9\\
236	12\\
237	11\\
238	13\\
239	13\\
240	13\\
241	12\\
242	12\\
243	13\\
244	12\\
245	11\\
246	15\\
247	13\\
248	10\\
249	10\\
250	11\\
251	11\\
252	11\\
253	12\\
254	11\\
255	10\\
256	9\\
257	8\\
258	10\\
259	11\\
260	7\\
261	8\\
262	7\\
263	8\\
264	8\\
265	8\\
266	8\\
267	9\\
268	9\\
269	8\\
270	7\\
271	7\\
272	8\\
273	14\\
274	10\\
275	10\\
276	11\\
277	9\\
278	11\\
279	11\\
280	11\\
281	12\\
282	8\\
283	12\\
284	12\\
285	11\\
286	10\\
287	10\\
288	5\\
289	11\\
290	11\\
291	10\\
292	7\\
293	9\\
294	11\\
295	10\\
296	10\\
297	12\\
298	9\\
299	9\\
300	10\\
301	12\\
302	13\\
303	8\\
304	9\\
305	9\\
306	9\\
307	6\\
308	8\\
309	8\\
310	8\\
311	8\\
312	7\\
313	7\\
314	7\\
315	7\\
316	6\\
317	6\\
318	5\\
319	6\\
320	5\\
321	5\\
322	4\\
323	5\\
324	5\\
325	5\\
326	5\\
327	5\\
328	5\\
329	4\\
330	4\\
331	4\\
332	5\\
333	5\\
334	3\\
335	4\\
336	4\\
337	5\\
338	5\\
339	4\\
340	4\\
341	5\\
342	5\\
343	4\\
344	4\\
345	4\\
346	5\\
347	4\\
348	4\\
349	4\\
350	5\\
351	5\\
352	5\\
353	5\\
354	4\\
355	5\\
356	5\\
357	6\\
358	5\\
359	4\\
360	5\\
361	5\\
362	5\\
363	5\\
364	5\\
365	5\\
366	6\\
367	7\\
368	6\\
369	7\\
370	9\\
371	9\\
372	9\\
373	7\\
374	8\\
375	10\\
376	9\\
377	14\\
378	6\\
379	8\\
380	7\\
381	9\\
382	11\\
383	8\\
384	7\\
385	9\\
386	12\\
387	10\\
388	10\\
389	11\\
390	9\\
391	9\\
392	10\\
393	10\\
394	11\\
395	7\\
396	8\\
397	7\\
398	8\\
399	7\\
400	7\\
401	9\\
402	9\\
403	5\\
404	5\\
405	6\\
406	7\\
407	5\\
408	5\\
409	4\\
410	5\\
411	5\\
412	5\\
413	5\\
414	6\\
415	5\\
416	6\\
417	6\\
418	7\\
419	6\\
420	7\\
421	6\\
422	7\\
423	6\\
424	6\\
425	6\\
426	6\\
427	6\\
428	7\\
429	7\\
430	7\\
431	6\\
432	7\\
433	7\\
434	8\\
435	7\\
436	7\\
437	7\\
438	7\\
439	8\\
440	7\\
441	8\\
442	8\\
443	7\\
444	9\\
445	7\\
446	7\\
447	8\\
448	7\\
449	8\\
450	8\\
451	6\\
452	6\\
453	8\\
454	8\\
455	6\\
456	6\\
457	7\\
458	6\\
459	6\\
460	6\\
461	7\\
462	7\\
463	6\\
464	7\\
465	6\\
466	8\\
467	7\\
468	7\\
469	8\\
470	8\\
471	6\\
472	6\\
473	6\\
474	7\\
475	9\\
476	6\\
477	8\\
478	9\\
479	8\\
480	7\\
481	9\\
482	9\\
483	5\\
484	6\\
485	7\\
486	7\\
487	7\\
488	7\\
489	7\\
490	8\\
491	6\\
492	6\\
493	6\\
494	7\\
495	9\\
496	6\\
497	6\\
498	6\\
499	7\\
500	5\\
501	5\\
};
\addlegendentry{data};

\addplot [color=black,dashed,line width=2.0pt]
  table[row sep=crcr]{%
1	9.47105788423154\\
2	9.47105788423154\\
3	9.47105788423154\\
4	9.47105788423154\\
5	9.47105788423154\\
6	9.47105788423154\\
7	9.47105788423154\\
8	9.47105788423154\\
9	9.47105788423154\\
10	9.47105788423154\\
11	9.47105788423154\\
12	9.47105788423154\\
13	9.47105788423154\\
14	9.47105788423154\\
15	9.47105788423154\\
16	9.47105788423154\\
17	9.47105788423154\\
18	9.47105788423154\\
19	9.47105788423154\\
20	9.47105788423154\\
21	9.47105788423154\\
22	9.47105788423154\\
23	9.47105788423154\\
24	9.47105788423154\\
25	9.47105788423154\\
26	9.47105788423154\\
27	9.47105788423154\\
28	9.47105788423154\\
29	9.47105788423154\\
30	9.47105788423154\\
31	9.47105788423154\\
32	9.47105788423154\\
33	9.47105788423154\\
34	9.47105788423154\\
35	9.47105788423154\\
36	9.47105788423154\\
37	9.47105788423154\\
38	9.47105788423154\\
39	9.47105788423154\\
40	9.47105788423154\\
41	9.47105788423154\\
42	9.47105788423154\\
43	9.47105788423154\\
44	9.47105788423154\\
45	9.47105788423154\\
46	9.47105788423154\\
47	9.47105788423154\\
48	9.47105788423154\\
49	9.47105788423154\\
50	9.47105788423154\\
51	9.47105788423154\\
52	9.47105788423154\\
53	9.47105788423154\\
54	9.47105788423154\\
55	9.47105788423154\\
56	9.47105788423154\\
57	9.47105788423154\\
58	9.47105788423154\\
59	9.47105788423154\\
60	9.47105788423154\\
61	9.47105788423154\\
62	9.47105788423154\\
63	9.47105788423154\\
64	9.47105788423154\\
65	9.47105788423154\\
66	9.47105788423154\\
67	9.47105788423154\\
68	9.47105788423154\\
69	9.47105788423154\\
70	9.47105788423154\\
71	9.47105788423154\\
72	9.47105788423154\\
73	9.47105788423154\\
74	9.47105788423154\\
75	9.47105788423154\\
76	9.47105788423154\\
77	9.47105788423154\\
78	9.47105788423154\\
79	9.47105788423154\\
80	9.47105788423154\\
81	9.47105788423154\\
82	9.47105788423154\\
83	9.47105788423154\\
84	9.47105788423154\\
85	9.47105788423154\\
86	9.47105788423154\\
87	9.47105788423154\\
88	9.47105788423154\\
89	9.47105788423154\\
90	9.47105788423154\\
91	9.47105788423154\\
92	9.47105788423154\\
93	9.47105788423154\\
94	9.47105788423154\\
95	9.47105788423154\\
96	9.47105788423154\\
97	9.47105788423154\\
98	9.47105788423154\\
99	9.47105788423154\\
100	9.47105788423154\\
101	9.47105788423154\\
102	9.47105788423154\\
103	9.47105788423154\\
104	9.47105788423154\\
105	9.47105788423154\\
106	9.47105788423154\\
107	9.47105788423154\\
108	9.47105788423154\\
109	9.47105788423154\\
110	9.47105788423154\\
111	9.47105788423154\\
112	9.47105788423154\\
113	9.47105788423154\\
114	9.47105788423154\\
115	9.47105788423154\\
116	9.47105788423154\\
117	9.47105788423154\\
118	9.47105788423154\\
119	9.47105788423154\\
120	9.47105788423154\\
121	9.47105788423154\\
122	9.47105788423154\\
123	9.47105788423154\\
124	9.47105788423154\\
125	9.47105788423154\\
126	9.47105788423154\\
127	9.47105788423154\\
128	9.47105788423154\\
129	9.47105788423154\\
130	9.47105788423154\\
131	9.47105788423154\\
132	9.47105788423154\\
133	9.47105788423154\\
134	9.47105788423154\\
135	9.47105788423154\\
136	9.47105788423154\\
137	9.47105788423154\\
138	9.47105788423154\\
139	9.47105788423154\\
140	9.47105788423154\\
141	9.47105788423154\\
142	9.47105788423154\\
143	9.47105788423154\\
144	9.47105788423154\\
145	9.47105788423154\\
146	9.47105788423154\\
147	9.47105788423154\\
148	9.47105788423154\\
149	9.47105788423154\\
150	9.47105788423154\\
151	9.47105788423154\\
152	9.47105788423154\\
153	9.47105788423154\\
154	9.47105788423154\\
155	9.47105788423154\\
156	9.47105788423154\\
157	9.47105788423154\\
158	9.47105788423154\\
159	9.47105788423154\\
160	9.47105788423154\\
161	9.47105788423154\\
162	9.47105788423154\\
163	9.47105788423154\\
164	9.47105788423154\\
165	9.47105788423154\\
166	9.47105788423154\\
167	9.47105788423154\\
168	9.47105788423154\\
169	9.47105788423154\\
170	9.47105788423154\\
171	9.47105788423154\\
172	9.47105788423154\\
173	9.47105788423154\\
174	9.47105788423154\\
175	9.47105788423154\\
176	9.47105788423154\\
177	9.47105788423154\\
178	9.47105788423154\\
179	9.47105788423154\\
180	9.47105788423154\\
181	9.47105788423154\\
182	9.47105788423154\\
183	9.47105788423154\\
184	9.47105788423154\\
185	9.47105788423154\\
186	9.47105788423154\\
187	9.47105788423154\\
188	9.47105788423154\\
189	9.47105788423154\\
190	9.47105788423154\\
191	9.47105788423154\\
192	9.47105788423154\\
193	9.47105788423154\\
194	9.47105788423154\\
195	9.47105788423154\\
196	9.47105788423154\\
197	9.47105788423154\\
198	9.47105788423154\\
199	9.47105788423154\\
200	9.47105788423154\\
201	9.47105788423154\\
202	9.47105788423154\\
203	9.47105788423154\\
204	9.47105788423154\\
205	9.47105788423154\\
206	9.47105788423154\\
207	9.47105788423154\\
208	9.47105788423154\\
209	9.47105788423154\\
210	9.47105788423154\\
211	9.47105788423154\\
212	9.47105788423154\\
213	9.47105788423154\\
214	9.47105788423154\\
215	9.47105788423154\\
216	9.47105788423154\\
217	9.47105788423154\\
218	9.47105788423154\\
219	9.47105788423154\\
220	9.47105788423154\\
221	9.47105788423154\\
222	9.47105788423154\\
223	9.47105788423154\\
224	9.47105788423154\\
225	9.47105788423154\\
226	9.47105788423154\\
227	9.47105788423154\\
228	9.47105788423154\\
229	9.47105788423154\\
230	9.47105788423154\\
231	9.47105788423154\\
232	9.47105788423154\\
233	9.47105788423154\\
234	9.47105788423154\\
235	9.47105788423154\\
236	9.47105788423154\\
237	9.47105788423154\\
238	9.47105788423154\\
239	9.47105788423154\\
240	9.47105788423154\\
241	9.47105788423154\\
242	9.47105788423154\\
243	9.47105788423154\\
244	9.47105788423154\\
245	9.47105788423154\\
246	9.47105788423154\\
247	9.47105788423154\\
248	9.47105788423154\\
249	9.47105788423154\\
250	9.47105788423154\\
251	9.47105788423154\\
252	9.47105788423154\\
253	9.47105788423154\\
254	9.47105788423154\\
255	9.47105788423154\\
256	9.47105788423154\\
257	9.47105788423154\\
258	9.47105788423154\\
259	9.47105788423154\\
260	9.47105788423154\\
261	9.47105788423154\\
262	9.47105788423154\\
263	9.47105788423154\\
264	9.47105788423154\\
265	9.47105788423154\\
266	9.47105788423154\\
267	9.47105788423154\\
268	9.47105788423154\\
269	9.47105788423154\\
270	9.47105788423154\\
271	9.47105788423154\\
272	9.47105788423154\\
273	9.47105788423154\\
274	9.47105788423154\\
275	9.47105788423154\\
276	9.47105788423154\\
277	9.47105788423154\\
278	9.47105788423154\\
279	9.47105788423154\\
280	9.47105788423154\\
281	9.47105788423154\\
282	9.47105788423154\\
283	9.47105788423154\\
284	9.47105788423154\\
285	9.47105788423154\\
286	9.47105788423154\\
287	9.47105788423154\\
288	9.47105788423154\\
289	9.47105788423154\\
290	9.47105788423154\\
291	9.47105788423154\\
292	9.47105788423154\\
293	9.47105788423154\\
294	9.47105788423154\\
295	9.47105788423154\\
296	9.47105788423154\\
297	9.47105788423154\\
298	9.47105788423154\\
299	9.47105788423154\\
300	9.47105788423154\\
301	9.47105788423154\\
302	9.47105788423154\\
303	9.47105788423154\\
304	9.47105788423154\\
305	9.47105788423154\\
306	9.47105788423154\\
307	9.47105788423154\\
308	9.47105788423154\\
309	9.47105788423154\\
310	9.47105788423154\\
311	9.47105788423154\\
312	9.47105788423154\\
313	9.47105788423154\\
314	9.47105788423154\\
315	9.47105788423154\\
316	9.47105788423154\\
317	9.47105788423154\\
318	9.47105788423154\\
319	9.47105788423154\\
320	9.47105788423154\\
321	9.47105788423154\\
322	9.47105788423154\\
323	9.47105788423154\\
324	9.47105788423154\\
325	9.47105788423154\\
326	9.47105788423154\\
327	9.47105788423154\\
328	9.47105788423154\\
329	9.47105788423154\\
330	9.47105788423154\\
331	9.47105788423154\\
332	9.47105788423154\\
333	9.47105788423154\\
334	9.47105788423154\\
335	9.47105788423154\\
336	9.47105788423154\\
337	9.47105788423154\\
338	9.47105788423154\\
339	9.47105788423154\\
340	9.47105788423154\\
341	9.47105788423154\\
342	9.47105788423154\\
343	9.47105788423154\\
344	9.47105788423154\\
345	9.47105788423154\\
346	9.47105788423154\\
347	9.47105788423154\\
348	9.47105788423154\\
349	9.47105788423154\\
350	9.47105788423154\\
351	9.47105788423154\\
352	9.47105788423154\\
353	9.47105788423154\\
354	9.47105788423154\\
355	9.47105788423154\\
356	9.47105788423154\\
357	9.47105788423154\\
358	9.47105788423154\\
359	9.47105788423154\\
360	9.47105788423154\\
361	9.47105788423154\\
362	9.47105788423154\\
363	9.47105788423154\\
364	9.47105788423154\\
365	9.47105788423154\\
366	9.47105788423154\\
367	9.47105788423154\\
368	9.47105788423154\\
369	9.47105788423154\\
370	9.47105788423154\\
371	9.47105788423154\\
372	9.47105788423154\\
373	9.47105788423154\\
374	9.47105788423154\\
375	9.47105788423154\\
376	9.47105788423154\\
377	9.47105788423154\\
378	9.47105788423154\\
379	9.47105788423154\\
380	9.47105788423154\\
381	9.47105788423154\\
382	9.47105788423154\\
383	9.47105788423154\\
384	9.47105788423154\\
385	9.47105788423154\\
386	9.47105788423154\\
387	9.47105788423154\\
388	9.47105788423154\\
389	9.47105788423154\\
390	9.47105788423154\\
391	9.47105788423154\\
392	9.47105788423154\\
393	9.47105788423154\\
394	9.47105788423154\\
395	9.47105788423154\\
396	9.47105788423154\\
397	9.47105788423154\\
398	9.47105788423154\\
399	9.47105788423154\\
400	9.47105788423154\\
401	9.47105788423154\\
402	9.47105788423154\\
403	9.47105788423154\\
404	9.47105788423154\\
405	9.47105788423154\\
406	9.47105788423154\\
407	9.47105788423154\\
408	9.47105788423154\\
409	9.47105788423154\\
410	9.47105788423154\\
411	9.47105788423154\\
412	9.47105788423154\\
413	9.47105788423154\\
414	9.47105788423154\\
415	9.47105788423154\\
416	9.47105788423154\\
417	9.47105788423154\\
418	9.47105788423154\\
419	9.47105788423154\\
420	9.47105788423154\\
421	9.47105788423154\\
422	9.47105788423154\\
423	9.47105788423154\\
424	9.47105788423154\\
425	9.47105788423154\\
426	9.47105788423154\\
427	9.47105788423154\\
428	9.47105788423154\\
429	9.47105788423154\\
430	9.47105788423154\\
431	9.47105788423154\\
432	9.47105788423154\\
433	9.47105788423154\\
434	9.47105788423154\\
435	9.47105788423154\\
436	9.47105788423154\\
437	9.47105788423154\\
438	9.47105788423154\\
439	9.47105788423154\\
440	9.47105788423154\\
441	9.47105788423154\\
442	9.47105788423154\\
443	9.47105788423154\\
444	9.47105788423154\\
445	9.47105788423154\\
446	9.47105788423154\\
447	9.47105788423154\\
448	9.47105788423154\\
449	9.47105788423154\\
450	9.47105788423154\\
451	9.47105788423154\\
452	9.47105788423154\\
453	9.47105788423154\\
454	9.47105788423154\\
455	9.47105788423154\\
456	9.47105788423154\\
457	9.47105788423154\\
458	9.47105788423154\\
459	9.47105788423154\\
460	9.47105788423154\\
461	9.47105788423154\\
462	9.47105788423154\\
463	9.47105788423154\\
464	9.47105788423154\\
465	9.47105788423154\\
466	9.47105788423154\\
467	9.47105788423154\\
468	9.47105788423154\\
469	9.47105788423154\\
470	9.47105788423154\\
471	9.47105788423154\\
472	9.47105788423154\\
473	9.47105788423154\\
474	9.47105788423154\\
475	9.47105788423154\\
476	9.47105788423154\\
477	9.47105788423154\\
478	9.47105788423154\\
479	9.47105788423154\\
480	9.47105788423154\\
481	9.47105788423154\\
482	9.47105788423154\\
483	9.47105788423154\\
484	9.47105788423154\\
485	9.47105788423154\\
486	9.47105788423154\\
487	9.47105788423154\\
488	9.47105788423154\\
489	9.47105788423154\\
490	9.47105788423154\\
491	9.47105788423154\\
492	9.47105788423154\\
493	9.47105788423154\\
494	9.47105788423154\\
495	9.47105788423154\\
496	9.47105788423154\\
497	9.47105788423154\\
498	9.47105788423154\\
499	9.47105788423154\\
500	9.47105788423154\\
501	9.47105788423154\\
};
\addlegendentry{mean 9.47};

\addplot [color=black,solid,line width=3.0pt]
  table[row sep=crcr]{%
1	7\\
2	7\\
3	7\\
4	7\\
5	7\\
6	7\\
7	7\\
8	7\\
9	7\\
10	7\\
11	7\\
12	7\\
13	7\\
14	7\\
15	7\\
16	7\\
17	7\\
18	7\\
19	7\\
20	7\\
21	7\\
22	7\\
23	7\\
24	7\\
25	7\\
26	7\\
27	7\\
28	7\\
29	7\\
30	7\\
31	7\\
32	7\\
33	7\\
34	7\\
35	7\\
36	7\\
37	7\\
38	7\\
39	7\\
40	7\\
41	7\\
42	7\\
43	7\\
44	7\\
45	7\\
46	7\\
47	7\\
48	7\\
49	7\\
50	7\\
51	7\\
52	7\\
53	7\\
54	7\\
55	7\\
56	7\\
57	7\\
58	7\\
59	7\\
60	7\\
61	7\\
62	7\\
63	7\\
64	7\\
65	7\\
66	7\\
67	7\\
68	7\\
69	7\\
70	7\\
71	7\\
72	7\\
73	7\\
74	7\\
75	7\\
76	7\\
77	7\\
78	7\\
79	7\\
80	7\\
81	7\\
82	7\\
83	7\\
84	7\\
85	7\\
86	7\\
87	7\\
88	7\\
89	7\\
90	7\\
91	7\\
92	7\\
93	7\\
94	7\\
95	7\\
96	7\\
97	7\\
98	7\\
99	7\\
100	7\\
101	7\\
102	7\\
103	7\\
104	7\\
105	7\\
106	7\\
107	7\\
108	7\\
109	7\\
110	7\\
111	7\\
112	7\\
113	7\\
114	7\\
115	7\\
116	7\\
117	7\\
118	7\\
119	7\\
120	7\\
121	7\\
122	7\\
123	7\\
124	7\\
125	7\\
126	7\\
127	7\\
128	7\\
129	7\\
130	7\\
131	7\\
132	7\\
133	7\\
134	7\\
135	7\\
136	7\\
137	7\\
138	7\\
139	7\\
140	7\\
141	7\\
142	7\\
143	7\\
144	7\\
145	7\\
146	7\\
147	7\\
148	7\\
149	7\\
150	7\\
151	7\\
152	7\\
153	7\\
154	7\\
155	7\\
156	7\\
157	7\\
158	7\\
159	7\\
160	7\\
161	7\\
162	7\\
163	7\\
164	7\\
165	7\\
166	7\\
167	7\\
168	7\\
169	7\\
170	7\\
171	7\\
172	7\\
173	7\\
174	7\\
175	7\\
176	7\\
177	7\\
178	7\\
179	7\\
180	7\\
181	7\\
182	7\\
183	7\\
184	7\\
185	7\\
186	7\\
187	7\\
188	7\\
189	7\\
190	7\\
191	7\\
192	7\\
193	7\\
194	7\\
195	7\\
196	7\\
197	7\\
198	7\\
199	7\\
200	7\\
201	7\\
202	7\\
203	7\\
204	7\\
205	7\\
206	7\\
207	7\\
208	7\\
209	7\\
210	7\\
211	7\\
212	7\\
213	7\\
214	7\\
215	7\\
216	7\\
217	7\\
218	7\\
219	7\\
220	7\\
221	7\\
222	7\\
223	7\\
224	7\\
225	7\\
226	7\\
227	7\\
228	7\\
229	7\\
230	7\\
231	7\\
232	7\\
233	7\\
234	7\\
235	7\\
236	7\\
237	7\\
238	7\\
239	7\\
240	7\\
241	7\\
242	7\\
243	7\\
244	7\\
245	7\\
246	7\\
247	7\\
248	7\\
249	7\\
250	7\\
251	7\\
252	7\\
253	7\\
254	7\\
255	7\\
256	7\\
257	7\\
258	7\\
259	7\\
260	7\\
261	7\\
262	7\\
263	7\\
264	7\\
265	7\\
266	7\\
267	7\\
268	7\\
269	7\\
270	7\\
271	7\\
272	7\\
273	7\\
274	7\\
275	7\\
276	7\\
277	7\\
278	7\\
279	7\\
280	7\\
281	7\\
282	7\\
283	7\\
284	7\\
285	7\\
286	7\\
287	7\\
288	7\\
289	7\\
290	7\\
291	7\\
292	7\\
293	7\\
294	7\\
295	7\\
296	7\\
297	7\\
298	7\\
299	7\\
300	7\\
301	7\\
302	7\\
303	7\\
304	7\\
305	7\\
306	7\\
307	7\\
308	7\\
309	7\\
310	7\\
311	7\\
312	7\\
313	7\\
314	7\\
315	7\\
316	7\\
317	7\\
318	7\\
319	7\\
320	7\\
321	7\\
322	7\\
323	7\\
324	7\\
325	7\\
326	7\\
327	7\\
328	7\\
329	7\\
330	7\\
331	7\\
332	7\\
333	7\\
334	7\\
335	7\\
336	7\\
337	7\\
338	7\\
339	7\\
340	7\\
341	7\\
342	7\\
343	7\\
344	7\\
345	7\\
346	7\\
347	7\\
348	7\\
349	7\\
350	7\\
351	7\\
352	7\\
353	7\\
354	7\\
355	7\\
356	7\\
357	7\\
358	7\\
359	7\\
360	7\\
361	7\\
362	7\\
363	7\\
364	7\\
365	7\\
366	7\\
367	7\\
368	7\\
369	7\\
370	7\\
371	7\\
372	7\\
373	7\\
374	7\\
375	7\\
376	7\\
377	7\\
378	7\\
379	7\\
380	7\\
381	7\\
382	7\\
383	7\\
384	7\\
385	7\\
386	7\\
387	7\\
388	7\\
389	7\\
390	7\\
391	7\\
392	7\\
393	7\\
394	7\\
395	7\\
396	7\\
397	7\\
398	7\\
399	7\\
400	7\\
401	7\\
402	7\\
403	7\\
404	7\\
405	7\\
406	7\\
407	7\\
408	7\\
409	7\\
410	7\\
411	7\\
412	7\\
413	7\\
414	7\\
415	7\\
416	7\\
417	7\\
418	7\\
419	7\\
420	7\\
421	7\\
422	7\\
423	7\\
424	7\\
425	7\\
426	7\\
427	7\\
428	7\\
429	7\\
430	7\\
431	7\\
432	7\\
433	7\\
434	7\\
435	7\\
436	7\\
437	7\\
438	7\\
439	7\\
440	7\\
441	7\\
442	7\\
443	7\\
444	7\\
445	7\\
446	7\\
447	7\\
448	7\\
449	7\\
450	7\\
451	7\\
452	7\\
453	7\\
454	7\\
455	7\\
456	7\\
457	7\\
458	7\\
459	7\\
460	7\\
461	7\\
462	7\\
463	7\\
464	7\\
465	7\\
466	7\\
467	7\\
468	7\\
469	7\\
470	7\\
471	7\\
472	7\\
473	7\\
474	7\\
475	7\\
476	7\\
477	7\\
478	7\\
479	7\\
480	7\\
481	7\\
482	7\\
483	7\\
484	7\\
485	7\\
486	7\\
487	7\\
488	7\\
489	7\\
490	7\\
491	7\\
492	7\\
493	7\\
494	7\\
495	7\\
496	7\\
497	7\\
498	7\\
499	7\\
500	7\\
501	7\\
};
\addlegendentry{median 7.00};

\end{axis}
\end{tikzpicture}%

%% file: figs/tsukuba_bitplanes_9_time.tex
%
%
\begin{tikzpicture}

\begin{axis}[%
width=0.951\fwidth,
height=\fheight,
at={(0\fwidth,0\fheight)},
scale only axis,
xmin=1,
xmax=501,
xlabel={Frame number},
xmajorgrids,
ymin=22,
ymax=234,
ylabel={Time (ms)},
ymajorgrids,
axis background/.style={fill=white},
title style={font=\bfseries},
title={Bit-Planes},
legend style={at={(0.03,0.97)},anchor=north west,legend cell align=left,align=left,draw=white!15!black}
]
\addplot [color=white!30!black,solid,mark=+,mark options={solid}]
  table[row sep=crcr]{%
1	31\\
2	22\\
3	44\\
4	48\\
5	42\\
6	45\\
7	52\\
8	54\\
9	47\\
10	31\\
11	36\\
12	33\\
13	40\\
14	46\\
15	40\\
16	64\\
17	34\\
18	33\\
19	87\\
20	39\\
21	45\\
22	36\\
23	31\\
24	39\\
25	39\\
26	38\\
27	45\\
28	42\\
29	41\\
30	43\\
31	44\\
32	103\\
33	48\\
34	37\\
35	38\\
36	38\\
37	36\\
38	43\\
39	39\\
40	42\\
41	53\\
42	51\\
43	52\\
44	50\\
45	49\\
46	49\\
47	33\\
48	41\\
49	35\\
50	37\\
51	40\\
52	39\\
53	39\\
54	38\\
55	39\\
56	34\\
57	38\\
58	36\\
59	31\\
60	33\\
61	35\\
62	50\\
63	50\\
64	51\\
65	35\\
66	86\\
67	77\\
68	36\\
69	35\\
70	34\\
71	37\\
72	33\\
73	40\\
74	45\\
75	47\\
76	31\\
77	31\\
78	32\\
79	27\\
80	36\\
81	31\\
82	36\\
83	42\\
84	29\\
85	31\\
86	45\\
87	30\\
88	234\\
89	83\\
90	34\\
91	32\\
92	44\\
93	38\\
94	34\\
95	39\\
96	31\\
97	28\\
98	35\\
99	36\\
100	28\\
101	35\\
102	39\\
103	30\\
104	68\\
105	41\\
106	43\\
107	36\\
108	40\\
109	39\\
110	67\\
111	37\\
112	39\\
113	41\\
114	35\\
115	70\\
116	32\\
117	34\\
118	36\\
119	29\\
120	29\\
121	29\\
122	41\\
123	70\\
124	34\\
125	32\\
126	32\\
127	44\\
128	47\\
129	60\\
130	48\\
131	54\\
132	45\\
133	37\\
134	83\\
135	36\\
136	49\\
137	43\\
138	63\\
139	66\\
140	63\\
141	94\\
142	71\\
143	67\\
144	82\\
145	107\\
146	49\\
147	50\\
148	66\\
149	74\\
150	66\\
151	68\\
152	62\\
153	68\\
154	59\\
155	52\\
156	121\\
157	47\\
158	59\\
159	65\\
160	58\\
161	53\\
162	59\\
163	59\\
164	61\\
165	66\\
166	113\\
167	64\\
168	63\\
169	75\\
170	171\\
171	73\\
172	123\\
173	118\\
174	79\\
175	102\\
176	160\\
177	93\\
178	134\\
179	97\\
180	183\\
181	128\\
182	134\\
183	143\\
184	106\\
185	170\\
186	121\\
187	132\\
188	141\\
189	158\\
190	224\\
191	115\\
192	106\\
193	105\\
194	230\\
195	70\\
196	98\\
197	111\\
198	130\\
199	99\\
200	82\\
201	104\\
202	180\\
203	74\\
204	85\\
205	221\\
206	80\\
207	75\\
208	76\\
209	92\\
210	167\\
211	90\\
212	91\\
213	87\\
214	91\\
215	90\\
216	87\\
217	89\\
218	126\\
219	66\\
220	66\\
221	62\\
222	74\\
223	77\\
224	77\\
225	75\\
226	125\\
227	71\\
228	72\\
229	75\\
230	73\\
231	83\\
232	133\\
233	206\\
234	76\\
235	70\\
236	80\\
237	80\\
238	73\\
239	116\\
240	77\\
241	78\\
242	93\\
243	111\\
244	123\\
245	100\\
246	143\\
247	69\\
248	82\\
249	89\\
250	110\\
251	110\\
252	105\\
253	143\\
254	61\\
255	69\\
256	69\\
257	82\\
258	106\\
259	68\\
260	116\\
261	49\\
262	61\\
263	64\\
264	67\\
265	64\\
266	66\\
267	115\\
268	57\\
269	60\\
270	62\\
271	66\\
272	72\\
273	140\\
274	71\\
275	75\\
276	63\\
277	80\\
278	68\\
279	119\\
280	72\\
281	78\\
282	71\\
283	83\\
284	85\\
285	131\\
286	70\\
287	78\\
288	76\\
289	83\\
290	79\\
291	134\\
292	65\\
293	70\\
294	73\\
295	74\\
296	74\\
297	127\\
298	73\\
299	73\\
300	64\\
301	68\\
302	79\\
303	104\\
304	55\\
305	65\\
306	58\\
307	60\\
308	92\\
309	43\\
310	61\\
311	60\\
312	59\\
313	97\\
314	52\\
315	57\\
316	58\\
317	48\\
318	75\\
319	38\\
320	43\\
321	37\\
322	43\\
323	69\\
324	35\\
325	38\\
326	37\\
327	37\\
328	74\\
329	41\\
330	34\\
331	36\\
332	39\\
333	75\\
334	25\\
335	36\\
336	34\\
337	37\\
338	60\\
339	34\\
340	40\\
341	41\\
342	51\\
343	80\\
344	38\\
345	45\\
346	50\\
347	40\\
348	93\\
349	45\\
350	34\\
351	41\\
352	37\\
353	73\\
354	30\\
355	37\\
356	43\\
357	44\\
358	77\\
359	36\\
360	35\\
361	39\\
362	37\\
363	73\\
364	34\\
365	34\\
366	36\\
367	41\\
368	71\\
369	30\\
370	38\\
371	46\\
372	41\\
373	80\\
374	44\\
375	48\\
376	45\\
377	54\\
378	82\\
379	30\\
380	33\\
381	40\\
382	43\\
383	77\\
384	31\\
385	52\\
386	45\\
387	82\\
388	37\\
389	44\\
390	47\\
391	76\\
392	39\\
393	32\\
394	41\\
395	79\\
396	34\\
397	36\\
398	72\\
399	79\\
400	47\\
401	53\\
402	37\\
403	77\\
404	32\\
405	36\\
406	38\\
407	73\\
408	30\\
409	32\\
410	69\\
411	72\\
412	32\\
413	44\\
414	37\\
415	80\\
416	43\\
417	43\\
418	45\\
419	80\\
420	44\\
421	43\\
422	44\\
423	77\\
424	41\\
425	42\\
426	45\\
427	81\\
428	43\\
429	42\\
430	44\\
431	88\\
432	47\\
433	44\\
434	47\\
435	90\\
436	44\\
437	46\\
438	45\\
439	88\\
440	71\\
441	44\\
442	51\\
443	91\\
444	46\\
445	49\\
446	67\\
447	97\\
448	53\\
449	49\\
450	50\\
451	100\\
452	49\\
453	54\\
454	57\\
455	108\\
456	45\\
457	54\\
458	46\\
459	91\\
460	43\\
461	53\\
462	53\\
463	88\\
464	43\\
465	60\\
466	58\\
467	100\\
468	49\\
469	54\\
470	54\\
471	91\\
472	46\\
473	53\\
474	49\\
475	94\\
476	47\\
477	52\\
478	52\\
479	109\\
480	58\\
481	62\\
482	58\\
483	103\\
484	47\\
485	56\\
486	59\\
487	99\\
488	53\\
489	52\\
490	56\\
491	102\\
492	46\\
493	51\\
494	50\\
495	104\\
496	50\\
497	49\\
498	229\\
499	93\\
500	44\\
501	46\\
};
\addlegendentry{data};

\addplot [color=black,dashed,line width=2.0pt]
  table[row sep=crcr]{%
1	64.187624750499\\
2	64.187624750499\\
3	64.187624750499\\
4	64.187624750499\\
5	64.187624750499\\
6	64.187624750499\\
7	64.187624750499\\
8	64.187624750499\\
9	64.187624750499\\
10	64.187624750499\\
11	64.187624750499\\
12	64.187624750499\\
13	64.187624750499\\
14	64.187624750499\\
15	64.187624750499\\
16	64.187624750499\\
17	64.187624750499\\
18	64.187624750499\\
19	64.187624750499\\
20	64.187624750499\\
21	64.187624750499\\
22	64.187624750499\\
23	64.187624750499\\
24	64.187624750499\\
25	64.187624750499\\
26	64.187624750499\\
27	64.187624750499\\
28	64.187624750499\\
29	64.187624750499\\
30	64.187624750499\\
31	64.187624750499\\
32	64.187624750499\\
33	64.187624750499\\
34	64.187624750499\\
35	64.187624750499\\
36	64.187624750499\\
37	64.187624750499\\
38	64.187624750499\\
39	64.187624750499\\
40	64.187624750499\\
41	64.187624750499\\
42	64.187624750499\\
43	64.187624750499\\
44	64.187624750499\\
45	64.187624750499\\
46	64.187624750499\\
47	64.187624750499\\
48	64.187624750499\\
49	64.187624750499\\
50	64.187624750499\\
51	64.187624750499\\
52	64.187624750499\\
53	64.187624750499\\
54	64.187624750499\\
55	64.187624750499\\
56	64.187624750499\\
57	64.187624750499\\
58	64.187624750499\\
59	64.187624750499\\
60	64.187624750499\\
61	64.187624750499\\
62	64.187624750499\\
63	64.187624750499\\
64	64.187624750499\\
65	64.187624750499\\
66	64.187624750499\\
67	64.187624750499\\
68	64.187624750499\\
69	64.187624750499\\
70	64.187624750499\\
71	64.187624750499\\
72	64.187624750499\\
73	64.187624750499\\
74	64.187624750499\\
75	64.187624750499\\
76	64.187624750499\\
77	64.187624750499\\
78	64.187624750499\\
79	64.187624750499\\
80	64.187624750499\\
81	64.187624750499\\
82	64.187624750499\\
83	64.187624750499\\
84	64.187624750499\\
85	64.187624750499\\
86	64.187624750499\\
87	64.187624750499\\
88	64.187624750499\\
89	64.187624750499\\
90	64.187624750499\\
91	64.187624750499\\
92	64.187624750499\\
93	64.187624750499\\
94	64.187624750499\\
95	64.187624750499\\
96	64.187624750499\\
97	64.187624750499\\
98	64.187624750499\\
99	64.187624750499\\
100	64.187624750499\\
101	64.187624750499\\
102	64.187624750499\\
103	64.187624750499\\
104	64.187624750499\\
105	64.187624750499\\
106	64.187624750499\\
107	64.187624750499\\
108	64.187624750499\\
109	64.187624750499\\
110	64.187624750499\\
111	64.187624750499\\
112	64.187624750499\\
113	64.187624750499\\
114	64.187624750499\\
115	64.187624750499\\
116	64.187624750499\\
117	64.187624750499\\
118	64.187624750499\\
119	64.187624750499\\
120	64.187624750499\\
121	64.187624750499\\
122	64.187624750499\\
123	64.187624750499\\
124	64.187624750499\\
125	64.187624750499\\
126	64.187624750499\\
127	64.187624750499\\
128	64.187624750499\\
129	64.187624750499\\
130	64.187624750499\\
131	64.187624750499\\
132	64.187624750499\\
133	64.187624750499\\
134	64.187624750499\\
135	64.187624750499\\
136	64.187624750499\\
137	64.187624750499\\
138	64.187624750499\\
139	64.187624750499\\
140	64.187624750499\\
141	64.187624750499\\
142	64.187624750499\\
143	64.187624750499\\
144	64.187624750499\\
145	64.187624750499\\
146	64.187624750499\\
147	64.187624750499\\
148	64.187624750499\\
149	64.187624750499\\
150	64.187624750499\\
151	64.187624750499\\
152	64.187624750499\\
153	64.187624750499\\
154	64.187624750499\\
155	64.187624750499\\
156	64.187624750499\\
157	64.187624750499\\
158	64.187624750499\\
159	64.187624750499\\
160	64.187624750499\\
161	64.187624750499\\
162	64.187624750499\\
163	64.187624750499\\
164	64.187624750499\\
165	64.187624750499\\
166	64.187624750499\\
167	64.187624750499\\
168	64.187624750499\\
169	64.187624750499\\
170	64.187624750499\\
171	64.187624750499\\
172	64.187624750499\\
173	64.187624750499\\
174	64.187624750499\\
175	64.187624750499\\
176	64.187624750499\\
177	64.187624750499\\
178	64.187624750499\\
179	64.187624750499\\
180	64.187624750499\\
181	64.187624750499\\
182	64.187624750499\\
183	64.187624750499\\
184	64.187624750499\\
185	64.187624750499\\
186	64.187624750499\\
187	64.187624750499\\
188	64.187624750499\\
189	64.187624750499\\
190	64.187624750499\\
191	64.187624750499\\
192	64.187624750499\\
193	64.187624750499\\
194	64.187624750499\\
195	64.187624750499\\
196	64.187624750499\\
197	64.187624750499\\
198	64.187624750499\\
199	64.187624750499\\
200	64.187624750499\\
201	64.187624750499\\
202	64.187624750499\\
203	64.187624750499\\
204	64.187624750499\\
205	64.187624750499\\
206	64.187624750499\\
207	64.187624750499\\
208	64.187624750499\\
209	64.187624750499\\
210	64.187624750499\\
211	64.187624750499\\
212	64.187624750499\\
213	64.187624750499\\
214	64.187624750499\\
215	64.187624750499\\
216	64.187624750499\\
217	64.187624750499\\
218	64.187624750499\\
219	64.187624750499\\
220	64.187624750499\\
221	64.187624750499\\
222	64.187624750499\\
223	64.187624750499\\
224	64.187624750499\\
225	64.187624750499\\
226	64.187624750499\\
227	64.187624750499\\
228	64.187624750499\\
229	64.187624750499\\
230	64.187624750499\\
231	64.187624750499\\
232	64.187624750499\\
233	64.187624750499\\
234	64.187624750499\\
235	64.187624750499\\
236	64.187624750499\\
237	64.187624750499\\
238	64.187624750499\\
239	64.187624750499\\
240	64.187624750499\\
241	64.187624750499\\
242	64.187624750499\\
243	64.187624750499\\
244	64.187624750499\\
245	64.187624750499\\
246	64.187624750499\\
247	64.187624750499\\
248	64.187624750499\\
249	64.187624750499\\
250	64.187624750499\\
251	64.187624750499\\
252	64.187624750499\\
253	64.187624750499\\
254	64.187624750499\\
255	64.187624750499\\
256	64.187624750499\\
257	64.187624750499\\
258	64.187624750499\\
259	64.187624750499\\
260	64.187624750499\\
261	64.187624750499\\
262	64.187624750499\\
263	64.187624750499\\
264	64.187624750499\\
265	64.187624750499\\
266	64.187624750499\\
267	64.187624750499\\
268	64.187624750499\\
269	64.187624750499\\
270	64.187624750499\\
271	64.187624750499\\
272	64.187624750499\\
273	64.187624750499\\
274	64.187624750499\\
275	64.187624750499\\
276	64.187624750499\\
277	64.187624750499\\
278	64.187624750499\\
279	64.187624750499\\
280	64.187624750499\\
281	64.187624750499\\
282	64.187624750499\\
283	64.187624750499\\
284	64.187624750499\\
285	64.187624750499\\
286	64.187624750499\\
287	64.187624750499\\
288	64.187624750499\\
289	64.187624750499\\
290	64.187624750499\\
291	64.187624750499\\
292	64.187624750499\\
293	64.187624750499\\
294	64.187624750499\\
295	64.187624750499\\
296	64.187624750499\\
297	64.187624750499\\
298	64.187624750499\\
299	64.187624750499\\
300	64.187624750499\\
301	64.187624750499\\
302	64.187624750499\\
303	64.187624750499\\
304	64.187624750499\\
305	64.187624750499\\
306	64.187624750499\\
307	64.187624750499\\
308	64.187624750499\\
309	64.187624750499\\
310	64.187624750499\\
311	64.187624750499\\
312	64.187624750499\\
313	64.187624750499\\
314	64.187624750499\\
315	64.187624750499\\
316	64.187624750499\\
317	64.187624750499\\
318	64.187624750499\\
319	64.187624750499\\
320	64.187624750499\\
321	64.187624750499\\
322	64.187624750499\\
323	64.187624750499\\
324	64.187624750499\\
325	64.187624750499\\
326	64.187624750499\\
327	64.187624750499\\
328	64.187624750499\\
329	64.187624750499\\
330	64.187624750499\\
331	64.187624750499\\
332	64.187624750499\\
333	64.187624750499\\
334	64.187624750499\\
335	64.187624750499\\
336	64.187624750499\\
337	64.187624750499\\
338	64.187624750499\\
339	64.187624750499\\
340	64.187624750499\\
341	64.187624750499\\
342	64.187624750499\\
343	64.187624750499\\
344	64.187624750499\\
345	64.187624750499\\
346	64.187624750499\\
347	64.187624750499\\
348	64.187624750499\\
349	64.187624750499\\
350	64.187624750499\\
351	64.187624750499\\
352	64.187624750499\\
353	64.187624750499\\
354	64.187624750499\\
355	64.187624750499\\
356	64.187624750499\\
357	64.187624750499\\
358	64.187624750499\\
359	64.187624750499\\
360	64.187624750499\\
361	64.187624750499\\
362	64.187624750499\\
363	64.187624750499\\
364	64.187624750499\\
365	64.187624750499\\
366	64.187624750499\\
367	64.187624750499\\
368	64.187624750499\\
369	64.187624750499\\
370	64.187624750499\\
371	64.187624750499\\
372	64.187624750499\\
373	64.187624750499\\
374	64.187624750499\\
375	64.187624750499\\
376	64.187624750499\\
377	64.187624750499\\
378	64.187624750499\\
379	64.187624750499\\
380	64.187624750499\\
381	64.187624750499\\
382	64.187624750499\\
383	64.187624750499\\
384	64.187624750499\\
385	64.187624750499\\
386	64.187624750499\\
387	64.187624750499\\
388	64.187624750499\\
389	64.187624750499\\
390	64.187624750499\\
391	64.187624750499\\
392	64.187624750499\\
393	64.187624750499\\
394	64.187624750499\\
395	64.187624750499\\
396	64.187624750499\\
397	64.187624750499\\
398	64.187624750499\\
399	64.187624750499\\
400	64.187624750499\\
401	64.187624750499\\
402	64.187624750499\\
403	64.187624750499\\
404	64.187624750499\\
405	64.187624750499\\
406	64.187624750499\\
407	64.187624750499\\
408	64.187624750499\\
409	64.187624750499\\
410	64.187624750499\\
411	64.187624750499\\
412	64.187624750499\\
413	64.187624750499\\
414	64.187624750499\\
415	64.187624750499\\
416	64.187624750499\\
417	64.187624750499\\
418	64.187624750499\\
419	64.187624750499\\
420	64.187624750499\\
421	64.187624750499\\
422	64.187624750499\\
423	64.187624750499\\
424	64.187624750499\\
425	64.187624750499\\
426	64.187624750499\\
427	64.187624750499\\
428	64.187624750499\\
429	64.187624750499\\
430	64.187624750499\\
431	64.187624750499\\
432	64.187624750499\\
433	64.187624750499\\
434	64.187624750499\\
435	64.187624750499\\
436	64.187624750499\\
437	64.187624750499\\
438	64.187624750499\\
439	64.187624750499\\
440	64.187624750499\\
441	64.187624750499\\
442	64.187624750499\\
443	64.187624750499\\
444	64.187624750499\\
445	64.187624750499\\
446	64.187624750499\\
447	64.187624750499\\
448	64.187624750499\\
449	64.187624750499\\
450	64.187624750499\\
451	64.187624750499\\
452	64.187624750499\\
453	64.187624750499\\
454	64.187624750499\\
455	64.187624750499\\
456	64.187624750499\\
457	64.187624750499\\
458	64.187624750499\\
459	64.187624750499\\
460	64.187624750499\\
461	64.187624750499\\
462	64.187624750499\\
463	64.187624750499\\
464	64.187624750499\\
465	64.187624750499\\
466	64.187624750499\\
467	64.187624750499\\
468	64.187624750499\\
469	64.187624750499\\
470	64.187624750499\\
471	64.187624750499\\
472	64.187624750499\\
473	64.187624750499\\
474	64.187624750499\\
475	64.187624750499\\
476	64.187624750499\\
477	64.187624750499\\
478	64.187624750499\\
479	64.187624750499\\
480	64.187624750499\\
481	64.187624750499\\
482	64.187624750499\\
483	64.187624750499\\
484	64.187624750499\\
485	64.187624750499\\
486	64.187624750499\\
487	64.187624750499\\
488	64.187624750499\\
489	64.187624750499\\
490	64.187624750499\\
491	64.187624750499\\
492	64.187624750499\\
493	64.187624750499\\
494	64.187624750499\\
495	64.187624750499\\
496	64.187624750499\\
497	64.187624750499\\
498	64.187624750499\\
499	64.187624750499\\
500	64.187624750499\\
501	64.187624750499\\
};
\addlegendentry{mean 64.19};

\addplot [color=black,solid,line width=3.0pt]
  table[row sep=crcr]{%
1	53\\
2	53\\
3	53\\
4	53\\
5	53\\
6	53\\
7	53\\
8	53\\
9	53\\
10	53\\
11	53\\
12	53\\
13	53\\
14	53\\
15	53\\
16	53\\
17	53\\
18	53\\
19	53\\
20	53\\
21	53\\
22	53\\
23	53\\
24	53\\
25	53\\
26	53\\
27	53\\
28	53\\
29	53\\
30	53\\
31	53\\
32	53\\
33	53\\
34	53\\
35	53\\
36	53\\
37	53\\
38	53\\
39	53\\
40	53\\
41	53\\
42	53\\
43	53\\
44	53\\
45	53\\
46	53\\
47	53\\
48	53\\
49	53\\
50	53\\
51	53\\
52	53\\
53	53\\
54	53\\
55	53\\
56	53\\
57	53\\
58	53\\
59	53\\
60	53\\
61	53\\
62	53\\
63	53\\
64	53\\
65	53\\
66	53\\
67	53\\
68	53\\
69	53\\
70	53\\
71	53\\
72	53\\
73	53\\
74	53\\
75	53\\
76	53\\
77	53\\
78	53\\
79	53\\
80	53\\
81	53\\
82	53\\
83	53\\
84	53\\
85	53\\
86	53\\
87	53\\
88	53\\
89	53\\
90	53\\
91	53\\
92	53\\
93	53\\
94	53\\
95	53\\
96	53\\
97	53\\
98	53\\
99	53\\
100	53\\
101	53\\
102	53\\
103	53\\
104	53\\
105	53\\
106	53\\
107	53\\
108	53\\
109	53\\
110	53\\
111	53\\
112	53\\
113	53\\
114	53\\
115	53\\
116	53\\
117	53\\
118	53\\
119	53\\
120	53\\
121	53\\
122	53\\
123	53\\
124	53\\
125	53\\
126	53\\
127	53\\
128	53\\
129	53\\
130	53\\
131	53\\
132	53\\
133	53\\
134	53\\
135	53\\
136	53\\
137	53\\
138	53\\
139	53\\
140	53\\
141	53\\
142	53\\
143	53\\
144	53\\
145	53\\
146	53\\
147	53\\
148	53\\
149	53\\
150	53\\
151	53\\
152	53\\
153	53\\
154	53\\
155	53\\
156	53\\
157	53\\
158	53\\
159	53\\
160	53\\
161	53\\
162	53\\
163	53\\
164	53\\
165	53\\
166	53\\
167	53\\
168	53\\
169	53\\
170	53\\
171	53\\
172	53\\
173	53\\
174	53\\
175	53\\
176	53\\
177	53\\
178	53\\
179	53\\
180	53\\
181	53\\
182	53\\
183	53\\
184	53\\
185	53\\
186	53\\
187	53\\
188	53\\
189	53\\
190	53\\
191	53\\
192	53\\
193	53\\
194	53\\
195	53\\
196	53\\
197	53\\
198	53\\
199	53\\
200	53\\
201	53\\
202	53\\
203	53\\
204	53\\
205	53\\
206	53\\
207	53\\
208	53\\
209	53\\
210	53\\
211	53\\
212	53\\
213	53\\
214	53\\
215	53\\
216	53\\
217	53\\
218	53\\
219	53\\
220	53\\
221	53\\
222	53\\
223	53\\
224	53\\
225	53\\
226	53\\
227	53\\
228	53\\
229	53\\
230	53\\
231	53\\
232	53\\
233	53\\
234	53\\
235	53\\
236	53\\
237	53\\
238	53\\
239	53\\
240	53\\
241	53\\
242	53\\
243	53\\
244	53\\
245	53\\
246	53\\
247	53\\
248	53\\
249	53\\
250	53\\
251	53\\
252	53\\
253	53\\
254	53\\
255	53\\
256	53\\
257	53\\
258	53\\
259	53\\
260	53\\
261	53\\
262	53\\
263	53\\
264	53\\
265	53\\
266	53\\
267	53\\
268	53\\
269	53\\
270	53\\
271	53\\
272	53\\
273	53\\
274	53\\
275	53\\
276	53\\
277	53\\
278	53\\
279	53\\
280	53\\
281	53\\
282	53\\
283	53\\
284	53\\
285	53\\
286	53\\
287	53\\
288	53\\
289	53\\
290	53\\
291	53\\
292	53\\
293	53\\
294	53\\
295	53\\
296	53\\
297	53\\
298	53\\
299	53\\
300	53\\
301	53\\
302	53\\
303	53\\
304	53\\
305	53\\
306	53\\
307	53\\
308	53\\
309	53\\
310	53\\
311	53\\
312	53\\
313	53\\
314	53\\
315	53\\
316	53\\
317	53\\
318	53\\
319	53\\
320	53\\
321	53\\
322	53\\
323	53\\
324	53\\
325	53\\
326	53\\
327	53\\
328	53\\
329	53\\
330	53\\
331	53\\
332	53\\
333	53\\
334	53\\
335	53\\
336	53\\
337	53\\
338	53\\
339	53\\
340	53\\
341	53\\
342	53\\
343	53\\
344	53\\
345	53\\
346	53\\
347	53\\
348	53\\
349	53\\
350	53\\
351	53\\
352	53\\
353	53\\
354	53\\
355	53\\
356	53\\
357	53\\
358	53\\
359	53\\
360	53\\
361	53\\
362	53\\
363	53\\
364	53\\
365	53\\
366	53\\
367	53\\
368	53\\
369	53\\
370	53\\
371	53\\
372	53\\
373	53\\
374	53\\
375	53\\
376	53\\
377	53\\
378	53\\
379	53\\
380	53\\
381	53\\
382	53\\
383	53\\
384	53\\
385	53\\
386	53\\
387	53\\
388	53\\
389	53\\
390	53\\
391	53\\
392	53\\
393	53\\
394	53\\
395	53\\
396	53\\
397	53\\
398	53\\
399	53\\
400	53\\
401	53\\
402	53\\
403	53\\
404	53\\
405	53\\
406	53\\
407	53\\
408	53\\
409	53\\
410	53\\
411	53\\
412	53\\
413	53\\
414	53\\
415	53\\
416	53\\
417	53\\
418	53\\
419	53\\
420	53\\
421	53\\
422	53\\
423	53\\
424	53\\
425	53\\
426	53\\
427	53\\
428	53\\
429	53\\
430	53\\
431	53\\
432	53\\
433	53\\
434	53\\
435	53\\
436	53\\
437	53\\
438	53\\
439	53\\
440	53\\
441	53\\
442	53\\
443	53\\
444	53\\
445	53\\
446	53\\
447	53\\
448	53\\
449	53\\
450	53\\
451	53\\
452	53\\
453	53\\
454	53\\
455	53\\
456	53\\
457	53\\
458	53\\
459	53\\
460	53\\
461	53\\
462	53\\
463	53\\
464	53\\
465	53\\
466	53\\
467	53\\
468	53\\
469	53\\
470	53\\
471	53\\
472	53\\
473	53\\
474	53\\
475	53\\
476	53\\
477	53\\
478	53\\
479	53\\
480	53\\
481	53\\
482	53\\
483	53\\
484	53\\
485	53\\
486	53\\
487	53\\
488	53\\
489	53\\
490	53\\
491	53\\
492	53\\
493	53\\
494	53\\
495	53\\
496	53\\
497	53\\
498	53\\
499	53\\
500	53\\
501	53\\
};
\addlegendentry{median 53.00};

\end{axis}
\end{tikzpicture}%

%% file: figs/eval_tsukuba_gt_zoom.tex
%
%
\definecolor{mycolor1}{rgb}{0.00000,0.44706,0.74118}%
\definecolor{mycolor2}{rgb}{1.00000,0.00000,1.00000}%
\begin{tikzpicture}
\begin{axis}[%
width=\fwidth,
height=\fheight,
at={(0\fwidth,0\fheight)},
scale only axis,
plot box ratio=1.71 1 2.155,
unit vector ratio=1 1 1,
xmin=0.8,
xmax=2.1,
xmajorgrids,
ymin=-0.0,
ymax=-0.12628403,
ymajorgrids,
zmin=0.5,
zmax=1.9,
zmajorgrids,
view={0}{0},
axis background/.style={fill=white}
]
\addplot3 [color=green,dashed,line width=0.75pt]
 table[row sep=crcr] {%
1.03157303	-0.60013275	1.86917145\\
1.05176521	-0.59961182	1.86218506\\
1.07189163	-0.59905182	1.85486542\\
1.09193398	-0.59845749	1.84720184\\
1.11187897	-0.59783356	1.83918152\\
1.13171509	-0.59718475	1.83079025\\
1.15142967	-0.59651657	1.82201355\\
1.17100616	-0.5958345	1.81283813\\
1.19041962	-0.59514526	1.80325348\\
1.20967117	-0.5944545	1.79323639\\
1.22873276	-0.59376953	1.78277679\\
1.24758499	-0.59309769	1.77185974\\
1.26620712	-0.59244659	1.76046997\\
1.28457428	-0.5918248	1.74859497\\
1.30266251	-0.591241	1.73622101\\
1.32044617	-0.59070419	1.72333496\\
1.33790283	-0.59022415	1.70992157\\
1.35498505	-0.5898111	1.69598419\\
1.37167542	-0.58947479	1.68150574\\
1.38794769	-0.58920578	1.6664827\\
1.4037764	-0.5889621	1.6509256\\
1.41914017	-0.5886969	1.63484467\\
1.43401627	-0.58836151	1.61825317\\
1.44837799	-0.58790619	1.60117157\\
1.46222519	-0.58727859	1.58359161\\
1.47551682	-0.58642685	1.56555496\\
1.48822403	-0.58529755	1.54706665\\
1.50031357	-0.58383682	1.52814392\\
1.51176804	-0.58199142	1.50881775\\
1.52257248	-0.57970856	1.48911926\\
1.5327121	-0.57693695	1.46908432\\
1.5421727	-0.57362778	1.44875458\\
1.55094345	-0.56973557	1.42817291\\
1.55901535	-0.56521835	1.40738571\\
1.56638077	-0.56004044	1.38644653\\
1.5730368	-0.55417572	1.36540939\\
1.57899567	-0.54765091	1.34431244\\
1.58427658	-0.54052933	1.32318573\\
1.58889999	-0.53287033	1.30205078\\
1.59288605	-0.52472839	1.28092499\\
1.59625381	-0.5161554	1.25982361\\
1.599021	-0.50719788	1.2387558\\
1.60120407	-0.49789948	1.21772781\\
1.60281708	-0.4883017	1.19674652\\
1.60387329	-0.47843903	1.17580627\\
1.60438141	-0.46835175	1.1549173\\
1.60434937	-0.4580751	1.13408035\\
1.60378265	-0.44764206	1.11329208\\
1.60268402	-0.4370842	1.09254807\\
1.60105347	-0.42643463	1.07184586\\
1.59888779	-0.41572525	1.05118027\\
1.59618027	-0.40498703	1.03054138\\
1.59292252	-0.3942627	1.00994156\\
1.58910065	-0.38356125	0.98936829\\
1.58470398	-0.37287766	0.96886124\\
1.57971863	-0.36218903	0.9484346\\
1.5741333	-0.35147949	0.92811966\\
1.5679364	-0.34073044	0.90794479\\
1.56111801	-0.32992416	0.88794296\\
1.55366852	-0.31904144	0.86814606\\
1.54557999	-0.30806305	0.84858795\\
1.53684708	-0.29697113	0.82930695\\
1.52746582	-0.28574677	0.81033997\\
1.51743469	-0.27437195	0.79172562\\
1.50675659	-0.26283112	0.77350555\\
1.49543671	-0.25110794	0.75571976\\
1.48347931	-0.2391835	0.73840164\\
1.4709108	-0.22705887	0.72161102\\
1.45773987	-0.21471512	0.70537277\\
1.44398682	-0.2021666	0.68971298\\
1.4296524	-0.18946907	0.67461044\\
1.41475708	-0.17670456	0.66006378\\
1.39930054	-0.16393616	0.64605087\\
1.38328842	-0.15123199	0.63255783\\
1.36672897	-0.13866257	0.6195752\\
1.3496109	-0.12628403	0.60707977\\
};
 \addplot3 [color=black,solid,line width=1.1pt]
 table[row sep=crcr] {%
1.04551718153184	-0.604718413982214	1.88839121817924\\
1.06578249641746	-0.604042938014353	1.8814224271549\\
1.08616461986163	-0.603330619636463	1.87415556916064\\
1.10644294153461	-0.602615906432813	1.8666483099827\\
1.1263608614221	-0.601984131008772	1.85912357457984\\
1.14640739938893	-0.601237475925781	1.85077531364051\\
1.1663064521135	-0.600396834359302	1.84212563178224\\
1.18600145763326	-0.599713218860952	1.83325311306729\\
1.20540291911352	-0.599123306436139	1.82416188086917\\
1.22486498040352	-0.598247414558858	1.8140480733908\\
1.24399265107964	-0.597433090991932	1.80375271567983\\
1.26297829977182	-0.596755228327664	1.79308973429192\\
1.28162368819965	-0.596241106893376	1.78220855513728\\
1.30014343637253	-0.595341698260224	1.77022184665784\\
1.31855829485427	-0.594937419312415	1.75814953362961\\
1.3365923742034	-0.594378042897089	1.7453760503264\\
1.35423413638732	-0.593946839962012	1.73215409189644\\
1.37145308626863	-0.593460919401117	1.71821763278091\\
1.3885011448548	-0.593299189713516	1.70388012890558\\
1.40514674921103	-0.592971587830085	1.68890149423316\\
1.42106269286276	-0.592730764750204	1.67357278485179\\
1.43679320994024	-0.592580617540447	1.65768803086937\\
1.45195679502422	-0.592290874584848	1.64118230937173\\
1.46660151644685	-0.591776664989145	1.62420973457788\\
1.4805002673807	-0.591040442228058	1.60668041969912\\
1.49412328465693	-0.590264562515859	1.58884278988696\\
1.50709020774475	-0.589106501842682	1.57032144105946\\
1.51959462762043	-0.587752669235438	1.55147553855568\\
1.53098239696363	-0.585657111230408	1.5320432212635\\
1.54215422785065	-0.583525785892431	1.51244549523381\\
1.55249013322132	-0.580796562915947	1.49249481557706\\
1.56212807385316	-0.5775604362704	1.47205335263834\\
1.57133308197266	-0.573663987771634	1.45152469554237\\
1.57972296270755	-0.569215068925493	1.43071316561194\\
1.58717791909661	-0.564039844486182	1.40984777175737\\
1.59392621028994	-0.558038928195386	1.38868867755981\\
1.6000798569607	-0.551579937481964	1.36775881092778\\
1.60566178559681	-0.5445456177943	1.3465690079331\\
1.61030354176921	-0.536644685432885	1.32536513340226\\
1.61479178679138	-0.528581961405775	1.30423674956326\\
1.61856902204364	-0.520171338020318	1.28291888570924\\
1.62150469362976	-0.511035894093066	1.26186074386123\\
1.62447990036898	-0.502169392846096	1.24071361962147\\
1.62621414930835	-0.492401373882411	1.21978478861777\\
1.62750019138622	-0.482498111335109	1.1986001398603\\
1.62875832236597	-0.472550214352079	1.17759929719167\\
1.62897024470776	-0.462445376298874	1.15677031193448\\
1.62876261316168	-0.452118276403004	1.13578415716593\\
1.62825685377031	-0.441531706543543	1.11485720202376\\
1.62703145178712	-0.431055566046942	1.0942824922018\\
1.62497857930948	-0.420407764461746	1.07342296500091\\
1.62264188973515	-0.409505133943591	1.05243566550376\\
1.61956435320038	-0.39871653087173	1.03181900029305\\
1.61598609784245	-0.388042182828316	1.01096335809182\\
1.61177073602855	-0.377242647490116	0.990391815346231\\
1.60706534004754	-0.366515486831293	0.96985656629806\\
1.60181044095019	-0.355921858657199	0.949185119227711\\
1.59573548356976	-0.345098293732932	0.929006675907936\\
1.58908312641339	-0.334357756445024	0.908817922004264\\
1.58169798809073	-0.323251532563253	0.888747980931602\\
1.57388620564582	-0.312178296510735	0.869162419579231\\
1.56554678375335	-0.301284687294198	0.84973251525627\\
1.55644305539783	-0.289983952091483	0.830575926608163\\
1.54662190044769	-0.27860227134509	0.81167250216012\\
1.53601933599673	-0.266981681992334	0.793151579416974\\
1.52496552770684	-0.255465981422296	0.775151189184091\\
1.51299538759551	-0.24342144793052	0.757586034015223\\
1.50068049498976	-0.231172196739296	0.740615554291348\\
1.48754722892015	-0.218793565301188	0.724157626453711\\
1.47381857493906	-0.206212355040002	0.70863101732089\\
1.45949814098089	-0.193467383697717	0.693295411416722\\
1.4448653086953	-0.180759841279184	0.678569723441823\\
1.42901565259096	-0.167741840407737	0.66429406645699\\
1.41344988724548	-0.155647641195399	0.650380578972188\\
1.39638820973429	-0.142654470190311	0.63730326662274\\
1.3791456168618	-0.129889966548948	0.62464143821132\\
};
 \addplot3 [color=mycolor1,solid,mark size=0.2pt,mark=square,mark options={solid}]
 table[row sep=crcr] {%
0.951245273334234	-0.558830494407669	1.87253964997745\\
0.972295338480598	-0.558788107734612	1.86685519128804\\
0.992197758516289	-0.557976529504733	1.86021914625258\\
1.01434525615734	-0.558812934782301	1.85385809552338\\
1.0343447642478	-0.557659631877976	1.84692287309168\\
1.05400558188037	-0.556628323362929	1.83938419650437\\
1.07498636894402	-0.557006849411692	1.83222170416708\\
1.09496513726405	-0.556592214802735	1.82425741016807\\
1.11576528239158	-0.556731741923081	1.81597700245555\\
1.13456494715981	-0.555610769339582	1.80619330451754\\
1.15487085675801	-0.555161234407098	1.79734772994733\\
1.17405983225143	-0.554093759977016	1.78688125862197\\
1.19399135399382	-0.554246397459861	1.7766063149482\\
1.21261555847632	-0.553741309047758	1.76622235183236\\
1.2327548415251	-0.553720218045354	1.75510733547127\\
1.24877020174866	-0.553514646888405	1.74363964476891\\
1.26662574398713	-0.553331620068646	1.73136303144746\\
1.28405635158583	-0.553460434845824	1.71868871228033\\
1.30130319613538	-0.552387741032461	1.70447279938539\\
1.31870205781453	-0.553217437015512	1.69140256911578\\
1.33618448729709	-0.551730492964535	1.67489278920509\\
1.35216995649536	-0.551320386079136	1.65940017738549\\
1.36754384686979	-0.551771496060903	1.64514253181788\\
1.383508303109	-0.55211637588944	1.62937847038338\\
1.39929816611434	-0.551743349874666	1.61273245761361\\
1.41268551264219	-0.550364608451462	1.59481792579268\\
1.4252955007041	-0.54954223635745	1.57732678091599\\
1.43808717577967	-0.547018346340045	1.55784575800296\\
1.4488382313104	-0.543912172041018	1.5384045913746\\
1.46227274889206	-0.542974263993305	1.51901092880121\\
1.47321049961365	-0.540388838266198	1.49940946542385\\
1.48381799025178	-0.536733728790498	1.47931947908935\\
1.49291103767847	-0.532153977849151	1.45918337897753\\
1.50257931081764	-0.527889562451435	1.43941672504082\\
1.50963563300167	-0.52171735545094	1.41871065047104\\
1.51737027233759	-0.51494975685299	1.39749042276673\\
1.52496931252665	-0.508899695507812	1.37756444698889\\
1.53312035879098	-0.502571014890074	1.35699452111801\\
1.53703446013507	-0.493738305697465	1.33550846383303\\
1.54301506661345	-0.48556597126022	1.31435656656114\\
1.54637994738936	-0.475896478828261	1.29397552509493\\
1.55302009103303	-0.467809351706753	1.27468152711154\\
1.55683531021529	-0.458367197590464	1.25380334947444\\
1.55792064338228	-0.446864534654002	1.23238965640394\\
1.55914113207515	-0.460863199582246	1.28350657721475\\
1.56376627737143	-0.456475426102524	1.28877852435278\\
1.56579766542358	-0.446046065472917	1.27010017537489\\
1.57277317601886	-0.448570579066082	1.29268131266893\\
1.57387768878965	-0.436961192814668	1.27273888873421\\
1.5760571903839	-0.425711790680266	1.25199727111431\\
1.58939328321674	-0.429841792705726	1.27878680301493\\
1.59012275575076	-0.418086753635344	1.25822575122535\\
1.59025689571918	-0.406423667356386	1.23742557774378\\
1.59112941211149	-0.395122433332789	1.21736476685763\\
1.59079247471183	-0.3835009224098	1.19683554101281\\
1.5897612691573	-0.372098840801237	1.17688552614706\\
1.60731508223217	-0.362459026247804	1.18243038651286\\
1.62928033681392	-0.362333368015535	1.19400209458485\\
1.62674013870721	-0.351095061435055	1.17357924559053\\
1.6235561385069	-0.338877781014558	1.15321645763918\\
1.62037703190138	-0.327424049954136	1.13317334223587\\
1.61574310633617	-0.315971543353901	1.11374194955882\\
1.61293101142331	-0.302419222734475	1.09604610114225\\
1.60725520033171	-0.290612015107659	1.07711642144199\\
1.60073288526069	-0.278303040846551	1.05875444133729\\
1.59441930625488	-0.266180052425334	1.03999079594213\\
1.58659071099746	-0.253951301558367	1.02187100623048\\
1.57850279271617	-0.241217890151537	1.00397409522902\\
1.56940465876367	-0.228048601201136	0.985463027662938\\
1.55908551915809	-0.213894605583828	0.968457519851925\\
1.54939018502235	-0.20063274674342	0.951592569883553\\
1.5396572722877	-0.187389729541438	0.935555339619127\\
1.52952302178101	-0.174483837089983	0.91921399942167\\
1.51731048146222	-0.161424836540478	0.90356600859826\\
1.50533307531662	-0.148186089002637	0.887889996607117\\
1.49258586241229	-0.134350819219985	0.872728847890213\\
};
 \addplot3 [color=mycolor2,solid,mark size=0.75pt,mark=+,mark options={solid}]
 table[row sep=crcr] {%
0.880266087233564	-0.696220776653798	1.85211187043047\\
0.898902850867547	-0.696616542518974	1.84954061491736\\
0.918674899863754	-0.697698265788086	1.84679605483549\\
0.938318240733947	-0.699323259721661	1.84345331408059\\
0.957940962300672	-0.700277537917824	1.83981977154531\\
0.97783674799186	-0.701550134383534	1.83584013999623\\
0.99805701063349	-0.702771668194559	1.83151737930766\\
1.01820274880496	-0.704155949012825	1.82698421353864\\
1.03844662489173	-0.70557998520945	1.82193956428348\\
1.05877193084816	-0.706772250603431	1.81591188264821\\
1.07898937787385	-0.707428580438718	1.80938308420283\\
1.09950954872871	-0.708421706421785	1.80256673573042\\
1.11921293995605	-0.70967274570977	1.79519965809509\\
1.13893378150722	-0.710485341519158	1.78763367157662\\
1.15875423445214	-0.711645778401825	1.77971566275187\\
1.17900289450155	-0.712936688291473	1.76982253670646\\
1.19704952562219	-0.714019237580064	1.75941730248498\\
1.21551800501539	-0.715600676193231	1.74906820874671\\
1.23427301532238	-0.716396627825885	1.7377518342941\\
1.25268192938335	-0.717175431331489	1.72559089639007\\
1.27240047587782	-0.719020545574191	1.71354148910457\\
1.29062654116033	-0.719891881771207	1.7012127985269\\
1.30819968340631	-0.720302624618381	1.68797351996725\\
1.32582819062849	-0.720513378543245	1.67425694583703\\
1.34414015539963	-0.720782885515672	1.66003429963791\\
1.3602758469601	-0.719418899845064	1.64545599747389\\
1.37660116651175	-0.718387113186099	1.63044465496158\\
1.38981507028133	-0.713694817005362	1.61507558026018\\
1.40513345287922	-0.711978418134968	1.59938872050092\\
1.40513345287922	-0.711978418134968	1.59938872050092\\
1.42126785215942	-0.710134360067965	1.58339401522404\\
1.43270069285173	-0.703229100496465	1.56775911896379\\
1.43270069285173	-0.703229100496465	1.56775911896379\\
1.43270069285173	-0.703229100496465	1.56775911896379\\
1.43270069285173	-0.703229100496465	1.56775911896379\\
1.43270069285173	-0.703229100496465	1.56775911896379\\
1.43270069285173	-0.703229100496465	1.56775911896379\\
1.43270069285173	-0.703229100496465	1.56775911896379\\
1.43270069285173	-0.703229100496465	1.56775911896379\\
1.43270069285173	-0.703229100496465	1.56775911896379\\
1.4498184935287	-0.69729401311741	1.55264117612054\\
1.46601067794677	-0.690673516797964	1.53845169619125\\
1.48325366582693	-0.684094292878479	1.5236577550475\\
1.51291827547951	-0.686867709965961	1.50488324328074\\
1.54339354350845	-0.689767691814404	1.48480228723373\\
1.56719938452323	-0.689574513732765	1.46813328304904\\
1.59065959642887	-0.689846166838216	1.45176374870745\\
1.62024447574527	-0.693065762429801	1.43030041450267\\
1.64430951617435	-0.693475365448874	1.41204367835362\\
1.66487609361492	-0.691020716480031	1.39381245025251\\
1.69025419999335	-0.692601442254947	1.37484619811215\\
1.71519409452691	-0.693163583891643	1.35336339199124\\
1.73607946656026	-0.692425144580806	1.33671534876774\\
1.75339135157195	-0.68975012931867	1.32087391699769\\
1.77970069046387	-0.693529285693214	1.30004694629596\\
1.80297386583007	-0.694649381730909	1.27774166845549\\
1.82005003506073	-0.692344959181554	1.2602032401047\\
1.84149126835935	-0.694794532761884	1.24054344423238\\
1.85463449572601	-0.689389014251396	1.22015746916709\\
1.86572935787708	-0.682623781359929	1.20070987127482\\
1.87758078263068	-0.676470882206086	1.17932225042896\\
1.88486109397624	-0.667549071802894	1.15899907822121\\
1.89076046609093	-0.657710607418991	1.13926981431799\\
1.8939099562275	-0.645713126691908	1.12014519281654\\
1.89605384378155	-0.633294979810506	1.10113818897239\\
1.89679637803228	-0.619659273874318	1.08267588178967\\
1.89625200150018	-0.606358072824401	1.06541308371092\\
1.89371871718183	-0.590858976140446	1.0509493685449\\
1.89107665221482	-0.577016537653352	1.03743494391405\\
1.88827915185235	-0.563244305704638	1.02255700207473\\
1.88383656638006	-0.551471057384725	1.01166950244712\\
1.87690932690547	-0.537682002478942	0.999267150166043\\
1.86901599236222	-0.522784296010961	0.986555573151246\\
1.86173122610868	-0.507245304210871	0.972209848422928\\
1.85392348444795	-0.491651235759296	0.957320958843921\\
1.84546808503666	-0.474961579928896	0.943144373809622\\
};
 \end{axis}
\end{tikzpicture}%

%% file: figs/kitti_eval_latest.tex
%
%
\begin{tikzpicture}

\begin{axis}[%
width=0.412\fwidth,
height=0.417\fheight,
at={(0\fwidth,0.583\fheight)},
scale only axis,
xmin=100,
xmax=800,
xlabel={Path Length [m]},
xmajorgrids,
ymin=1.4245,
ymax=5.2866,
ylabel={Translation [\%]},
ymajorgrids,
axis background/.style={fill=white}
]
\addplot [color=black,solid,line width=1.0pt,mark=+,mark options={solid},forget plot]
  table[row sep=crcr]{%
100	2.1686\\
200	2.4059\\
300	2.6252\\
400	2.7927\\
500	2.9074\\
600	2.966\\
700	2.9501\\
800	3.0137\\
};
\addplot [color=white!30!black,dashed,line width=1.0pt,mark=o,mark options={solid},forget plot]
  table[row sep=crcr]{%
100	2.5984\\
200	3.0791\\
300	3.5648\\
400	4.01\\
500	4.4326\\
600	4.7414\\
700	4.9874\\
800	5.2866\\
};
\addplot [color=white!70!black,solid,line width=1.0pt,mark=square,mark options={solid},forget plot]
  table[row sep=crcr]{%
100	1.4245\\
200	1.7086\\
300	2.1131\\
400	2.4794\\
500	2.8073\\
600	3.0865\\
700	3.2694\\
800	3.4543\\
};
\end{axis}

\begin{axis}[%
width=0.412\fwidth,
height=0.417\fheight,
at={(0.542\fwidth,0.583\fheight)},
scale only axis,
xmin=100,
xmax=800,
xlabel={Path Length [m]},
xmajorgrids,
ymin=0.00412529612494193,
ymax=0.0138082828626528,
ylabel={Rotation Error [deg/m]},
ymajorgrids,
axis background/.style={fill=white}
]
\addplot [color=black,solid,line width=1.0pt,mark=+,mark options={solid},forget plot]
  table[row sep=crcr]{%
100	0.00509932437666433\\
200	0.00521391593569049\\
300	0.00509932437666433\\
400	0.00475554969958583\\
500	0.00446907080202042\\
600	0.00429718346348117\\
700	0.00412529612494193\\
800	0.00418259190445501\\
};
\addplot [color=white!30!black,dashed,line width=1.0pt,mark=o,mark options={solid},forget plot]
  table[row sep=crcr]{%
100	0.00922462050160625\\
200	0.0101413529738156\\
300	0.0106570149894333\\
400	0.0107716065484595\\
500	0.0107716065484595\\
600	0.0106570149894333\\
700	0.0103705360918679\\
800	0.0103705360918679\\
};
\addplot [color=white!70!black,solid,line width=1.0pt,mark=square,mark options={solid},forget plot]
  table[row sep=crcr]{%
100	0.0138082828626528\\
200	0.012547775713365\\
300	0.011860226359208\\
400	0.0111726770050511\\
500	0.0105997192099202\\
600	0.0101986487533287\\
700	0.00991216985576324\\
800	0.009740282517224\\
};
\end{axis}

\begin{axis}[%
width=0.412\fwidth,
height=0.417\fheight,
at={(0\fwidth,0\fheight)},
scale only axis,
xmin=14.4,
xmax=86.4,
xlabel={Speed [km/h]},
xmajorgrids,
ymin=1.4806,
ymax=33.3811,
ylabel={Translation Error [\%]},
ymajorgrids,
axis background/.style={fill=white}
]
\addplot [color=black,solid,line width=1.0pt,mark=+,mark options={solid},forget plot]
  table[row sep=crcr]{%
14.4	3.4907\\
21.6	2.2341\\
28.8	1.9044\\
36	1.8664\\
43.2	2.1872\\
50.4	2.3598\\
57.6	2.3967\\
64.8	8.8362\\
72	12.8054\\
79.2	13.6943\\
86.4	18.0474\\
};
\addplot [color=white!30!black,dashed,line width=1.0pt,mark=o,mark options={solid},forget plot]
  table[row sep=crcr]{%
14.4	3.3174\\
21.6	2.6947\\
28.8	2.4045\\
36	2.3943\\
43.2	2.5942\\
50.4	2.1298\\
57.6	2.7886\\
64.8	11.8856\\
72	19.3349\\
79.2	22.8833\\
86.4	33.3811\\
};
\addplot [color=white!70!black,solid,line width=1.0pt,mark=square,mark options={solid},forget plot]
  table[row sep=crcr]{%
14.4	1.4806\\
21.6	2.6249\\
28.8	2.4264\\
36	2.2185\\
43.2	1.9242\\
50.4	1.7607\\
57.6	1.7809\\
64.8	2.6994\\
72	2.773\\
79.2	2.7473\\
86.4	2.4187\\
};
\end{axis}

\begin{axis}[%
width=0.412\fwidth,
height=0.417\fheight,
at={(0.542\fwidth,0\fheight)},
scale only axis,
xmin=14.4,
xmax=86.4,
xlabel={Speed [km/h]},
xmajorgrids,
ymin=0.00326585943224569,
ymax=0.0678954987230026,
ylabel={Rotation Error [deg/m]},
ymajorgrids,
axis background/.style={fill=white},
legend style={legend cell align=left,align=left,draw=white!15!black}
]
\addplot [color=black,solid,line width=1.0pt,mark=+,mark options={solid}]
  table[row sep=crcr]{%
14.4	0.0154698604685322\\
21.6	0.00635983152595214\\
28.8	0.00469825392007275\\
36	0.00366692988883727\\
43.2	0.00360963410932419\\
50.4	0.00441177502250734\\
57.6	0.00326585943224569\\
64.8	0.00624523996692597\\
72	0.00635983152595214\\
79.2	0.00538580327422974\\
86.4	0.00762033867523995\\
};
\addlegendentry{Raw Intensity};

\addplot [color=white!30!black,dashed,line width=1.0pt,mark=o,mark options={solid}]
  table[row sep=crcr]{%
14.4	0.0203400017271442\\
21.6	0.0101413529738156\\
28.8	0.00784952179329228\\
36	0.00618794418741289\\
43.2	0.00572957795130823\\
50.4	0.00504202859715124\\
57.6	0.00968298673771091\\
64.8	0.0373568482425297\\
72	0.0482430463500153\\
79.2	0.0521391593569049\\
86.4	0.0678954987230026\\
};
\addlegendentry{Bit-Planes};

\addplot [color=white!70!black,solid,line width=1.0pt,mark=square,mark options={solid}]
  table[row sep=crcr]{%
14.4	0.0139801702011921\\
21.6	0.011860226359208\\
28.8	0.0105997192099202\\
36	0.0100840571943025\\
43.2	0.0115164516821295\\
50.4	0.0103705360918679\\
57.6	0.0087662542655016\\
64.8	0.00928191628111934\\
72	0.0140947617602183\\
79.2	0.0143239448782706\\
86.4	0.0116310432411557\\
};
\addlegendentry{Viso2};

\end{axis}
\end{tikzpicture}%